\documentclass[journal]{IEEEtran}
\usepackage{amsmath,amsfonts}
\usepackage{algorithmic}
\usepackage{algorithm}
\usepackage{array}
\usepackage[caption=false,font=normalsize,labelfont=sf,textfont=sf,labelformat=simple]{subfig} %
\usepackage{textcomp}
\usepackage{stfloats}
\usepackage{url}
\usepackage{verbatim}
\usepackage{graphicx}
\usepackage{cite}
\usepackage{multirow} 
\usepackage{booktabs}
\usepackage{makecell}
\usepackage{pifont}     %
\usepackage[T1]{fontenc} 
\usepackage{hyperref}  
\usepackage[capitalize]{cleveref} 
\hypersetup{hidelinks,colorlinks=true,allcolors=black,pdfstartview=Fit,breaklinks=true}
\usepackage{color} 

\begin{document}

\title{Dual-frequency Selected Knowledge Distillation with Statistical-based Sample Rectification for PolSAR Image Classification}
        
\author{Xinyue Xin, Ming Li,~\IEEEmembership{Member,~IEEE,} Yan Wu,~\IEEEmembership{Member,~IEEE,} Xiang Li, Peng Zhang,~\IEEEmembership{Member,~IEEE,} and Dazhi Xu 

\thanks{
Xinyue Xin, Ming Li, Peng Zhang and Dazhi Xu are with the National Laboratory of Radar Signal Processing, Xidian University, Xi'an 710071, China, and also with the Collaborative Innovation Center of Information Sensing and Understanding, Xidian University, Xi'an 710071, China (e-mail:liming@xidian.edu.cn).

Yan Wu is with the Remote Sensing Image Processing and Fusion Group, School of Electronics Engineering, Xidian University, Xi'an 710071, China (e-mail:ywu@mail.xidian.edu.cn).

Xiang Li is with the Beijing Institute of Radio Measurement, Beijing 100854, China.}}

\maketitle
\begin{abstract}
The collaborative classification of dual-frequency PolSAR images is a meaningful but also challenging research. The effect of regional consistency on classification information learning and the rational use of dual-frequency data are two main difficulties for dual-frequency collaborative classification. To tackle these problems, a selected knowledge distillation network with statistical-based sample rectification (SKDNet-SSR) is proposed in this article. First, in addition to applying CNN and ViT as local and global feature extractors, a statistical-based dynamic sample rectification (SDSR) module is designed to avoid the impact of poor regional consistency on spatial information learning process. Specifically, based on the fact that the PolSAR covariance matrix conforms to the complex Wishart distribution, SDSR first dynamically evaluates the sample purity, and then performs pixel selection and pixel generation to remove noisy pixels, thereby avoiding the feature interaction between informative pixels and noisy pixels and improving the classification feature extraction process. Next, a dual-frequency gate-selected distillation (DGSD) module is constructed to emphasize the advantages of different frequency bands and perform complementary learning on dual-frequency data. It uses the dominant single-frequency branch on each sample as teacher model to train the dual-frequency student model, enabling the student model to learn the optimal results and realizing complementary utilization of dual-frequency data on different terrain objects. Comprehensive experiments on four measured dual-frequency PolSAR data demonstrate that the proposed SKDNet-SSR outperforms other related methods.  
\end{abstract}


\section{Introduction}
\IEEEPARstart{D}{ue} to its advantage of earth observing in all-time and all-weather conditions, polarimetric synthetic aperture radar (PolSAR) plays an important role in land cover classification\cite{2023_C}, segmentation\cite{2020_Seg}, change detection\cite{2020_CD}, object detection\cite{2023_OD}, etc. Among them, the PolSAR image classification task, which determines the ground area as grassland, ocean, urban, road, etc. by using obtained electromagnetic wave signal, is useful for plenty of practical applications, i.e., precision agriculture, urban development, environmental protection, and so on. 

In order to achieve accurate classification, numerous PolSAR image classification methods have emerged, which can be divided into two categories: algorithms based on traditional expert knowledge and algorithms based on deep learning. Among them, the algorithms based on traditional expert knowledge mainly utilize polarimetric scattering mechanism or polarimetric statistical models. First, the typical polarimetric scattering mechanism-based methods include Pauli decomposition \cite{Pauli_1996}, Freeman decomposition \cite{Freeman_1998}, Cloude-Pottier method \cite{Cloude_1997}, and so on. Wherein the Pauli decomposition \cite{Pauli_1996} is a coherent target decomposition method and is suitable for determining targets with stable scattering mechanism. The Freeman decomposition \cite{Freeman_1998} is an incoherent target decomposition method and is suitable for distributed targets. The Cloude-Pottier method\cite{Cloude_1997} uses entropy $H$ and angle $\alpha$ to divide feature space into eight regions, and each region represents a category. Second, due to the unique statistical distribution of PolSAR data, researchers have successively proposed Wishart distribution\cite{2015_TIP_Nie_wishart}, Wishart generalized gamma distribution\cite{2021_GRSL_Song_WGGMRF}, and mixed Wishart generalized gamma distribution\cite{2019_TIP_Liu_WMM}, in order to achieve increasingly accurate description for PolSAR data. By combining these distribution models with maximum likelihood estimation\cite{1999_TGRS_JSLee_wML, 2017_TGRS_Bi_wML}, Markov random field\cite{2015_TGRS_MRF, 2018_TGRS_Song_MRF}, and triplet Markov field\cite{2014_GRSL_Liu_TMF}, the classification of PolSAR images can be achieved. Based on the effectiveness of the polarimetric scattering mechanism and statistical models, some researchers have attempted to combine these two principles for classification. For example, Freeman et al.\cite{JSLee_1999} proposed to get the preliminary classification results by the Cloude-Pottier method, and then iterate based on the Wishart distribution to achieve more accurate classification results. Cao et al.\cite{Cao_2007} proposed to combine the backward scattering power with entropy $H$ and angle $\alpha$ to obtain the initial cluster center, and then use the Wishart distribution for hierarchical clustering.

In addition to the traditional algorithms mentioned above, there are also classification algorithms based on deep learning. Due to its ability to automatically extract deep discriminative features, deep learning-based classification algorithms have gradually become the mainstream in the field of PolSAR image interpretation. As the most classic neural network, convolutional neural network (CNN) has been successfully applied for PolSAR image classification\cite{2016_GRSL_Zhou_CNN,2018_TGRS_tbCNN}. Furthermore, stacked autoencoder (SAE)\cite{2018_GRSL_Hu_SAE, 2022_JSTARS_Gui_SAE}, graph neural network (GNN)\cite{2023_TGRS_Bi_GCN, 2023_NC_Xu} and vision Transformer (ViT)\cite{2024_JSTARS_Xin_ViT, 2022_TGRS_Dong_ViT} are also effective image classification frameworks. Among them, the combination of CNN and ViT, which together utilize the advantages of CNN in local spatial information learning and ViT in global spatial information learning, has become the most commonly used framework for PolSAR image classification\cite{2023_TGRS_ExViT, 2024_TGRS_Geng_CNNViT}. The essence of CNN is to realize information interaction at close range through the local convolutional kernel, and the essence of ViT is to realize information interaction at far range through the self-attention block.

\begin{figure}[!t]
	\centering
	{\includegraphics[width=3in,height=1.6in]{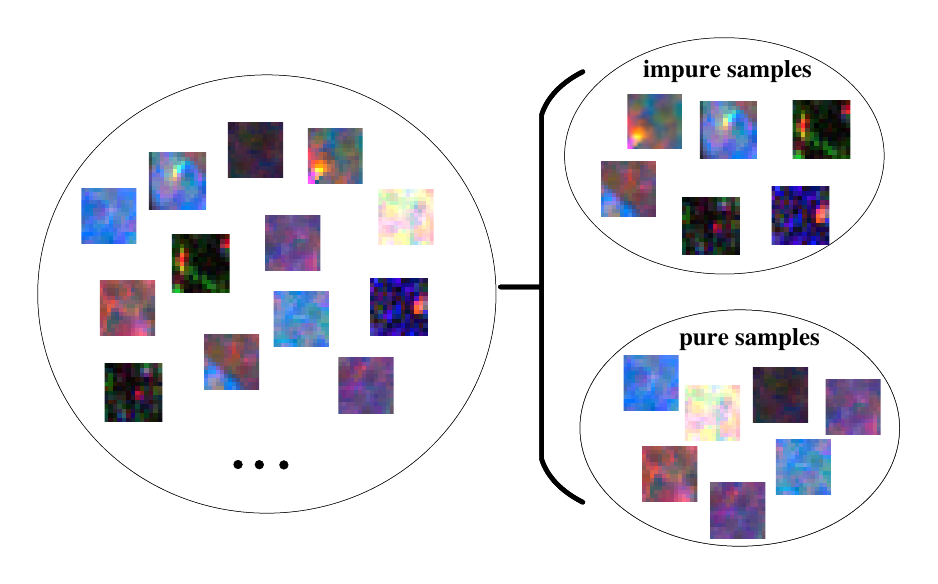}%
		\label{sample1}}
	\hfil
	\caption{Samples from the Flevoland\_CL dataset.}
	\label{Fig_sample}
\end{figure}

These deep learning-based models automatically extract advanced features from the image spatial structure, which has the advantage of being fast and efficient. However, due to the complex real-world scenarios and the coherent speckle imaging mechanism, there are some PolSAR samples with poor regional consistency\cite{2012_M_PolSAR, 2011_M_PolSAR}. As shown in Fig. \ref{Fig_sample}, it is clear that in pure samples, all pixels have similar properties, whereas in impure samples, there are some isolated pixels that differ significantly from the majority of pixels. We refer to the majority of similar pixels in the sample, i.e., pixels relevant to classification, as informative pixels, and the minority of isolated pixels, i.e., pixels irrelevant to classification, as noisy pixels. The essence of classification information learning is to learn advanced features related to classification task, and filter out features that are irrelevant to classification or even weaken classification performance. For neural network-based classification information learning process, all pixels within the sample are considered in the same category and equally participate in local and global feature interactions. For example, in CNN, the convolutional kernel learns spatial information by sliding over the sample space, and the operations in different regions are the same. In ViT, the self-attention operation learns spatial information by constructing query, key, and value in global sample space. Both these two typical networks traverse all pixels within the sample to obtain spatial information. However, when informative pixels and noisy pixels learn from each other, unnecessary computations are added and classification feature extraction is affected, thereby reducing the classification accuracy. To solve this problem, some studies\cite{2023_CVPR_Long, 2023_CVPR_Wei, 2023_CVPR_Li} have proposed token pruning strategies to select topK patches to participate in advanced feature extraction process, thus avoiding background noisy pixels from affecting the classification information learning. However, these hard-sparsification methods, which used the same topK for all samples, may result in discarding some informative pixels in pure samples (over-sparsification) or retaining some noisy pixels in impure samples (under-sparsification). And if an adaptive topK is used, the number of patches for each sample is inconsistent, which does not meet the batch processing requirements of neural networks. Therefore, how to select a reasonable topK for each sample and solve the inconsistency of the sample size after patch sparsification, so as to perform classification information learning in the case of poor region consistency is still to be solved.

Besides, since electromagnetic waves of different frequencies have different sensitivity to ground objects, single-frequency PolSAR data cannot obtain excellent observation results in complex ground environments. To this end, with the breakthrough development of remote sensing equipment, dual-frequency PolSAR data have gradually emerged in remote sensing research. The dual-frequency collaborative classification refers to the process of utilizing the rich information brought by dual-frequency data for accurate classification. At present, there are two frameworks for dual-frequency collaborative classification algorithms: single-branch framework and dual-branch framework. Among them, the single-branch framework first uses stacking\cite{2014_RS_Gao} or sparsity\cite{2015_RS_Yang} to obtain fused features, and then inputs the fused features into the single-branch network for advanced feature extraction and classification. This framework separates feature fusion from classification task and has advantage of simplicity. However, the increased feature dimension may increase computational complexity, and the loss of some important features may decrease accuracy. Therefore, current researches often use the dual-branch framework\cite{2018_TGRS_tbCNN}. Based on the consistency and complementarity between dual-frequency data, the dual-branch framework uses two networks to learn two frequency data separately, and performs dual-frequency fusion in the feature extraction stage or classification stage. For example, in the feature extraction stage, advanced feature representations that balance richness and compactness can be obtained based on attention mechanisms\cite{2022_TGRS_Cao,2022_ISPRS_CMGFNet} or feature constraints\cite{2022_TGRS_Xin}. In the classification stage, the classification results of each branch can be used as classification confidence to guide the fusion process\cite{2024_JSTARS_Xin_ViT}, or the consistency of different classification results can be constrained to obtain joint decisions\cite{2023_TGRS_Cao}. The above dual-frequency PolSAR image classification algorithms have been validated in some measured datasets. However, compared with mature developed single-frequency PolSAR image classification algorithms, studies on dual-frequency data are still limited. Besides, if dual-frequency data is processed by two stacked single-frequency networks, the two branches are treated equally and the single-frequency branch with poorer classification performance may reduce the joint results. In addition, most of the developed networks are trained by hard labels, ignoring the information contained in the numerical results. According to the existing research on knowledge distillation\cite{2015_KD}, the output of the neural network not only shows the classification results, but also implies the differences between categories and the similarity between categories. Therefore, it is an urgent problem to be solved that how to reasonably utilize the soft outputs of different frequency data and perform dual-frequency complementary fusion to obtain more accurate results.

To cope with the above-mentioned challenges, a selected knowledge distillation network with statistical-based sample rectification (SKDNet-SSR) is designed for dual-frequency PolSAR image collaborative classification. On the one hand, in addition to applying CNN and ViT as local and global feature extractors, a statistical-based dynamic sample rectification (SDSR) module is designed to improve classification information learning under poor regional consistency conditions. Specifically, based on the classification results of each sample, SDSR first evaluates the sample purity to adaptively obtain the topK for each sample. Then, the SDSR sequentially performs pixel selection and pixel generation based on the statistical Wishart distribution, thus obtaining reconstructed samples and avoiding feature interaction between noisy pixels and informative pixels. On the other hand, a dual-frequency gate-selected distillation (DGSD) module is proposed to enhance the learning of dual-frequency complementarity over different terrain objects. On the basis of knowledge distillation technology, the DGSD adds a teacher branch selection operation. By selecting the single-frequency branch with better classification results on each sample as the teacher model, the dual-frequency student model can learn the optimal results for each sample. In addition, through the cooperation of hard label loss and soft label loss, the supplementary information contained in another branch is also considered during the training process.

The major contributions of this work are summarized as follows:

\begin{enumerate}
	\item{We develop an SKDNet-SSR for dual-frequency PolSAR image collaborative classification. It effectively avoids the impact of poor regional consistency on PolSAR image classification information learning, and solves the problem of complementary utilization between dual-frequency data.}
	\item{To reduce the impact of poor regional consistency on PolSAR image classification information learning, an SDSR module is introduced. It not only adaptively evaluates the sample purity to obtain the topK value, but also designs a statistical-based pixel selection unit and a statistical-based pixel generation unit. By filtering out noisy pixels and balancing the sequence length of each sample, irrelevant noisy pixels are prevented from participating in classification information learning, and the advanced feature extraction process is improved.}
	\item{To utilize the complementarity of dual-frequency PolSAR data for joint classification, a DGSD module is proposed. It takes the single-frequency branch with better classification performance on each sample as the teacher model, and then trains the dual-frequency student model based on the knowledge distillation principle. These two steps enable the student model to learn the most outstanding single-frequency branch, and realize the effective utilization of the dual-frequency complementarity on different terrain objects.}
\end{enumerate}

The remainder of this paper is organized as follows. In Section \ref{section2}, the related works about remote sensing image classification and knowledge distillation are briefly introduced. Section \ref{section3} detailed the network architecture of the proposed method. Experimental results on four measured PolSAR datasets are displayed in Section \ref{section4}. Finally, the conclusion with a possible future outlook is shown in Section \ref{section5}.

\section{Preliminaries}\label{section2}

\subsection{CNN and ViT for remote sensing image classification}
CNN is a classic image feature extractor that utilizes local convolution kernels to learn image advanced information. For remote sensing image classification, Zhou et al.\cite{2016_GRSL_Zhou_CNN} converted the polarimetric coherence matrix into a 9-D real-valued vector as network input, achieving the first utilization of CNN in PolSAR images. Yang et al.\cite{2022_TGRS_SepDGConv} further proposed a separable dynamic grouping convolution network, which made grouping convolution a hyperparameter and enabled the entire network architecture learnable during the network training. Subsequently, numerous CNN-based algorithms are proposed for different task requirements. For example, Xu et al.\cite{2018_TGRS_tbCNN} proposed using two branch CNNs to process dual-frequency PolSAR data separately, achieving dual-frequency PolSAR image joint classification; Hosseinpour et al. \cite{2022_ISPRS_CMGFNet} constructed an encoder-decoder structure and gate fusion module, achieving feature extraction and multimodal feature fusion; Li et al. \cite{2023_TNNLS_AsyFFNet} used the scaling factor of batch normalization to determine the redundancy of the current feature layer, eliminating unnecessary features and improving network generalization ability. 

Transformer\cite{2017_arXiv_Transformer} was first proposed for natural language processing (NLP), and it excels in learning global information over long distances. ViT\cite{2021_arXiv_ViT} can be seen as the image processing version of Transformer, which directly migrates the standard Transformer model to image processing with minimal modifications. To be specific, ViT divides the image into many patches to form a linear embedding sequence, which is then used as the input of Transformer to simulate phrase sequence in NLP. Since then, a series of improved algorithms for ViT have been proposed, such as Swin Transformer\cite{2021_SwinTran}, which utilized sliding window and window attention to reduce computational complexity; Transformer in Transformer\cite{2021_TNT}, which combined intra-window self-attention and inter-window self-attention to achieve local-global spatial information learning; mCrossPA\cite{2023_TGRS_mCrossPA}, which used class tokens as representative information of different modalities and achieved complementary learning through cross token attention.

In response to the advantages of CNN in local spatial information learning and ViT in global spatial information learning, the combinations of CNN and ViT have gradually appeared. For example, CvT\cite{2021_ICCV_Wu_CvT} proposed the convolutional token embedding and convolutional mapping, effectively introducing the translation invariance, scaling, and other properties of CNN into Transformer; CeiT\cite{2021_CeiT} replaced the feedforward network in Transformer with a locally enhanced feedforward network to promote correlation between adjacent tokens. In remote sensing image interpretation tasks, Ding et al. \cite{2022_TGRS_GLT} proposed a global-local Transformer for the joint classification of hyperspectral images and lidar data, which not only learned the local-global spatial information through the combination of CNN and ViT, but also designed a multi-scale feature fusion strategy to improve the classification performance; Yao et al. \cite{2023_TGRS_ExViT} proposed to concatenate ViT and separable CNN to form a feature extraction network, and processed remote sensing data of different modalities separately through parallel branches; Geng et al. \cite{2024_TGRS_Geng_CNNViT} proposed a hierarchical scattering-spatial interaction Transformer, which not only learned the local-global spatial information through the combination of CNN and ViT, but also designed a Transformer-based scattering feature learning branch to realize the effective extraction of scattering features and spatial features in PolSAR images.
\begin{figure*}[!t]
	\centering
	\includegraphics[width=7in,height=2.8in]{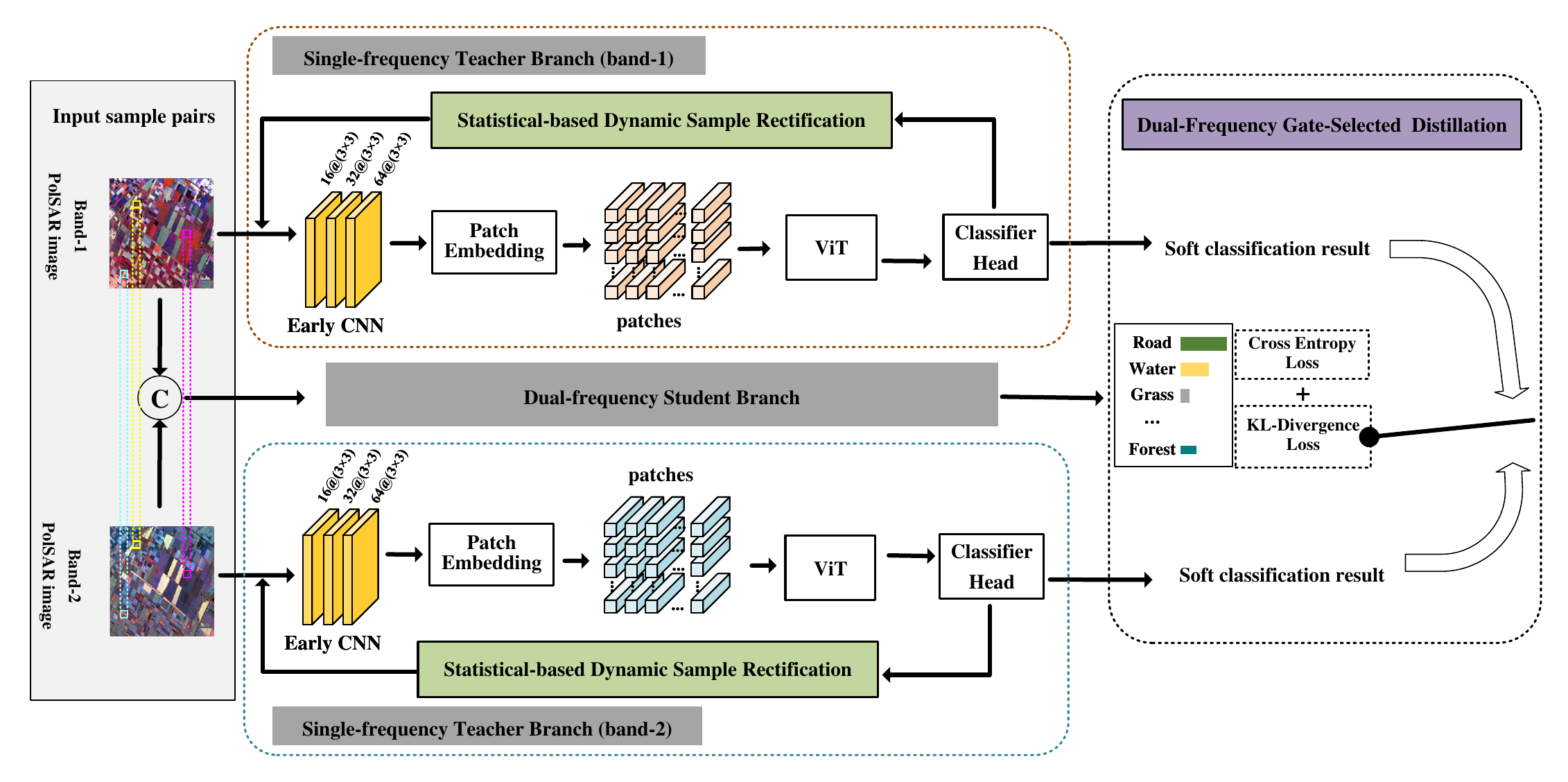}
	\caption{Overview of the proposed SKDNet-SSR.}
	\label{Fig1}
\end{figure*}
\subsection{knowledge distillation}
Knowledge distillation (KD) \cite{2015_KD} is a technique that transfers knowledge from one model to another, initially applied in model compression. Generally, the pre-trained model is referred as teacher model and the model that requires training is referred as student model. The purpose of KD is to enable the student model to fit the output of teacher model. Due to the fact that large-scale deep neural networks pose a significant challenge for devices with limited memory and computing power, KD is widely used for its ability to transfer knowledge from a large teacher model to a lightweight student model without any significant performance loss.
	
In multi-modal scenarios, KD can effectively integrate information from different modalities and deliver it to student model, so that complementary information from inter-modality can be borrowed by student model to get better results. Various works have been proposed to demonstrate the feasibility of KD in multi-modal tasks. For example, in response to the issue of inherent heterogeneities in cross-modal classification tasks, Li et al.\cite{2023_CVPR_KD} proposed a decoupled multimodal distillation approach to decouple each modality into two parts: modal-irrelevant features and modal-exclusive features, and designed graph distillation unit for each decoupled part to achieve dynamic cross-modal knowledge distillation. Similar cross-modal distillation methods also use optical RGB images for the teacher model and optical flow images for the student model\cite{2016_CVPR_KD}. As for the issue of modal missing in multi-modal research, Chen et al.\cite{2022_TMI_KD} proposed using multi-modal data for teacher model to deliver information to the unimodal student model, so as to leverage the privileged knowledge of multimodal data; Wei et al.\cite{2023_TGRS_KD} designed a general diversity-guided distillation network that transfers knowledge by matching feature diversity between the teacher model and the student model, thereby achieving accurate inference when modalities are missing. Besides, in order to achieve information fusion between modalities, Xu et al.\cite{2024_TGRS_KD} proposed a distillation in model (DIM) module, which used an encoder-decoder structure to fuse multimodal remote sensing data and guided the encoder training with the final fused information; Xue et al.\cite{2021_ICCV_KD} utilized the unimodal teacher model to obtain pseudo labels of unlabeled samples, and then trained the multimodal student model through both true labels of labeled samples and pseudo labels of unlabeled samples, achieving semi-supervised multimodal fusion.

The proposed SKDNet-SSR distinguishes itself from previous KD works by employing a teacher branch selection operation before knowledge transfer. By selecting the single-frequency branch with better classification results on each sample as the teacher model, the dual-frequency student model can learn the optimal results for each sample. Besides, SKDNet-SSR considers the supplementary information contained in another branch by combining the hard label loss and soft label loss, leading to significant performance boosts on PolSAR image classification task.

\section{Proposed Method}\label{section3}

\subsection{Method Overview}
Dual-frequency PolSAR data is obtained by transmitting and receiving electromagnetic waves in different frequency bands. Different frequencies of electromagnetic waves have varying sensitivities to terrain objects. For example, the C-band PolSAR data performs well in sea ice detection, while the L-band PolSAR data has advantages in crop recognition. In addition, the neural network-based feature exactors learn classification information equally among pixels within the sample, allowing some noisy pixels to participate in information interaction and affect the final classification results. These theories inspire us to design an SKDNet-SSR for dual-frequency PolSAR collaborative classification. The overall architecture of the proposed SKDNet-SSR is presented in Fig. \ref{Fig1}.

First, the two frequency PolSAR data is fed into two single-frequency teacher branches (band-1 and band-2). These two branches are trained independently and do not affect each other. Therefore, it is only necessary to ensure that the inputs of the two branches are from the same region. Each branch takes the combination of early CNN, patch embedding, ViT, and SDSR as the feature exactor. Among them, early CNN and ViT are used to extract local-global spatial information within the sample; patch embedding serves as a sample format conversion operation between CNN and ViT, which can convert samples from image blocks to sequences; the SDSR enhances the regional consistency by discarding irrelevant noisy pixels and adaptively rectificating samples, so that the classification information extractions by CNN and ViT are only performed among informative pixels. Next, the two well-trained single-frequency branches are used for DGSD. Specifically, the single-frequency branch with better classification results is selected as the teacher model for each sample, while the stacked dual-frequency PolSAR data is fed into the student model. The student model is trained using both the soft labels of the selected teacher model and the hard labels of the ground truth, so as to learn the optimal single-frequency classification results and realize the utilization of dual-frequency complementarity.

\subsection{Single-frequency Teacher Branch}

\subsubsection{Early CNN}
The PolSAR image classification task generally selects the covariance matrix as the original feature \cite{2016_GRSL_Zhou_CNN}. For neural network-based classification algorithms, the input is a 3-D matrix, with the first two dimensions being the height and width of the sample, and the third dimension being the number of features. Therefore, we convert the polarimetric covariance matrix on each pixel into a 9-D real-valued vector and use window sliding to create samples. Specifically, the polarimetric covariance matrix of each pixel is:
\begin{equation}\label{equ_1_1}
\text{C} = 
\begin{bmatrix}
C_{11} & C_{12}  & C_{13}  \\ 
C_{12}^{*} & C_{22}  & C_{23}  \\ 
C_{13}^{*} & C_{23}^{*}  & C_{33}  
\end{bmatrix}.
\end{equation}
The superscript $*$ represents the complex conjugate operation. And the corresponding 9-D vector is [$C_{11}$, $Re(C_{12})$, $Im(C_{12})$, $Re(C_{13})$, $Im(C_{13})$, $C_{22}$, $Re(C_{23})$, $Im(C_{23})$, $C_{33}$], where $Re(\cdot)$ and $Im(\cdot)$ represent the real and imaginary part, respectively. And then, we select a rectangular box of size $s\times s$ for overlapping window segmentation to obtain samples with a size of $(s\times s \times 9)$.

As shown in the single-frequency teacher branch in Fig. \ref{Fig1}, for a training sample $x\in \mathbb{R}^{s\times s\times 9}$, the early CNN is first applied to learn local spatial information:
\begin{equation}\label{equ_1_2}
x'= CNNs(x)=W_{3}(W_{2}(W_{1}*x+b_{1})+b_{2})+b_{3}.
\end{equation}
Each CNN layer consists of three items: convolution, batch normalization (BN), and rectified linear unit (ReLU). The early CNN block can supplement the learning of local spatial information ignored in subsequent ViT and provide more information by increasing the input dimension.

\subsubsection{patch embedding and ViT}
Based on the basic framework of ViT \cite{2022_TGRS_Dong_ViT}, $x'$ is first divided into non overlapping patches through the patch embedding block. The patches and the class patch "cls" are combined to form $p=\left\{cls, p_{1}, p_{2},..., p_{n}, ..., p_{N}\right\}$, where $N$ represents the number of patches. The patch embedding block is a significant improvement of ViT compared to Transformer. This improvement is due to the fact that the traditional Transformer is designed for sequential input. When applying Transformer to image domain, the image needs to be divided into multiple patches and tiled into sequences.

And then, ViT completes global spatial information learning through three components: self-attention, multi-layer perceptron (MLP), and residual connection. Among them, the self-attention is generally considered as the core component of ViT, which interacts information over all patches to achieve feature learning on the global space:
\begin{equation}\label{equ_1_3}
p'= (p \times W^{v}) \times SoftMax( \frac{(p \times W^{Q})(p \times W^{K})^{T}}{\sqrt{d}})
\end{equation}
where $p'$ is the patches after self-attention operation, $d$ is the feature dimension used to ensure the gradient stability of softmax. $W^{Q}$, $W^{K}$and $W^{V}$ represent the query weight matrix, key weight matrix, and value weight matrix. The superscript $T$ means transpose operation.

After ViT, the "cls" is fed into the classifier head as representative of all patches to obtain the classification result $out = [y_{1},...,y_{m},...,y_{M}]$, $M$ is the number of categories. Based on the principle of softmax in multi-category classification, the $y_{m}$ is the $m$th item of $out$ and also the probability that the sample belongs to the $m$th class.

\subsubsection{statistical-based dynamic sample rectification (SDSR)}
Due to its advantage in learning local and global spatial information, CNN and ViT can effectively extract advanced feature representations from PolSAR images. However, the inherent regional mixing makes some PolSAR samples contain not only informative pixels but also irrelevant noisy pixels. If CNN and ViT are directly used for learning advanced classification features, noisy pixels may lead to misclassification and reduce classification accuracy. To address this issue, we propose using an SDSR module to adaptively remove noisy pixels, and retain only informative pixels for information interaction.

\begin{figure*}[!t]
	\centering
	\includegraphics[width=6.5in]{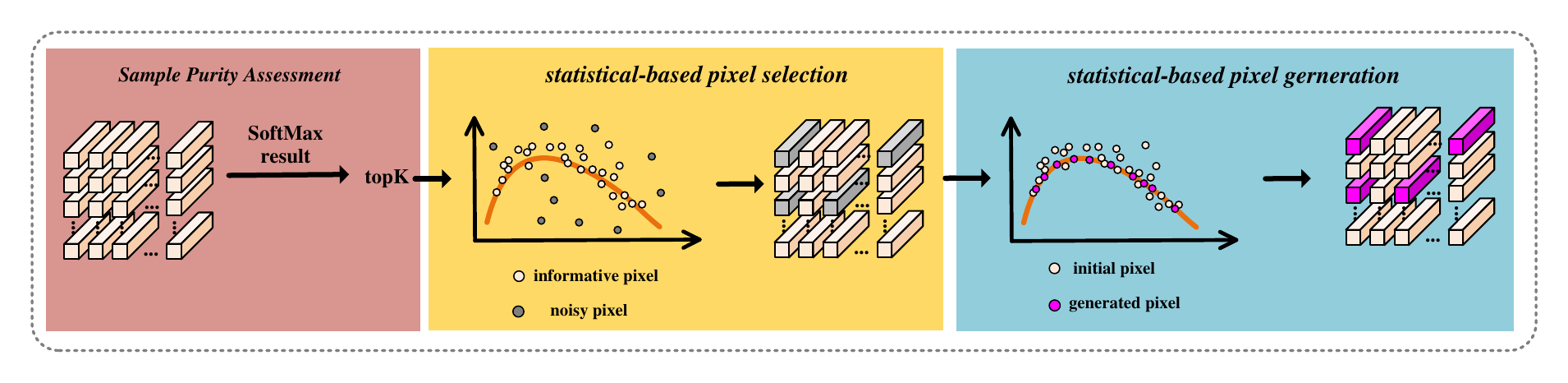}
	\caption{The detail of the SDSR module.}
	\label{Fig2}
\end{figure*}

Specifically, as shown in Fig. \ref{Fig2}, the SDSR consists of three steps. The first step is sample purity assessment, which is used to obtain the topK value; the second step is statistical-based pixel selection, which is used to discard the noisy pixels and retain the informative pixels; the third step is statistical-based pixel generation, which is used to generate new informative pixels to replace the noisy pixels discarded in the second step.

As mentioned earlier, the "cls" is fed into the classifier head as representative of all patches to obtain the classification result. In the first step of SDSR, in order to evaluate the sample purity, we input the advanced features of each patch into the classifier head to obtain a set of classification results: $\left\{out_1, out_2, ..., out_n, ..., out_N\right\}$, where $out_n$ represents the classification result of $n$th patch. If the sample is pure, then all patches should belong to the same category, i.e., the indexes of the maximum value in classification results are the same for all patches. On the contrary, if the sample is impure, then there exist some patches whose indexes do not coincide with the index of the "cls". Therefore, we take the proportion of patches that are consistent with "cls" as the sample purity $r$, and determine the number of pixels to be retained: $topK = r*s^2$.

After determining the number of pixels that need to be retained, the next question is which pixels should be kept and which ones should be rectificated. As presented in Fig. \ref{Fig2}, a statistical-based pixel selection unit is designed to separate informative pixels and noisy pixels. Considering the small sample size and the difficulty of calculating complex statistical models, we choose the Wishart distribution to fit the PolSAR data in the following steps.

In statistical-based pixel selection unit, we take the mean covariance matrix of each sample as the sample center, and then measure the statistical Wishart distance between all pixels and the sample center. The specific formula of statistical Wishart distance $d_{w}$ between a pixel with covariance matrix as $C_i$ and a sample with feature center as $\Sigma$ is:
\begin{equation}
d_{w}(C_i,\Sigma)=Tr(\Sigma^{-1}C_i) +In|\Sigma|.
\end{equation}
Where $\Sigma=\frac{1}{s^2} \sum_{i=1}^{s^2} C_i$, $In(\cdot)$ represents the logarithmic operation and $Tr(\cdot)$ represents finding the trace of a matrix. After measuring the statistical distance between each pixel and the sample center, the topK pixels with minimum distance are considered as informative pixels to retain, while other pixels are considered as noisy pixels to discard. 

The above operations make the sequence length of each sample relate to its own situation, which avoids the over-sparsification or under-sparsification caused by using the fixed topK value. However, for the convenience of batch processing in neural networks, the size of input samples should be consistent. If topK pixels are adaptively selected to update samples, the length of the sample vectors will be different, making it impossible to batch process and increasing training difficulty. In response to this issue, a statistical-based pixel generation unit is developed. 

In statistical-based pixel generation unit, we first capture the corresponding statistical Wishart distribution based on the sample center $\Sigma$, and then perform sampling to generate new pixel-level features. The probability density function (PDF) of complex Wishart distribution is:
\begin{equation}\label{equ_1_5}
p_{w}=\frac{L^{Lq}|C|^{L-q}exp(-LTr(\Sigma ^{-1}C))}{K(L,q)|\Sigma|^{L}},
\end{equation}
where $q$ is the dimensionality of the scattering vector, $L$ represents the number of looks, $|\cdot|$ represents finding the trace of a matrix. $K(L,q)=\pi^{q(q-1/)/2}\prod_{i=1}^{q} \Gamma(L-q+1)$ is the Gamma function. Then, the generated new pixels are used to replace the original noisy pixels to form reconstructed samples, thereby solving the problem of inconsistent sample size after adaptive pixel sparsification.

Overall, the sample purity assessment, the statistical-based pixel selection unit, and the statistical-based pixel generation unit jointly ensure that the global-local information interactions are carried out on informative pixels, avoiding irrelevant noisy pixels from participating in classification information learning and improving the advanced feature extraction process.

\subsection{Dual-frequency Gate-Selected Distillation (DGSD) module}
The DGSD module concludes two steps. First, based on the classification results of two single-frequency branches, DGSD selects the branch with better classification performance as the teacher model for each sample. Second, based on the KD technique of the teacher-student model, DGSD transmits information from teacher to student through the combination of soft labels and hard labels. Below, we will provide a detailed introduction to these two steps.

\subsubsection{teacher branch selection}
Given two single-frequency PolSAR data, two independent single-frequency branches are trained by using cross entropy loss functions. Assuming that $out^{1} = [y^{1}_{1},...,y^{1}_{m},...,y^{1}_{M}]$ and $out^{2} = [y^{2}_{1},...,y^{2}_{m},...,y^{2}_{M}]$ represent the outputs of band-1 branch and band-2 branch respectively, $y$ represents the one-hot encoding of real label. The cross entropy loss functions of two single-frequency branches are:
\begin{align}\label{equ_2_1}
loss_{1} = -\frac{1}{M} \sum_{i=1}^{M} (y(i)log(out^{1}(i))),\\
loss_{2} = -\frac{1}{M} \sum_{i=1}^{M} (y(i)log(out^{2}(i))).
\end{align}
The two single-frequency branches are trained separately, resulting in two independently well-trained classification networks. 

Due to the varying sensitivity of electromagnetic waves in different frequencies to terrain objects, $out^{1}$ and $out^{2}$ are different for the same sample. If band-1 performs better on sample $x$, i.e. $loss_{1}$ is smaller than $loss_{2}$, then the maximum value of $out^{1}$ (which is also the class-index entry of $out^{1}$) should be greater than the maximum value of $out^{2}$ (which is also the class-index entry of $out^{2}$). Based on this property, we can select the dominant single-frequency branch as the teacher model for sample $x$:
\begin{equation}\label{equ_2_5}
Model_{t}=
\left\{
\begin{array}{lr}
Model_{t1},\ if \ max(out^{1}) \geq max(out^{2}), & \\
Model_{t2},\ if \ max(out^{1}) < max(out^{2}). &
\end{array}
\right. 
\end{equation}
where $Model_{t1}$ and $Model_{t2}$ represent the single-frequency teacher branch of band-1 and band-2. For example, if $out^{1}=[0.05, 0.80, 0.10, 0.05]$, $out^{2}=[0.10, 0.60, 0.25, 0.05]$, their maximum values are $0.80$ and $0.60$ respectively. Obviously, the classification performance of the band-1 branch is better, so we choose band-1 branch as the teacher model for this sample.

After the gated selection described above, the dominant branch for each sample can be obtained and used as the teacher model, achieving the purpose of teacher branch selection. This process lays a good foundation for subsequent knowledge transfer, which allows the dual-frequency student model to learn the optimal results in two frequency bands, thus capturing different advantages in different frequency bands and realizing the complementary exploitation of dual-frequency data on different terrain objects.

\subsubsection{transferring information from teacher to student}
For the dual-frequency student model, we take the concatenation of two single-frequency data as input, and the student model learns advanced features by simulating the output of teacher model, regardless of the input dimension. The framework of student model is similiar to the teacher model, including early CNN, patch embedding, ViT, and SDSR.

According the the research of KD \cite{2015_KD}, in addition to positive labels, the negative labels of the softmax output also contain a lot of information for network inductive reasoning. For example, if the probability corresponding to a certain negative label is much higher than other negative labels, it indicates that the sample has a certain similarity with that negative label during inference. In the traditional hard target training process, all negative labels are treated uniformly, the differences between categories are emphasized and the similarities between categories are ignored. Therefore, the training method of KD allows each sample to bring more information to the student model than traditional training methods. The soft label KL-divergence $L_{KL}$ between teacher model and student model is:
\begin{align}\label{equ_2_2}
L_{KL} = \sum_{i=1}^{M} (out_{t}(i)log(\frac{out_{t}(i)}{out_{s}(i)})),
\end{align}where $out_{t}$ and $out_{s}$ represent the softmax output of teacher model and the softmax output of student model respectively.

Besides, as the teacher model is trained on single-frequency data, its soft results show inter-class differences and inter-class similarities under that single-frequency data. The performance of different frequency data on the same sample varies, that is, the numerical results of $out^1$ and $out^2$ are different, and their biased negative labels may also be different. For example, if $out^1=[0, 0.2, 0.8, 0]$ and $out^2=[0.2, 0, 0.8, 0]$, these two sets of results both indicate that the sample belongs to the third category. But $out1$ suggests that the sample has a certain similarity with the second category in band-1 PolSAR data, while $out^2$ suggests that the sample has a certain similarity with the first category in band-2 PolSAR data. If the training method of KD is strictly followed, that is, using the advantage branch as the teacher model for knowledge transfer, the dual-frequency student model will ignore the supplementary information brought by another frequency PolSAR data. To alleviate this issue, we have incorporated traditional hard label loss, namely:
\begin{align}\label{equ_2_3}
L_{CE} = -\frac{1}{M} \sum_{i=1}^{M} (y(i)log(out_{s}(i))).
\end{align}

As shown in Fig. \ref{Fig1}, after selecting the teacher model, we perform backpropagation by minimizing the soft label KL-divergence $L_{KL}$ between teacher model and student model, as well as the cross entropy loss $L_{CE}$ between the student model and the hard label, thus realizing network training.
\begin{align}\label{equ_2_4}
L_{all} = \alpha * L_{KL} + (1-\alpha) * L_{CE}
\end{align}
where $\alpha$ is the balance coefficient of $L_{KL}$ and $L_{CE}$.

When $\alpha=0$, the student model is only trained through hard labels, while when $\alpha=1$, the training process of the student model is performed solely by KD. The $\alpha$ helps to prevent the student model from overfitting the teacher model with some bias, thus improving the model's generalization ability.

\section{Experimental Result and Analysis} \label{section4}
In this section, we will first introduce the measured datasets, algorithms for comparison, and experimental settings. Second, the ablation experiments will be conducted to verify the effectiveness of the proposed SDSR module and DGSD module. Finally, we will compare the proposed SKDNet-SSR with several related algorithms and provide a detailed analysis.

\subsection{Description of Dataset}
\subsubsection{Hebei Dataset}
\begin{figure}[!t]
	\centering
	\subfloat[]{\includegraphics[width=1in]{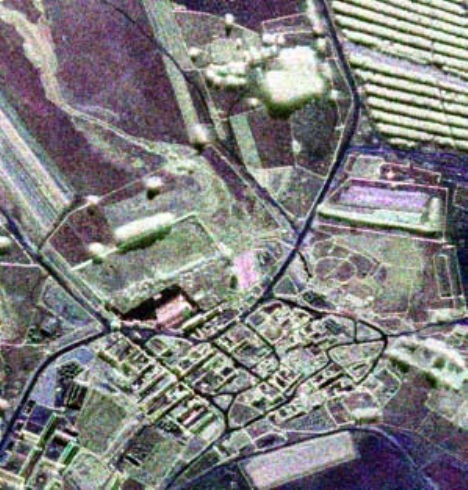}%
		\label{Pauli_HB_S}}
	\hfil
	\subfloat[]{\includegraphics[width=1in]{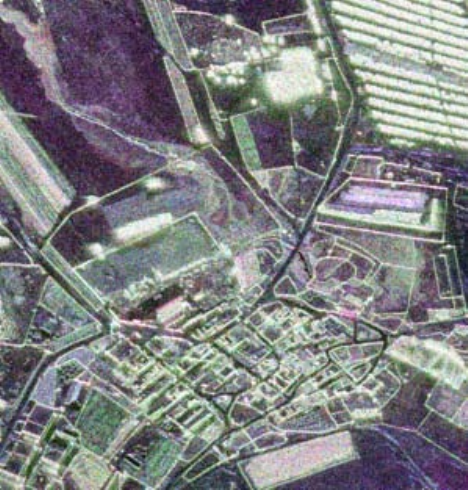}%
		\label{Pauli_HB_L}}
	\hfil
	\subfloat[]{\includegraphics[width=1in]{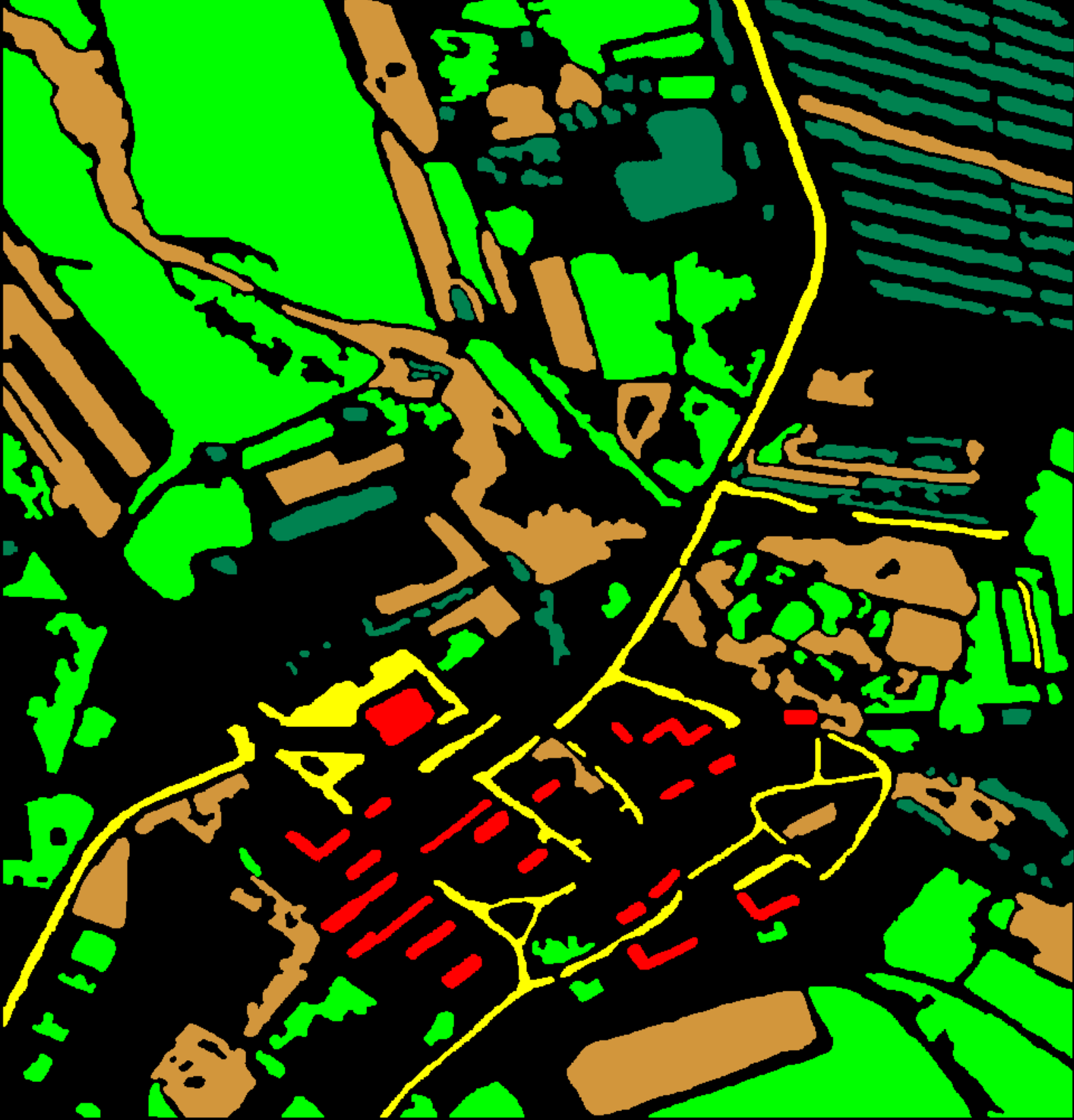}%
		\label{GroundTruth_HB}}
	\hfil
	\subfloat{\includegraphics[width=3in]{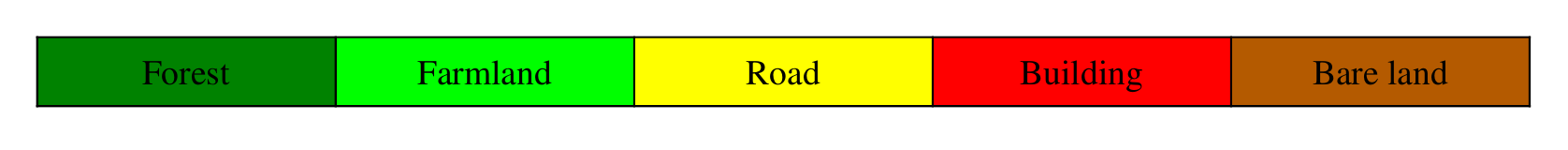}}
	\caption{The Hebei dataset. (a) S-band. (b) L-band. (c) Ground Truth.}
	\label{Fig_HB}
\end{figure}

The Hebei dataset consists of two frequency bands PolSAR data (S-band and L-band, they are both obtained by the airborne system), forming a dual-frequency dataset: Hebei\_SL. The image size of this dataset is 1005$\times$962. As shown in Fig. \ref{Fig_HB}, the categories included in this dataset are: Forest, Farmland, Road, Building, and Bare land. Each category is marked with different colors on ground truth map. In the following experiments, 500 samples (100 samples for each category) are selected as training sample set. 

\subsubsection{SanFrancisco Dataset}
\begin{figure}[!t]
	\centering
	\subfloat[]{\includegraphics[width=1in]{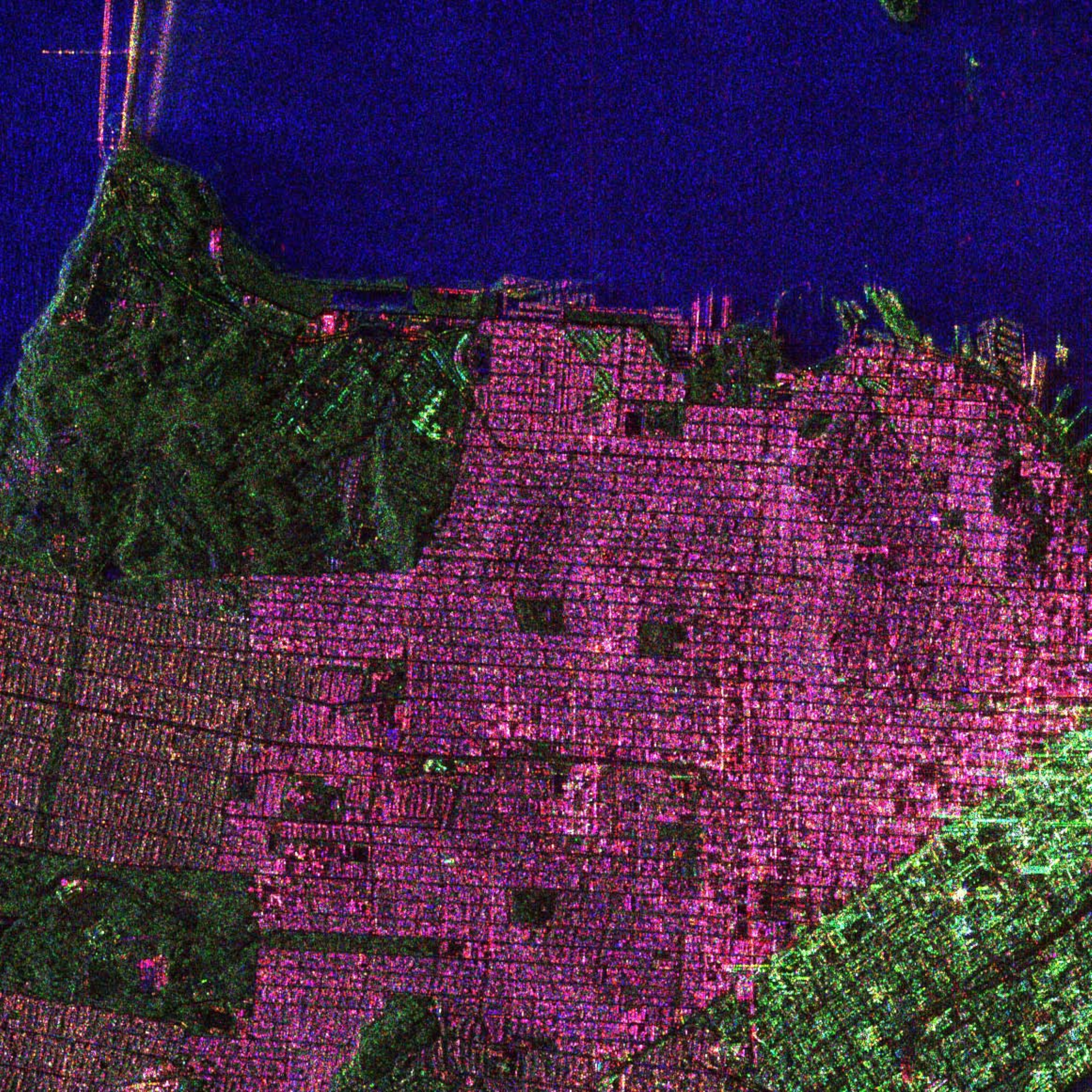}%
		\label{Pauli_San_C}}
	\hfil
	\subfloat[]{\includegraphics[width=1in]{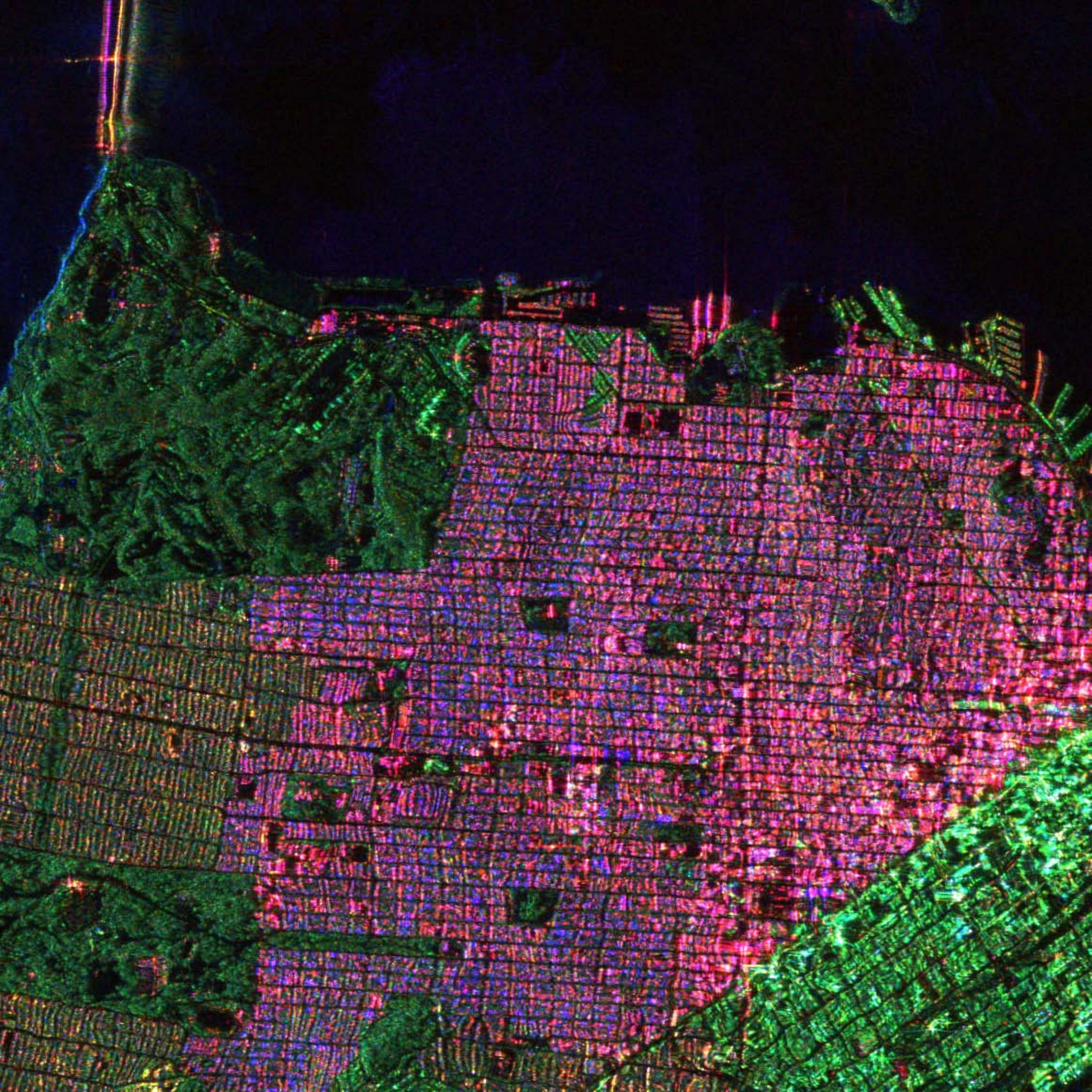}%
		\label{Pauli_San_L}}
	\hfil
	\subfloat[]{\includegraphics[width=1in]{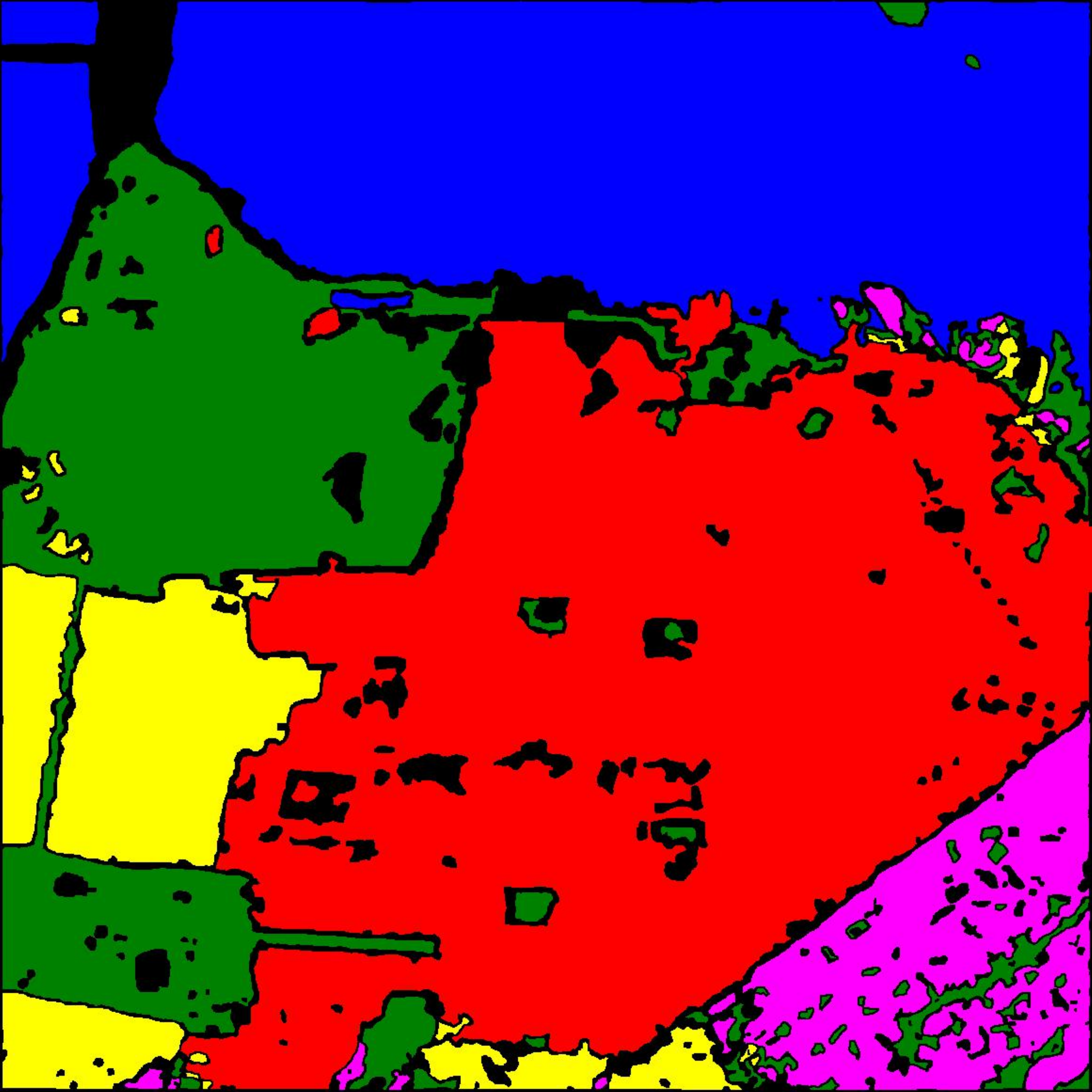}%
		\label{GroundTruth_San}}
	\hfil
	\subfloat{\includegraphics[width=3in]{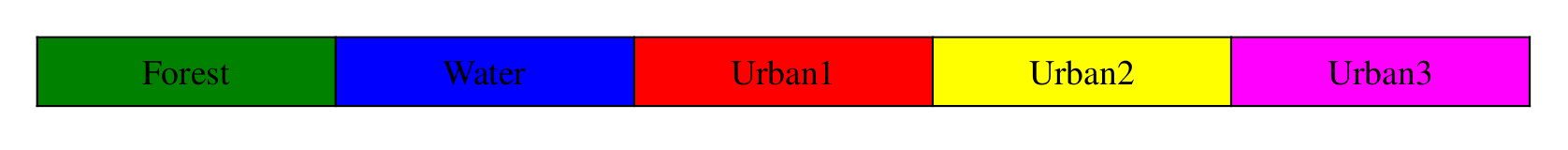}}  
	\caption{The SanFrancisco dataset. (a) C-band. (b) L-band. (c) Ground Truth.}
	\label{Fig_San}
\end{figure}

The SanFrancisco dataset has two frequency bands PolSAR data (C-band data is obtained by the GF-3 system, L-band data is obtained by the ALOS system), forming a dual-frequency dataset: SanFrancisco\_CL. The image size is 1161$\times$1161. As shown in Fig. \ref{Fig_San}, the categories included in this dataset are: Forest, Water, Urban1, Urban2, and Urban3. Each category is marked with different colors on ground truth map. In the following experiments, 250 samples (50 samples for each category) are selected as training sample set. 


\begin{figure}[!t]
	\centering
	\subfloat[]{\includegraphics[width=1.3in,height=1.1in]{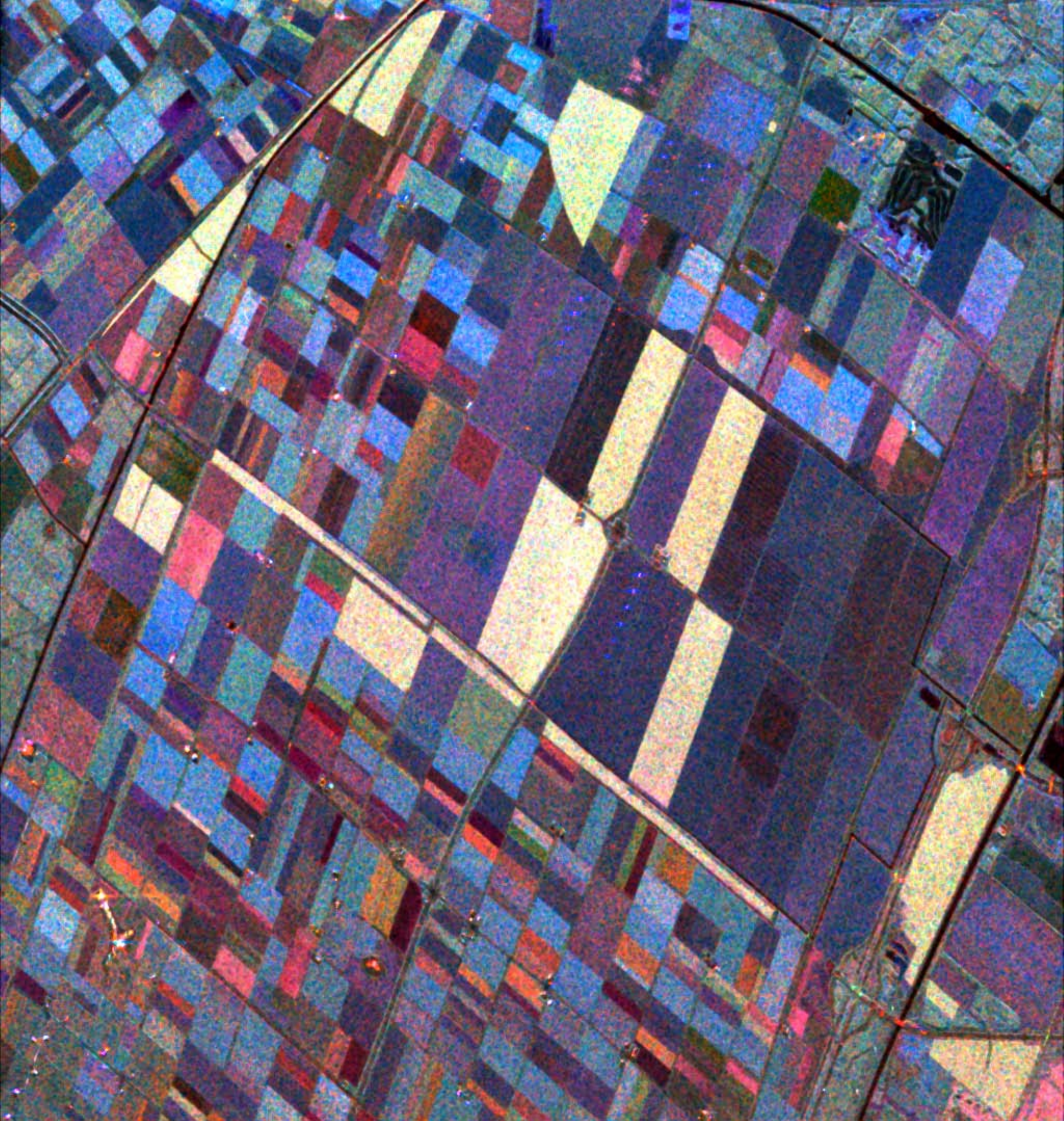}%
		\label{Pauli_Fle_C}}
	\hfil
	\subfloat[]{\includegraphics[width=1.3in,height=1.1in]{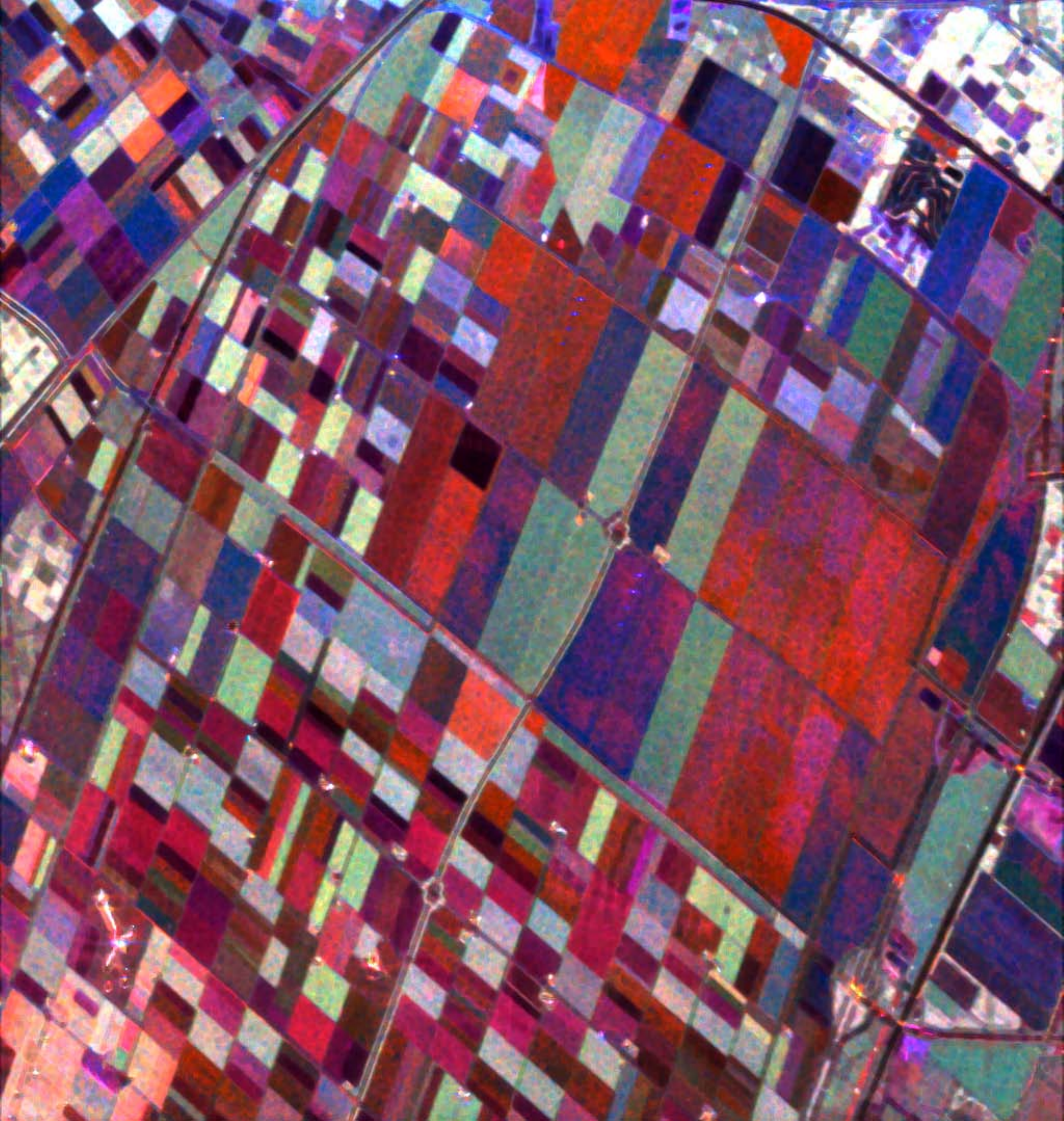}%
		\label{Pauli_Fle_L}}
	\hfil
	\subfloat[]{\includegraphics[width=1.3in,height=1.1in]{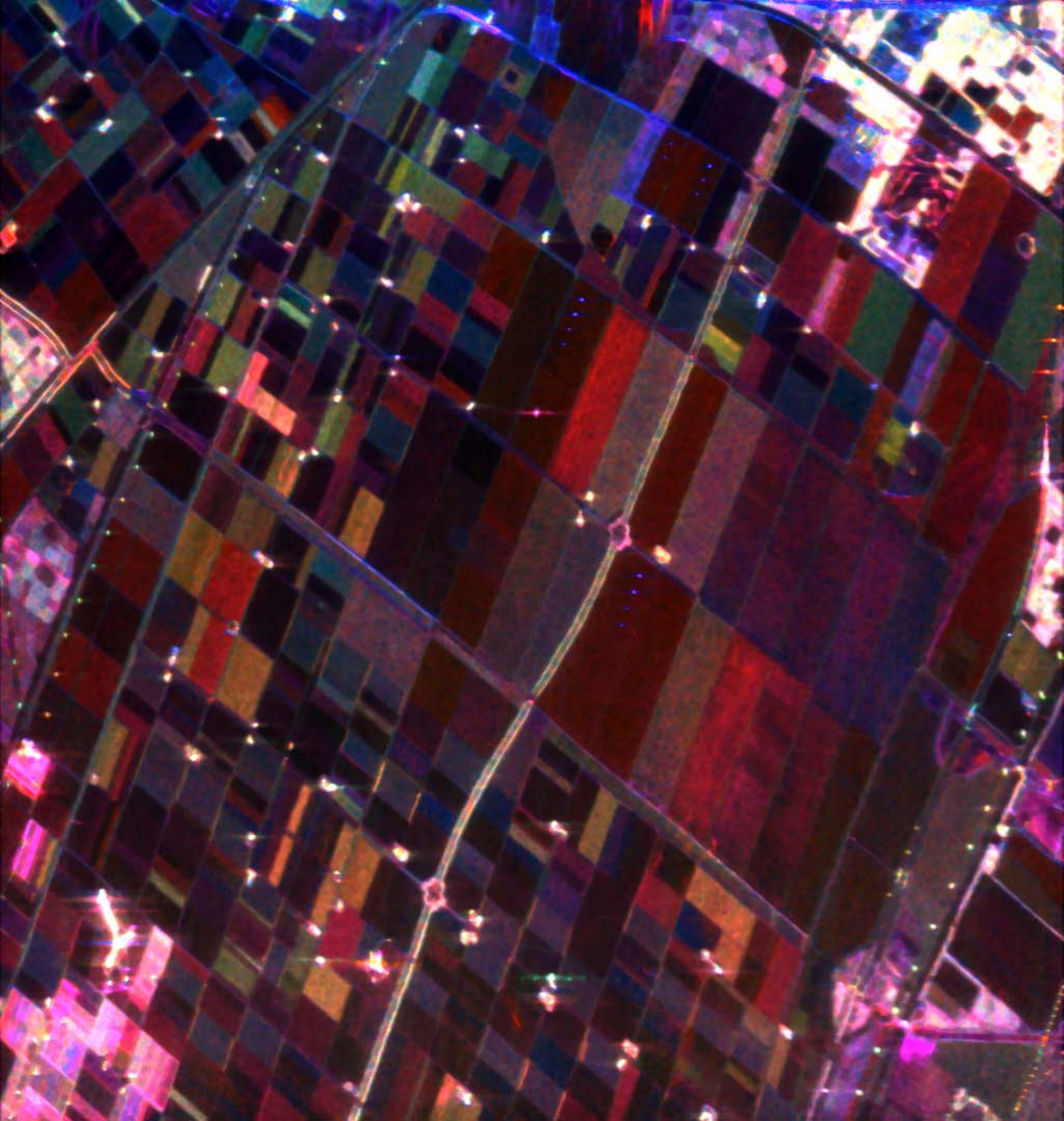}%
		\label{Pauli_Fle_P}}
	\hfil
	\subfloat[]{\includegraphics[width=1.3in,height=1.1in]{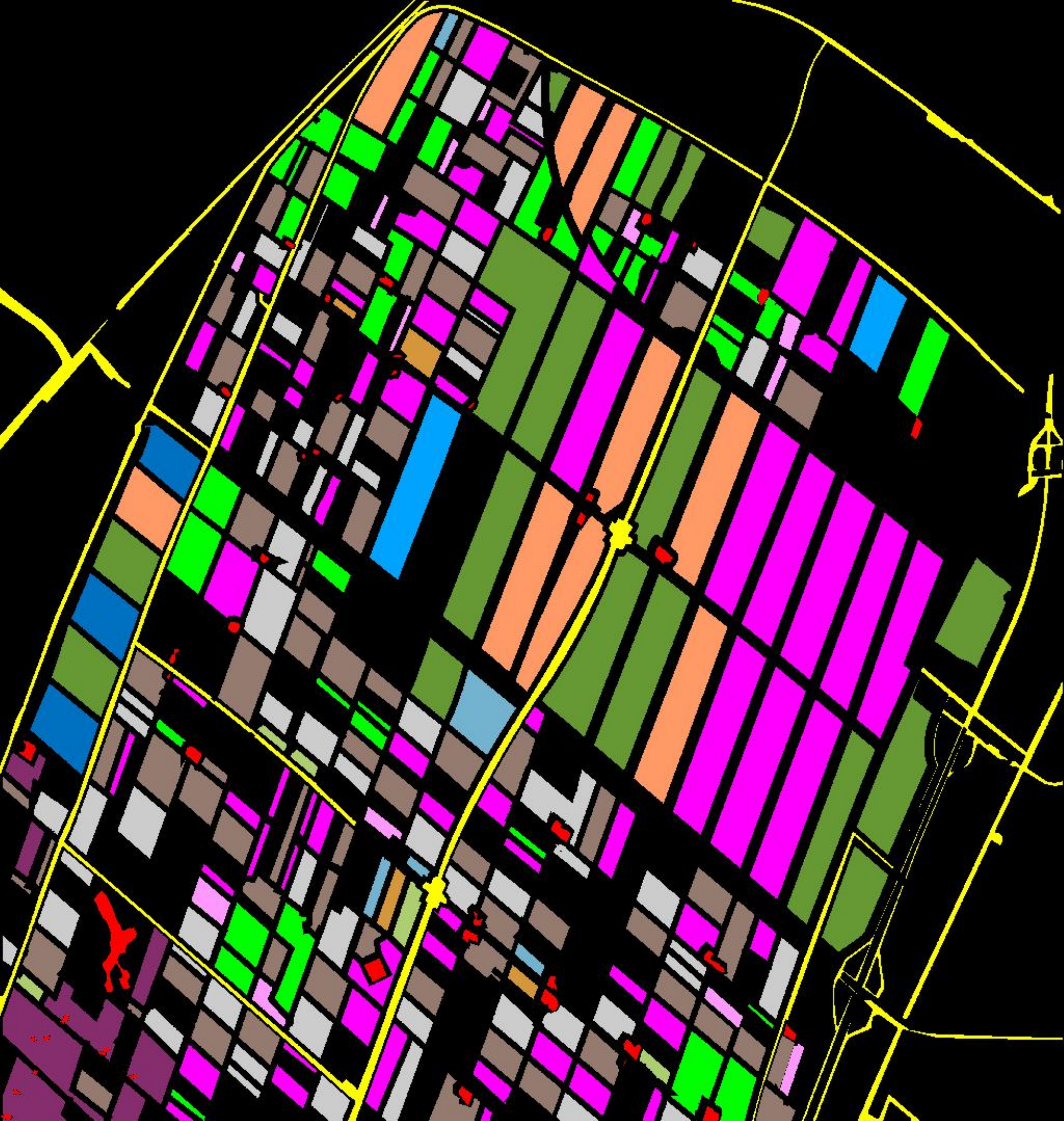}%
		\label{GroundTruth_Fle}}
	\hfil
	\subfloat{\includegraphics[width=2.8in]{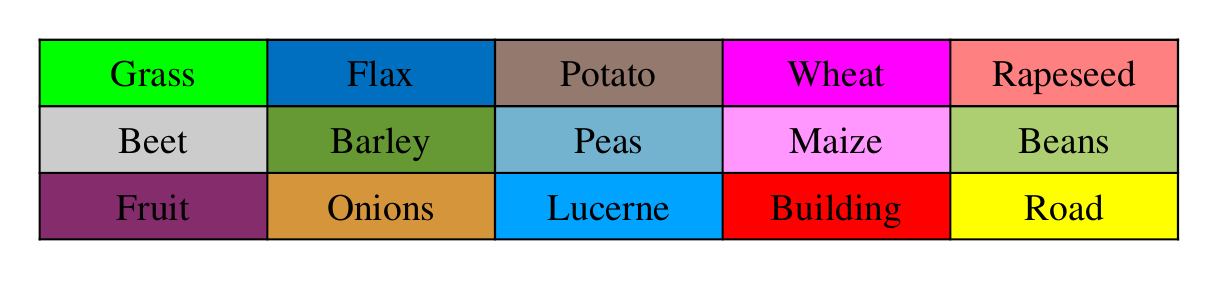}}%
	\caption{The Flevoland dataset. (a) C-band. (b) L-band. (c) P-band. (d) Ground Truth.}
	\label{Fig_Fle}
\end{figure}

\subsubsection{Flevoland Dataset}
The Flevoland dataset conclude two dual-frequency PolSAR datasets: Flevoland\_CL (C-band PolSAR data and L-band PolSAR data) and Flevoland\_CP (C-band PolSAR data and P-band PolSAR data). All data is obtained by the NASA/JPL AIRSAR system. In addition, the image size of this dataset is 1079$\times$1024, and the categories included in this dataset are: Grass, Flax, Potato, Wheat, Rapeseed, Beet, Barley, Peas, Maize, Beans, Fruit, Onions, Lucerne, Building and Road. Fig. \ref{Fig_Fle} shows the pauli RGB images of each band and the ground truth map plotted against Google Earth. In the following experiments, 1500 samples (100 samples for each category) are selected as training sample set.

\subsection{Ralated Methods for Comparison}
This article selects eight relevant classification algorithms for comparative experiments, namely:
(1) two branch CNN (tbCNN)\cite{2018_TGRS_tbCNN}, which used paralleled CNNs to learn remote sensing data of different modalities separately; 
(2) spatial–spectral cross-modal enhancement network (S2ENet)\cite{2022_GRSL_S2ENet}, which proposed a spatial-spectral enhancement module for cross-modal information interaction in both spatial and feature dimensions; 
(3) global-local Transformer (GLT)\cite{2022_TGRS_GLT}, which characterized local and global correlated features by CNN and ViT; 
(4) extended vision transformer (ExViT) \cite{2023_TGRS_ExViT}, which processed multi-modal remote sensing data with paralleled ViTs and performed multi-modal fusion through cross-modality attention; 
(5) multi-head cross-patch attention (mCrossPA) \cite{2023_TGRS_mCrossPA}, which utilized the "cls" of one source data to transfer complementary information to another source data; 
(6) separable dynamic grouping convolution (SepDGConv)\cite{2022_TGRS_SepDGConv}, which used a single-stream network instead of the commonly used dual-stream network for dual-modal remote sensing data fusion; 
(7) asymmetric feature fusion network (AsyFFNet)\cite{2023_TNNLS_AsyFFNet}, which sparsified multi-modal features through the scaling factor of BN layer; 
(8) cross-modal gated fusion network (CMGFNet)\cite{2022_ISPRS_CMGFNet}, which constructed a gated fusion module to combine two modalities and proposed a multi-level feature fusion strategy to fuse the high-level features and low-level features.

\begin{table*}[!t]
	\renewcommand{\arraystretch}{1.3} 
	\centering
	\caption{Ablation Studies on SDSR and DGSD.}
	\begin{tabular}{c|c|c|c|c}
		\hline
		\multirow{2}{*}{Dataset} &  baseline &  baseline+SDSR  & baseline+SDSR+cat & baseline+SDSR+DGSD \\
		\cline{2-5} 
		& \multicolumn{4}{c}{OA(\%)/AA(\%)/$\kappa \times$ 100}\\
		\hline
		\multirow{2}{*}{Hebei\_SL} &S|88.15/87.42/82.74 & S|90.30/91.85/85.62 & \multirow{2}{*}{92.77/92.81/88.99} & \multirow{2}{*}{95.76/95.04/93.51}\\
		&L|85.51/86.43/78.02 & L|89.79/89.52/83.46 & & \\
		\hline
		\multirow{2}{*}{SanFrancisco\_CL} &C|96.35/95.13/94.98 & C|96.86/96.21/95.69 & \multirow{2}{*}{98.76/98.46/98.29} & \multirow{2}{*}{99.08/98.64/98.73}\\
		&L|96.64/96.33/95.38 & L|97.09/96.55/96.01 & & \\
		\hline
		\multirow{2}{*}{Flevoland\_CL} & C|88.35/89.96/86.65 & C|90.22/91.20/88.79 & \multirow{2}{*}{96.16/96.45/95.59}& \multirow{2}{*}{97.66/97.57/97.31}\\
		& L|92.79/93.66/91.71 & L|93.10/94.06/92.07 & & \\
		\hline
		\multirow{2}{*}{Flevoland\_CP} & C|88.35/89.96/86.65 & C|90.22/91.20/88.79 & \multirow{2}{*}{94.76/95.49/93.99} & \multirow{2}{*}{96.35/96.69/95.80}\\
		& P|86.25/86.52/84.57 & P|87.12/87.02/85.28 & & \\
		\hline
	\end{tabular}%
	\label{Tab_ablation}%
\end{table*}%

\subsection{Experimental Settings}
All experiments are run on the Hewlett-Packard (HP)-Z840 Workstation with Nvidia GeForce RTX Ti 4060 GPU, 64-GB RAM, PyTorch environment, and Windows 10 operating system. The neural networks are trained with the optimizer as Adam, epoch as 200, and the basic learning rate as 1e-3 which is decayed by 0.9 every 50 epochs.

Moreover, four evaluation indicators are used to measure and analyze the classification results: the accuracies of each category, overall accuracy (OA), average accuracy (AA), and kappa coefficient ($\kappa$). The formulas for solving OA, AA, and $\kappa$ are:
\begin{align}\label{equ_4_1}
OA = \frac{\sum_{i=1}^{M} (F(i,i))}{N_{all}},
AA = \frac{1}{M}\sum_{i=1}^{M}\frac{F(i,i)}{F(i,:)},
\kappa = \frac{OA-p_e}{1-p_e}.
\end{align}
Where $p_e=\frac{1}{N_{all}^2}\sum_{i=1}^{M}F(i,:)*F(:,i)$, $F\in \mathbb{R}^{M\times M}$ is the confusion matrix and $N_{all}$ represents the number of samples. The higher these indicators are, the better the classification results are.

\subsection{Ablation Studies}
\subsubsection{Effect of the SDSR and DGSD}
To verify the effectiveness of the two major modules proposed in this article, we refer to the combination of early CNN, patch embedding and ViT as \textit{baseline}, and constructe \textit{baseline+SDSR}, \textit{baseline+SDSR+cat} and \textit{baseline+SDSR+DGSD}. The \textit{baseline} and \textit{baseline+SDSR} are designed for single-frequency data, the \textit{baseline+SDSR+cat} and \textit{baseline+SDSR+DGSD} are designed for dual-frequency data. "cat" implies the concatenation operation on dual-frequency data.

As shown in Table \ref{Tab_ablation}, adding the SDSR module to the \textit{baseline} has improved the accuracy on all datasets. This is because the SDSR removes noisy pixels and reconstructs a relatively pure sample, so that the classification information learning of CNN and ViT only occurs among informative pixels, avoiding the involvement of noisy pixels in information interaction and affecting classification performance. 

The difference between \textit{baseline+SDSR} and \textit{baseline+SDSR+cat} is that one is a single branch network for single-frequency data, the other is a dual branch network for two frequency data and a feature stacking operation is added to achieve dual-frequency data fusion. In Table \ref{Tab_ablation}, the classification performance of \textit{baseline+SDSR+cat} is significantly better than that of \textit{baseline+SDSR}. This is because dual-frequency data provides richer classification information, allowing the network to learn more comprehensively. However, simple feature stacking can also lead to issues such as feature redundancy and inter-band competition. Therefore, we propose the DGSD strategy for dual-frequency data fusion. It selects the dominant single-frequency branch on each sample through gating operation, and guides the dual-frequency student model based on the advantageous branch, thereby achieving reasonable utilization of dual-frequency complementary on different terrain objects. The experimental results in Table \ref{Tab_ablation} have shown that using the DGSD strategy for joint classification (\textit{baseline+SDSR+DGSD}) has higher accuracy than simple stacking strategy (\textit{baseline+SDSR+cat}).

\subsubsection{Analysis of the Balance Factor $\alpha$}

\begin{figure}[!t]
	\centering
	\subfloat[]{\includegraphics[width=1.5in]{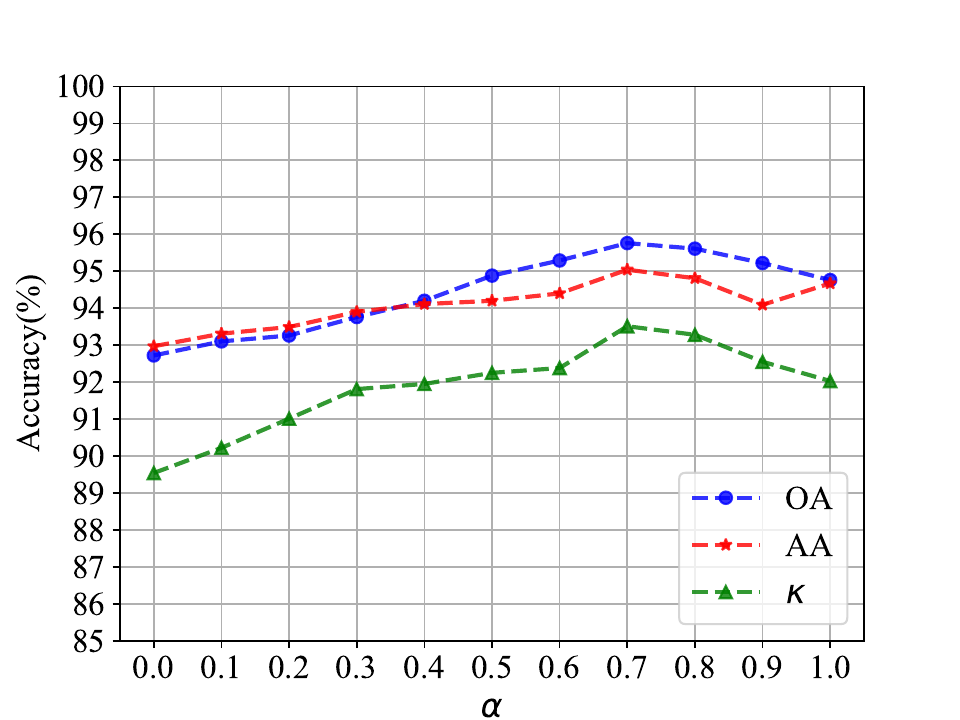}%
		\label{alpha_HB_SL}}
	\hfil
	\subfloat[]{\includegraphics[width=1.5in]{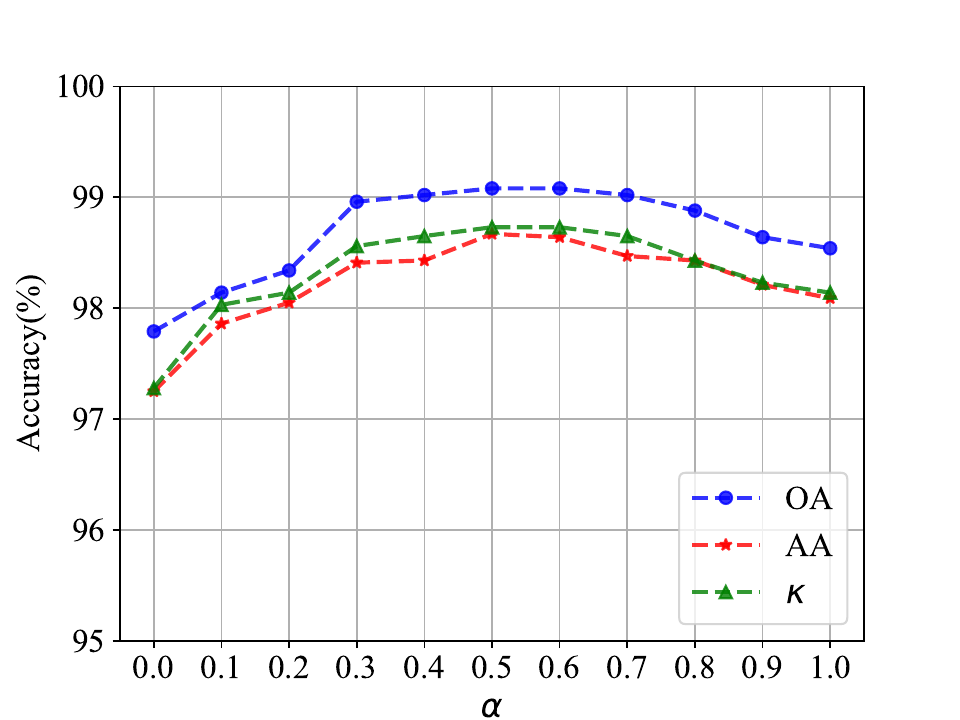}}%
	\label{alpha_San_CL}
	\hfil
	\subfloat[]{\includegraphics[width=1.5in]{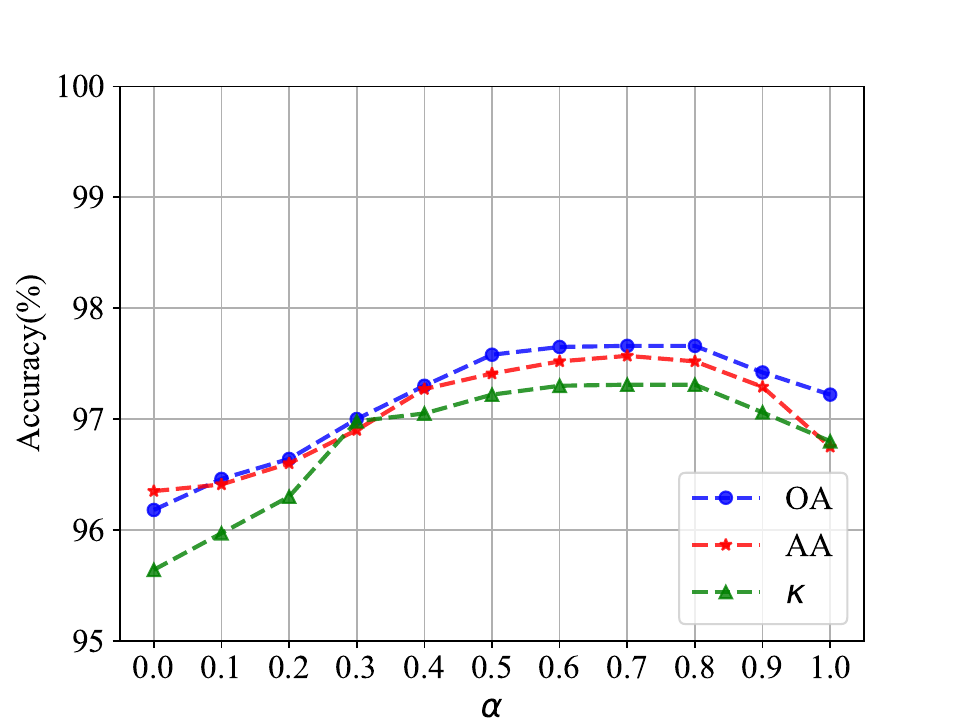}%
		\label{alpha_Fle_CL}}
	\hfil
	\subfloat[]{\includegraphics[width=1.5in]{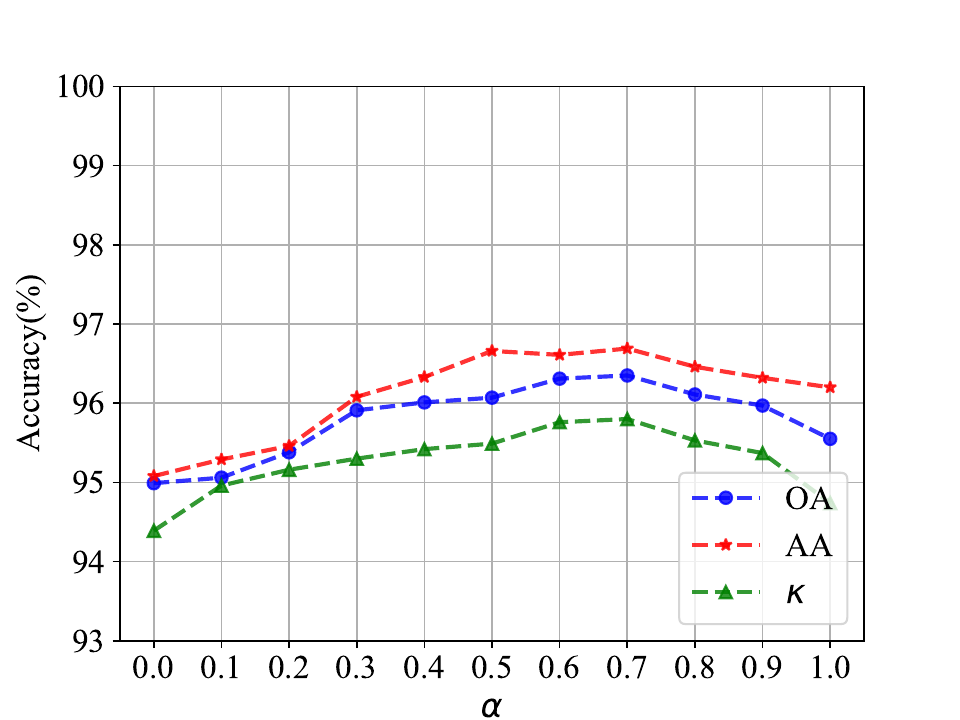}%
		\label{alpha_Fle_CP}}
	\hfil
	\caption{The evaluation of the balance factor $\alpha$. (a) Hebei\_SL. (b) SanFrancisco\_CL.  (c) Flevoland\_CL. (d) Flevoland\_CP.}
	\label{Fig_alpha}
\end{figure}

In order to find the optimal balance factor $\alpha$, we conducted experiments on $\alpha=\{0,0.1, 0.2, 0.3, 0.4, 0.5, 0.6, 0.7, 0.8, 0.9, 1\}$, and the experimental results are shown in Fig. \ref{Fig_alpha}. $\alpha=0$ means that the two single-frequency teacher models do not participate in the training process, and the student model only relies on the hard labels; $\alpha=1$ means that the dual-frequency student model is trained only by teacher models. In Fig. \ref{Fig_alpha}, it can be observed that the classification accuracy using only KD is higher than the classification accuracy using only hard labels. This is because the teacher model not only preserves the difference information among categories, but also contains the similarity information among categories. Therefore, the soft labels of the teacher model bring more inductive reasoning knowledge than the hard labels. But if we only use the soft labels of teacher model for training, the dual-frequency student branch will immerse itself in learning the knowledge of the dominant teacher branch. Although in this case, the accuracy of the student model will approach that of the dominant teacher branch, it will lose the utilization of features contained in another frequency data. Therefore, the proposed DGSD strategy combines the loss functions of $L_{KL}$ and $L_{CE}$, with a view to realizing the full utilization of the soft labels obtained by the teacher model and the rich information provided by the dual-frequency data. From Fig. \ref{Fig_alpha}, we find that the accuracy of Hebei\_SL reaches a peak at $\alpha=0.7$. Similarly, when $\alpha=0.4\sim0.6$, SanFracncisco\_CL has the highest accuracy; when $\alpha=0.6\sim0.8$, Flevoland\_CL has the highest accuracy; and when $\alpha=0.5\sim0.7$, Flevoland\_CP has the highest accuracy.

In the following experiments, we select $\alpha=0.7$ for Hebei\_SL, Flevoland\_CL, and Flevoland\_CP, $\alpha=0.5$ for SanFracncisco\_CL.

\begin{table*}[!t]
	\renewcommand{\arraystretch}{1.2} 
	\centering
	\caption{Classification Accuracy of different algorithms on Hebei\_SL.}
	\begin{tabular}{ccccccccc|ccc}  
		\hline
		\hline
		\multirow{3}{*}{Class} &  \multicolumn{11}{c}{Performance} \\
		\cline{2-12} 
		& \multirow{2}{*}{tbCNN} & \multirow{2}{*}{S2ENet} & \multirow{2}{*}{GLT} & \multirow{2}{*}{ExViT} & \multirow{2}{*}{mCrossPA} & \multirow{2}{*}{SepDGConv} & \multirow{2}{*}{AsyFFNet} &  \multirow{2}{*}{CMGFNet} & \multicolumn{3}{c}{Ours} \\
		\cline{10-12} 
		& & & &  &  & & & & S & L & S+L\\
		\hline
		Forest&	94.25&	97.02&	\textbf{97.18}&\textbf{	97.18}&	95.83&	96.16	&96.27	&96.18  &
		\underline{96.39}	& 96.06&	96.56\\
		
		Farmland&91.44&	90.87&	90.16&	91.76&	87.88&93.18	& 90.74 &92.93&	
		85.42 & \underline{90.20}& \textbf{95.25}\\
		
		Road&56.58&	82.39&	89.98&	88.90&	89.77&89.59	 & 88.33 &91.17&
		\underline{89.64}	&83.31 & \textbf{91.37}\\
		
		Building&85.64&	92.63& 92.64&	92.29&	92.00&94.50	& 93.92 &93.09&	
		89.59 & \underline{91.08}& \textbf{94.96}\\
		
		Bare land&95.47	& 97.77&	95.63&	\textbf{98.02}&	98.01&96.57	& 97.56 &97.82&
		\underline{96.71}	 &93.43 & 97.74 \\
		\hline
		OA(\%)&90.50	&	93.03&	93.58&	93.98&	93.88&93.78	& 94.14 &94.48&	
		\underline{90.30} & 89.79& \textbf{95.76}\\
		AA(\%)	&84.61&	92.14&	93.12&	93.63&	93.70&93.80&	94.16 &94.64&	
		\underline{91.85} & 89.52& \textbf{95.04}\\
		$\kappa\times$100&	85.50& 89.47&	88.87&	90.90&	90.86&90.59& 92.22 &92.18&
		\underline{85.62}	 & 83.46& \textbf{93.51}	\\
		\hline
		\hline
	\end{tabular}%
	\label{Tab_HB_SL}%
\end{table*}%

\begin{figure*}[!t]
	\centering
	\subfloat[]{\includegraphics[width=1.15in,height=1.15in]{Fig/2_Pauli/GroundTruth_HB}
		\label{fig_HB_SL_1}}
	\hfil
	\subfloat[]{\includegraphics[width=1.15in,height=1.15in]{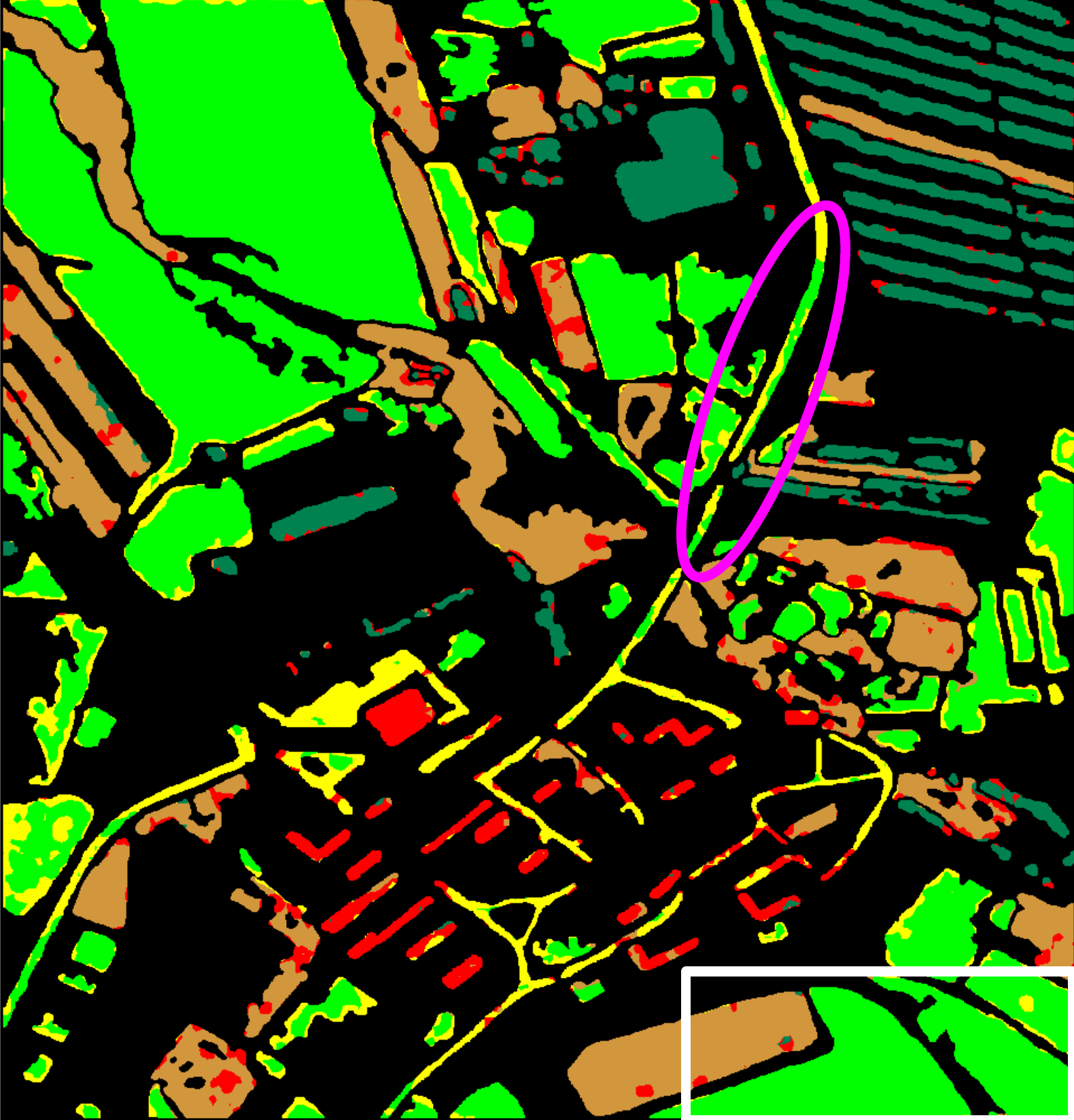}%
		\label{fig_HB_SL_2}}
	\hfil
	\subfloat[]{\includegraphics[width=1.15in,height=1.15in]{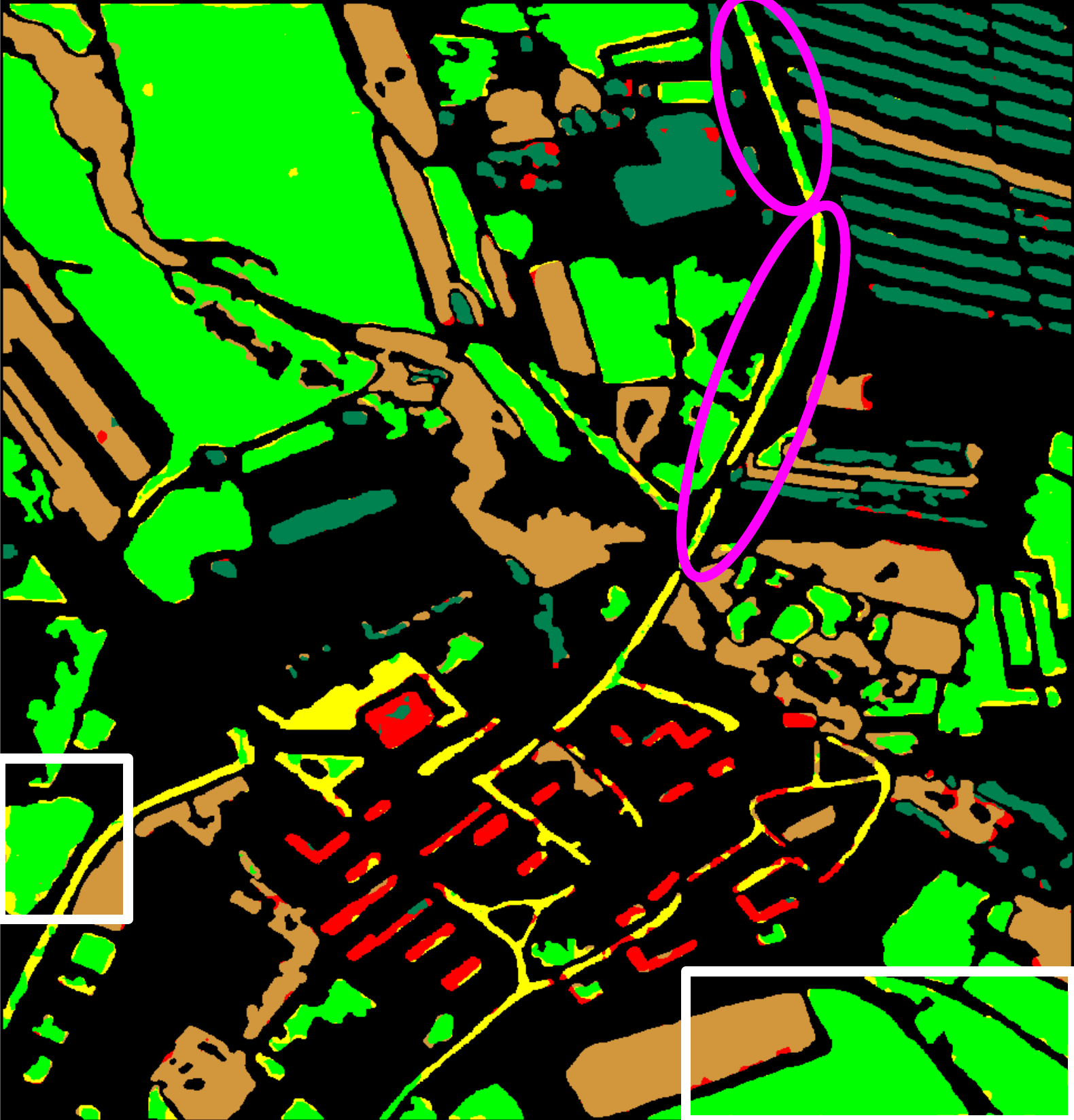}%
		\label{fig_HB_SL_3}}
	\hfil
	\subfloat[]{\includegraphics[width=1.15in,height=1.15in]{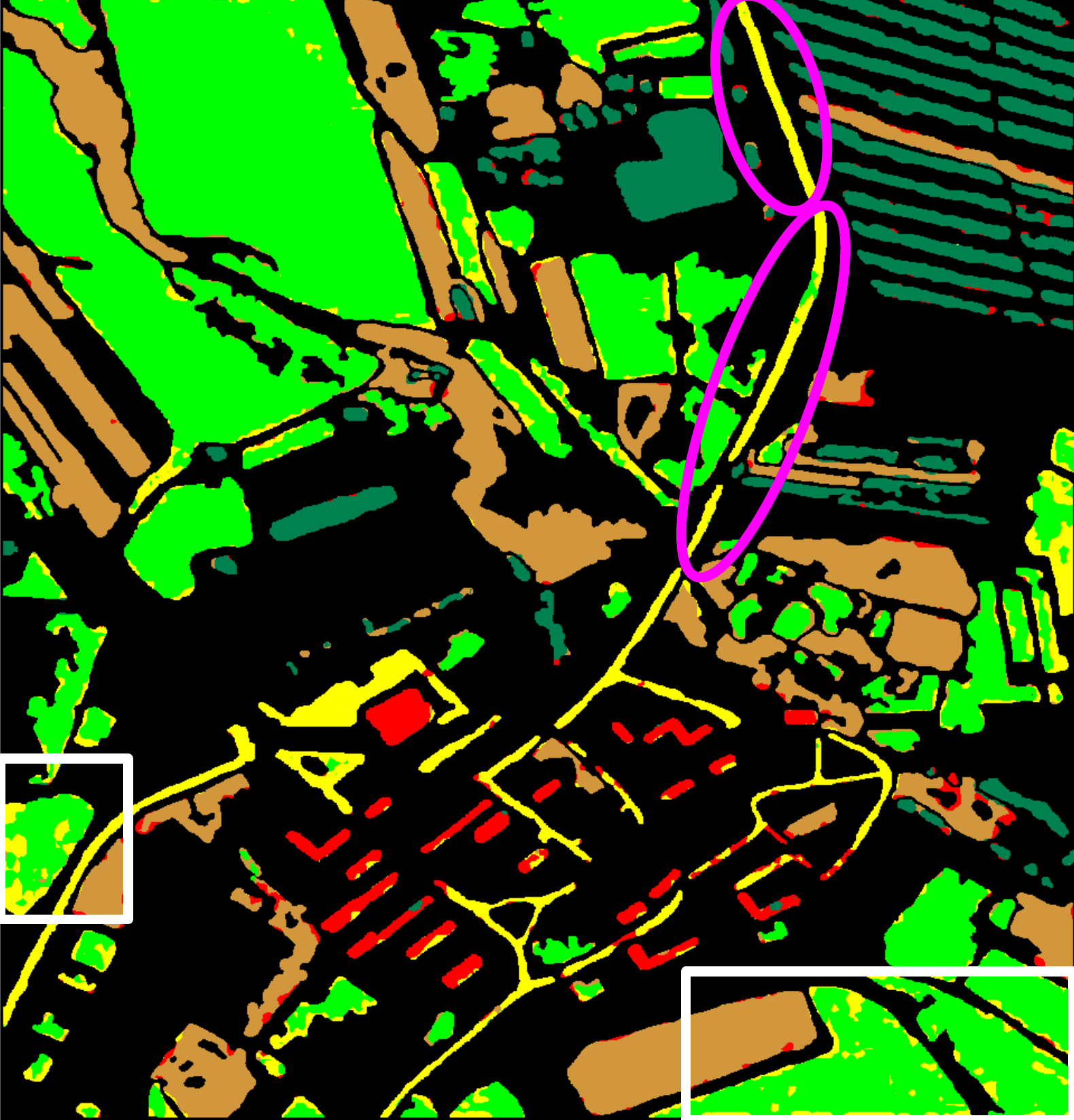}%
		\label{fig_HB_SL_4}}
	\hfil
	\subfloat[]{\includegraphics[width=1.15in,height=1.15in]{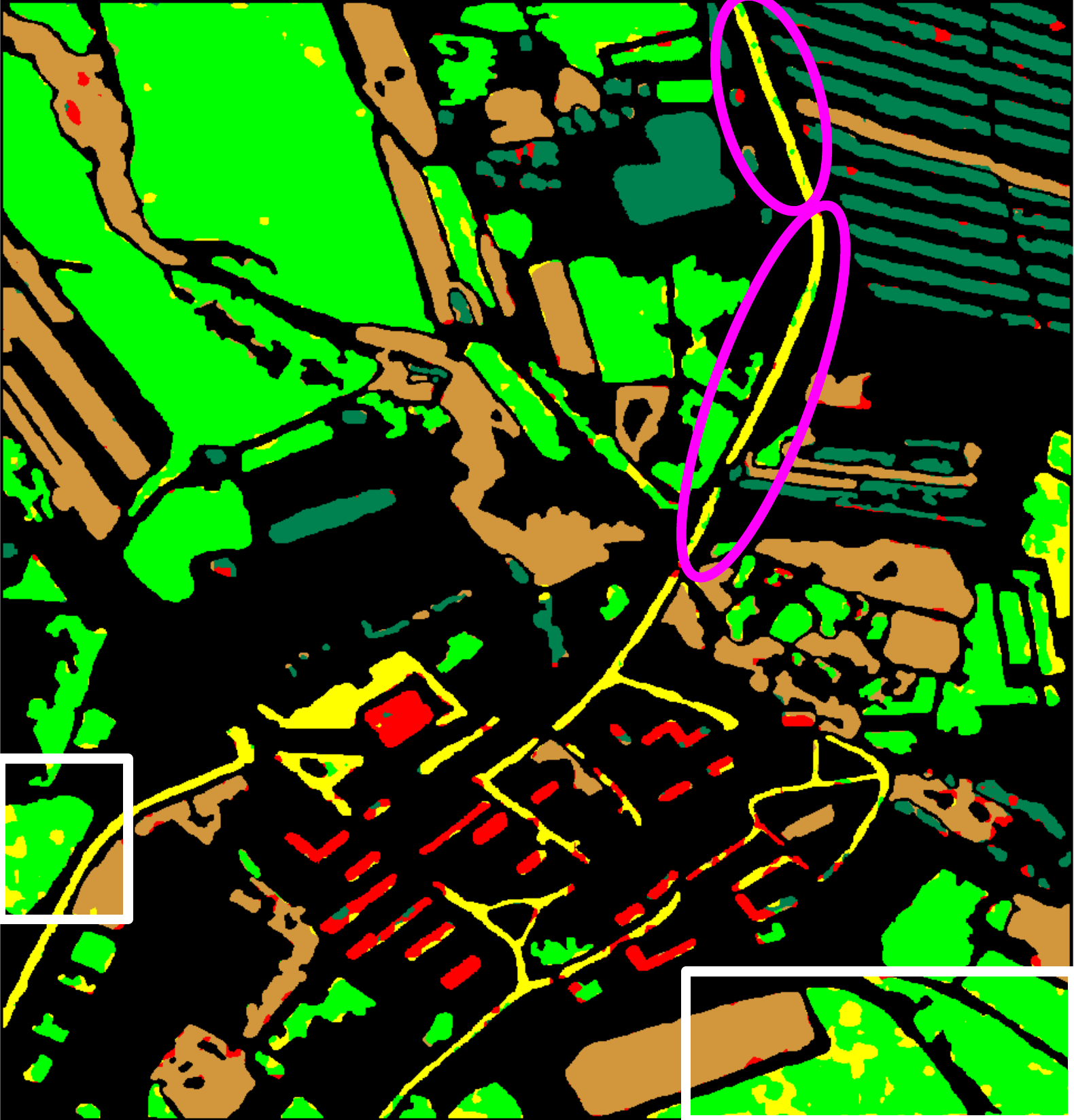}%
		\label{fig_HB_SL_5}}
	\hfil
	\subfloat[]{\includegraphics[width=1.15in,height=1.15in]{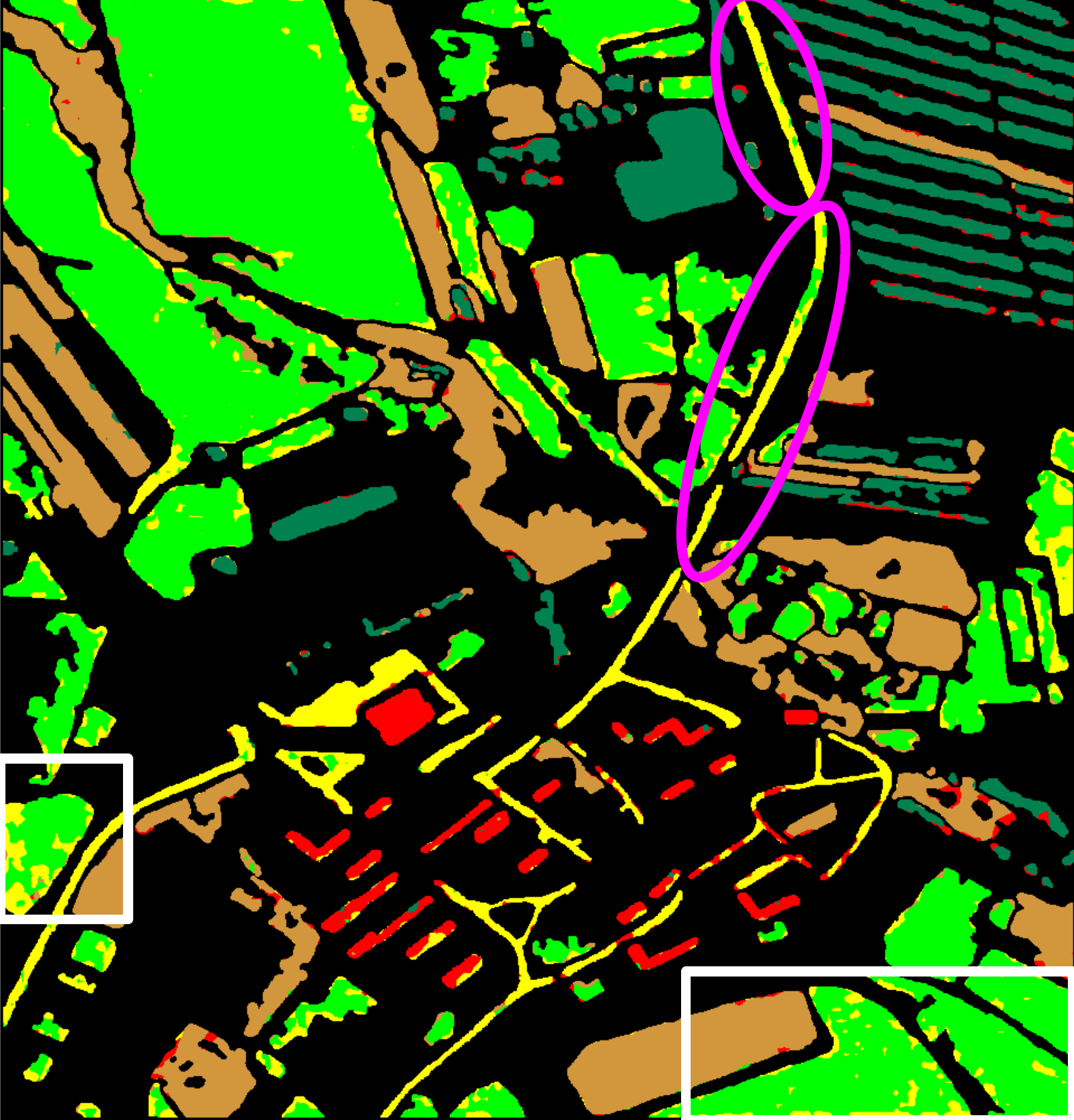}%
		\label{fig_HB_SL_6}}
	\hfil
	\subfloat[]{\includegraphics[width=1.15in,height=1.15in]{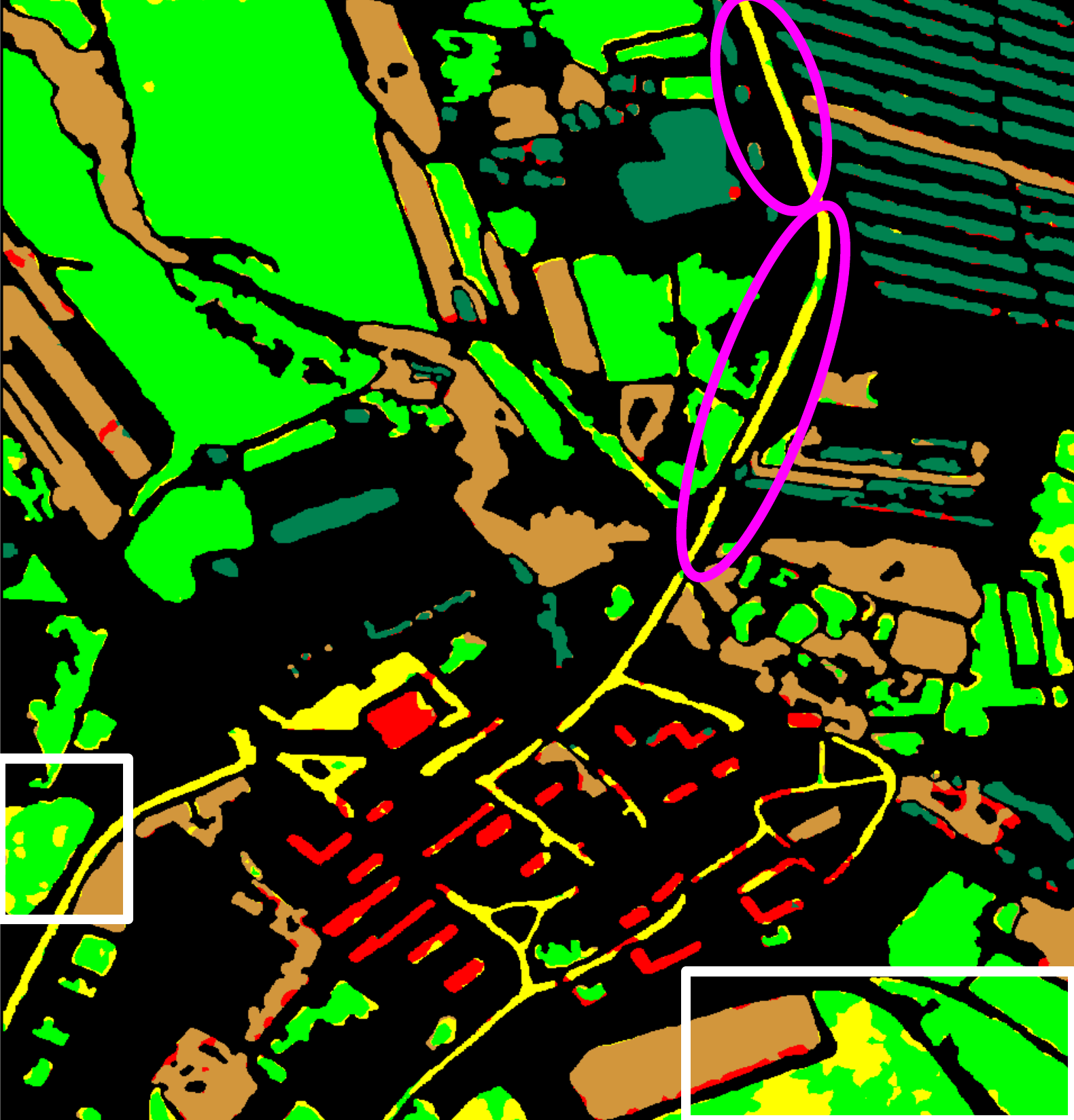}%
		\label{fig_HB_SL_7}}
	\hfil
	\subfloat[]{\includegraphics[width=1.15in,height=1.15in]{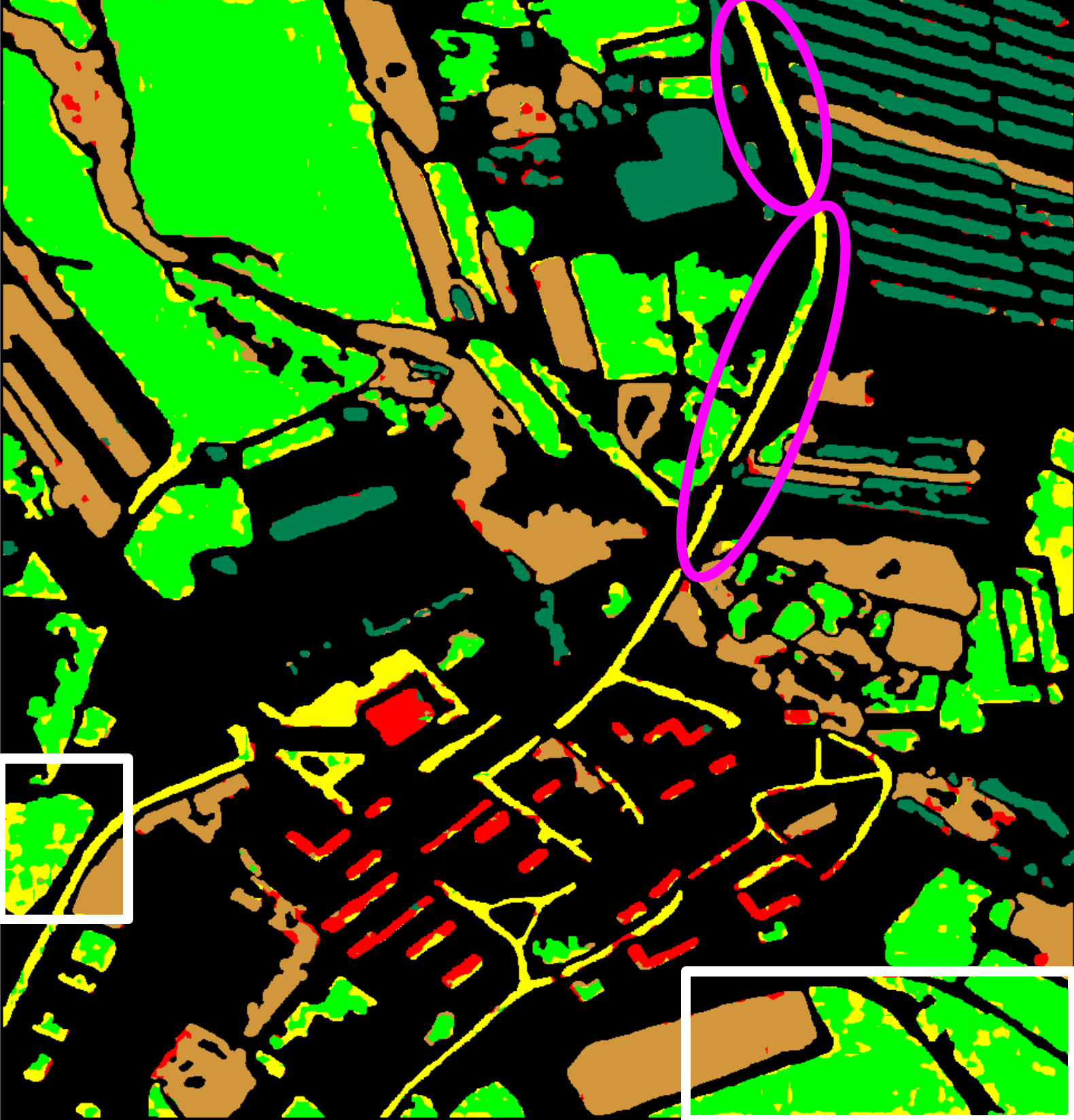}%
		\label{fig_HB_SL_8}}
	\hfil
	\subfloat[]{\includegraphics[width=1.15in,height=1.15in]{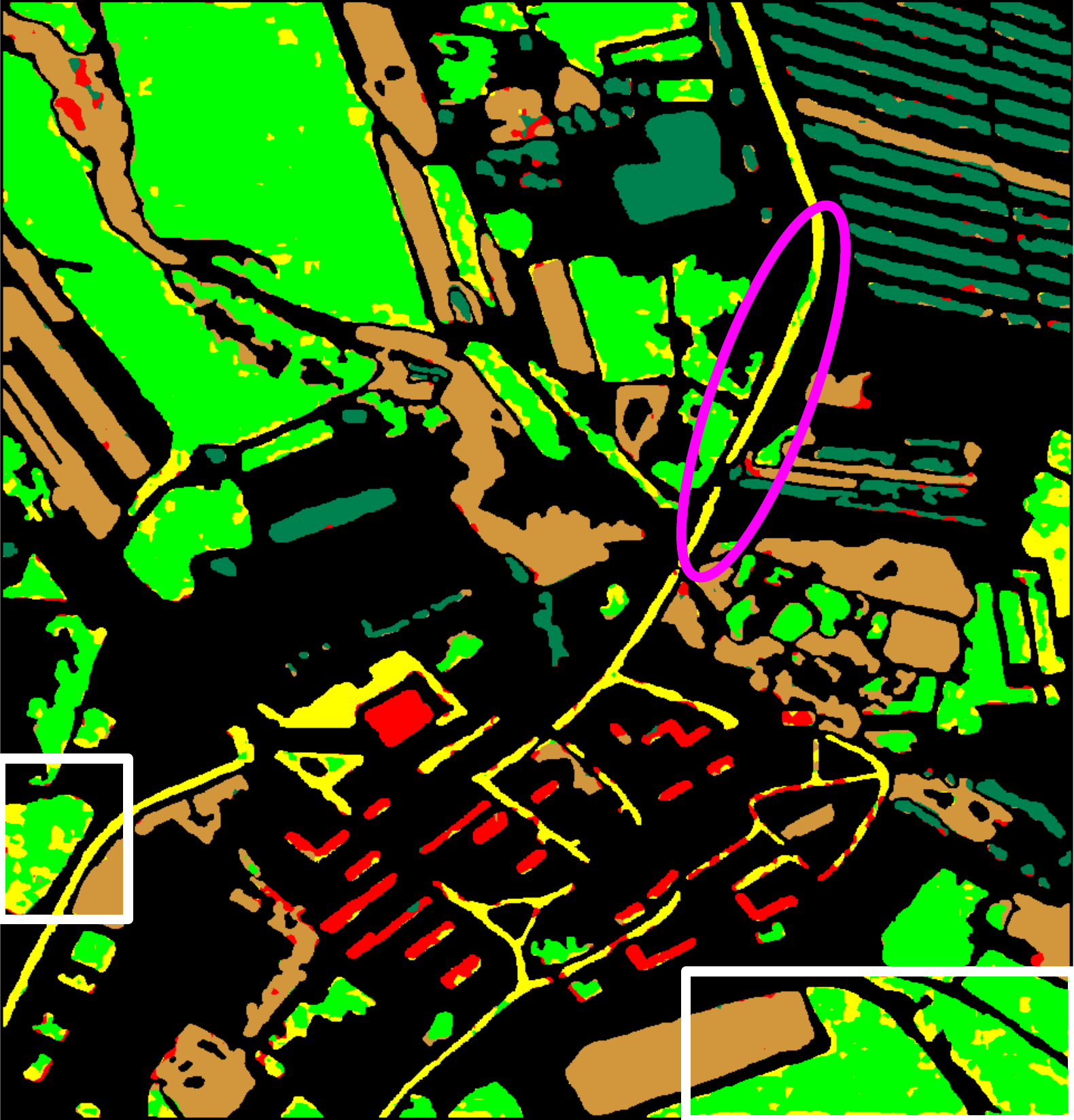}%
		\label{fig_HB_SL_9}}
	\hfil
	\subfloat[]{\includegraphics[width=1.15in,height=1.15in]{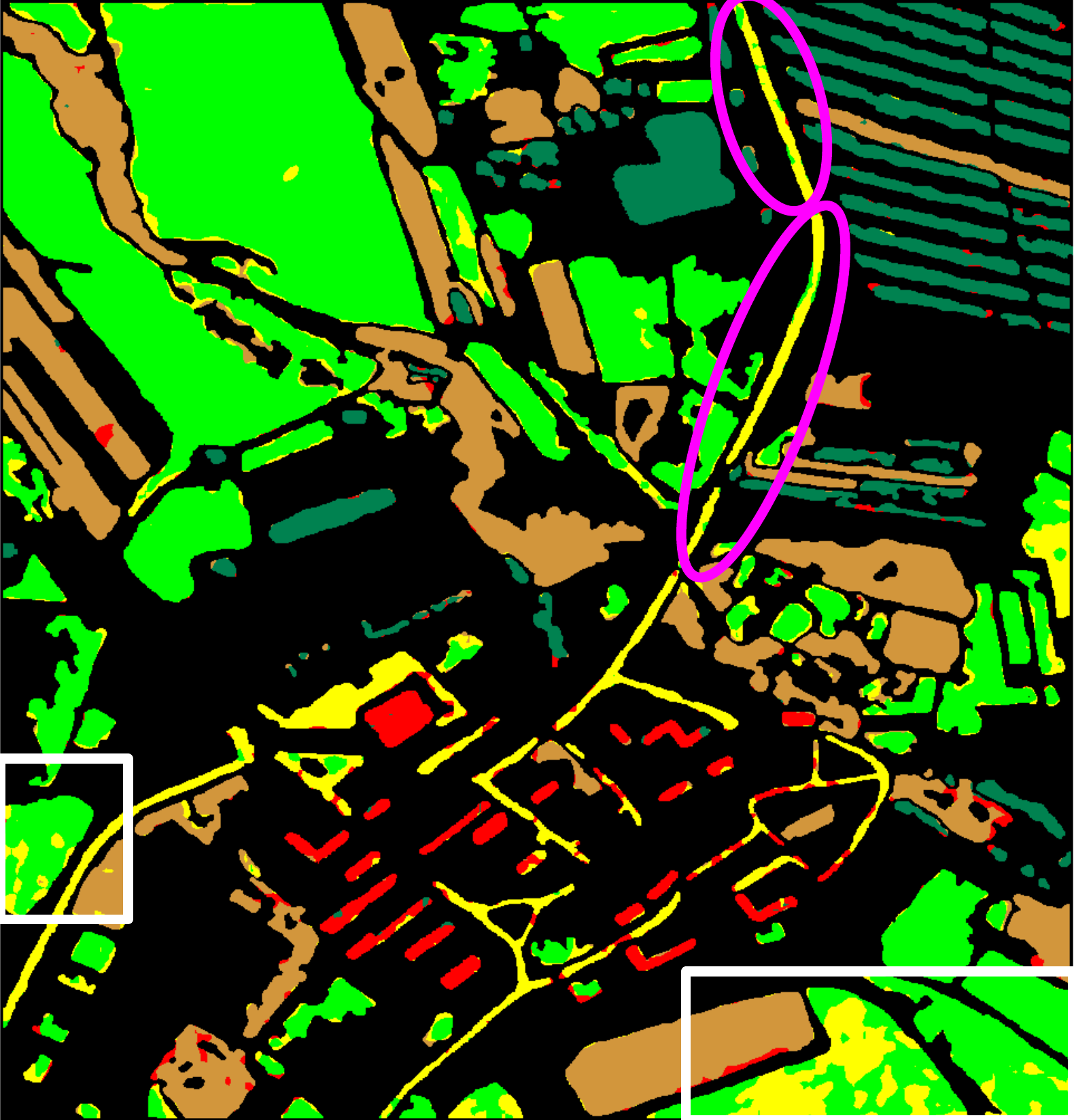}
		\label{fig_HB_SL_S}}
	\hfil
	\subfloat[]{\includegraphics[width=1.15in,height=1.15in]{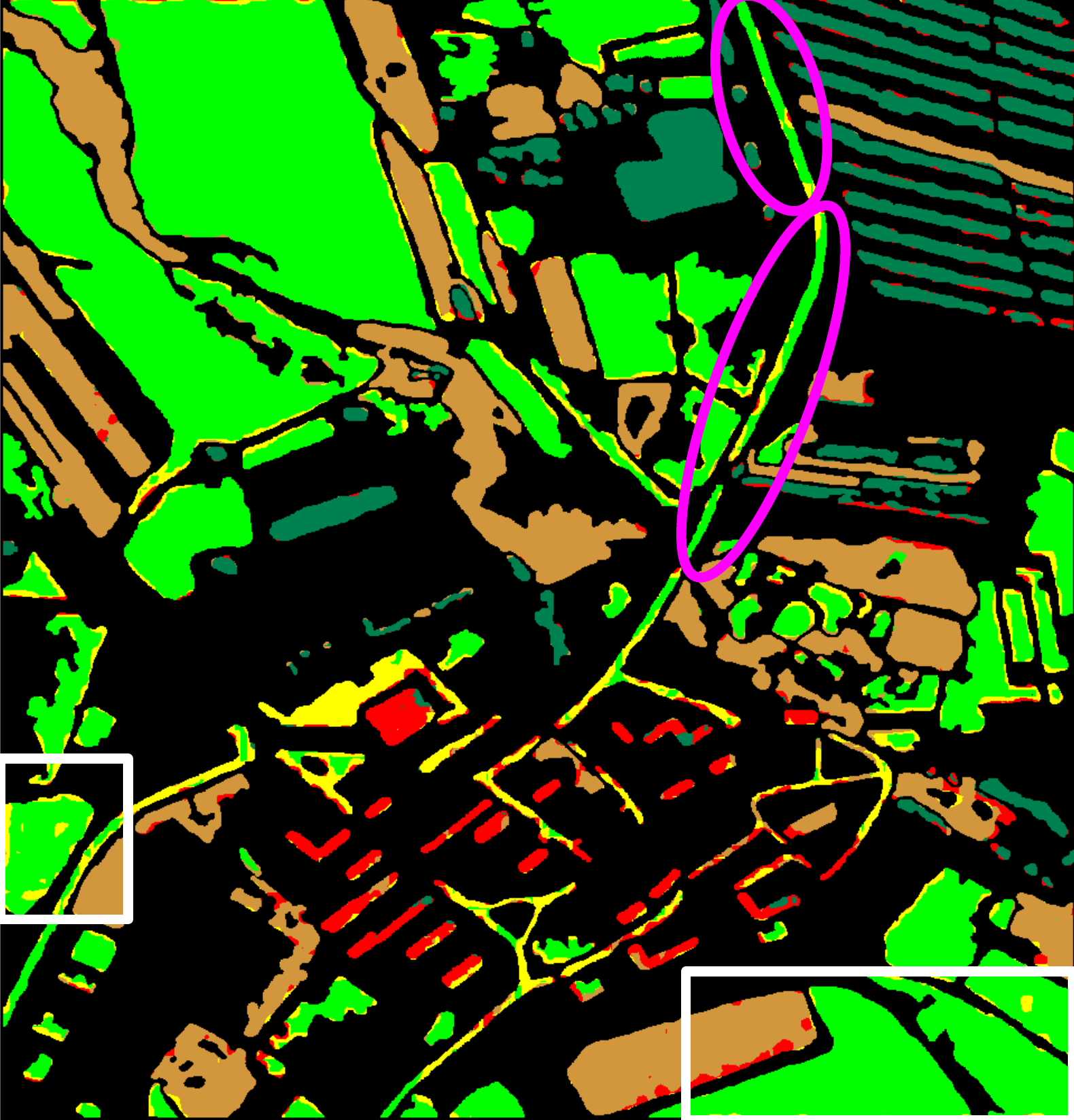}
		\label{fig_HB_SL_L}}
	\hfil
	\subfloat[]{\includegraphics[width=1.15in,height=1.15in]{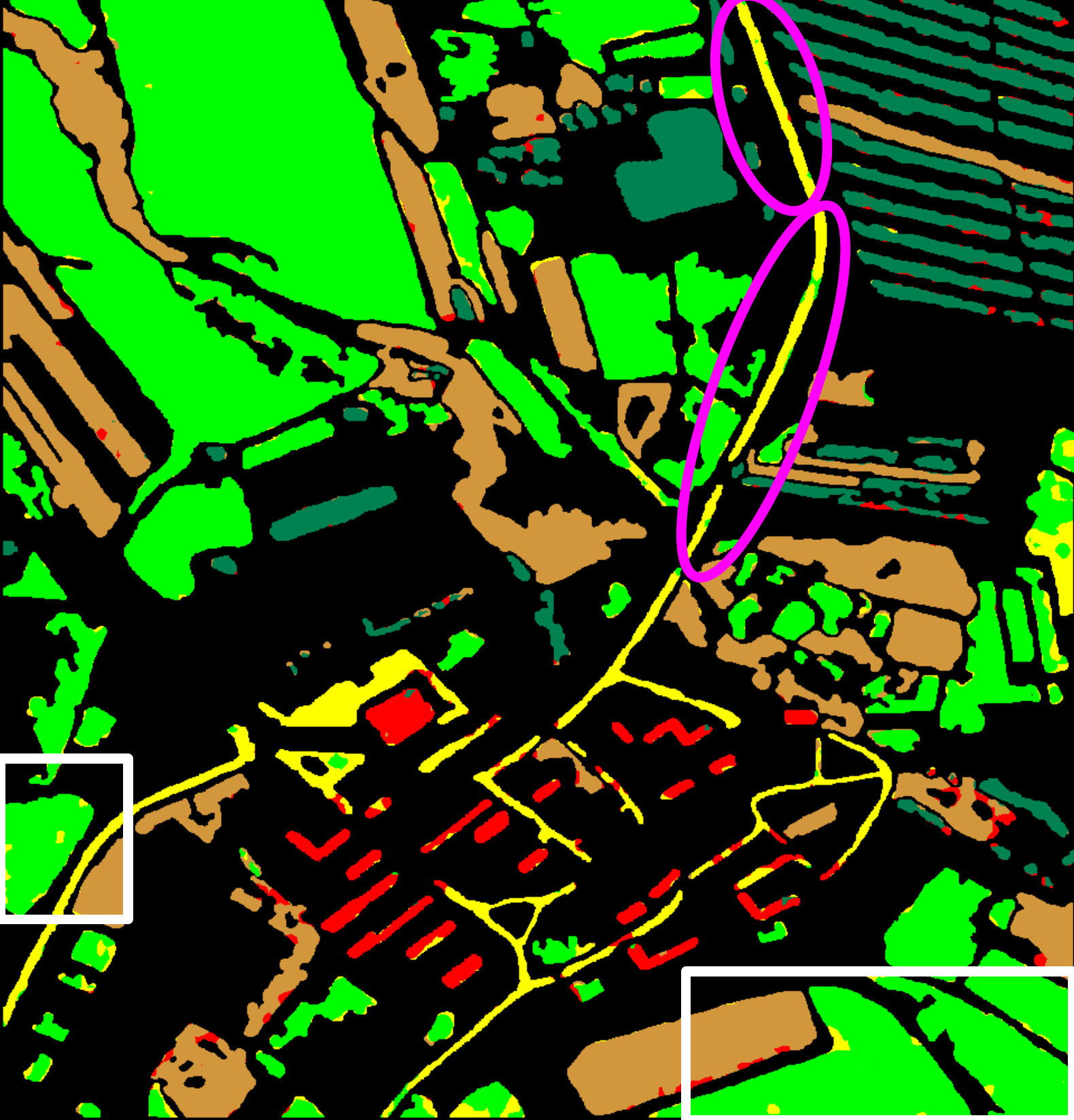}%
		\label{fig_HB_SL_proposed}}
	\caption{Classification maps obtained by different algorithms on Hebei\_SL. (a) Ground Truth. (b) tbCNN. (c) S2ENet.(d) GLT.(e) ExViT.(f) mCrossPA.(g) SepDGConv.(h) AsyFFNet. (i) CMGFNet. (j) Ours(S). (k) Ours(L). (l) Ours(S+L).}
	\label{Fig_HB_SL}
\end{figure*}

\subsection{Performance Comparison}
\begin{figure*}[!t]
	\centering
	\subfloat[]{\includegraphics[width=1.15in,height=1.15in]{Fig/2_Pauli/GroundTruth_San}
		\label{fig_San_CL_1}}
	\hfil
	\subfloat[]{\includegraphics[width=1.15in,height=1.15in]{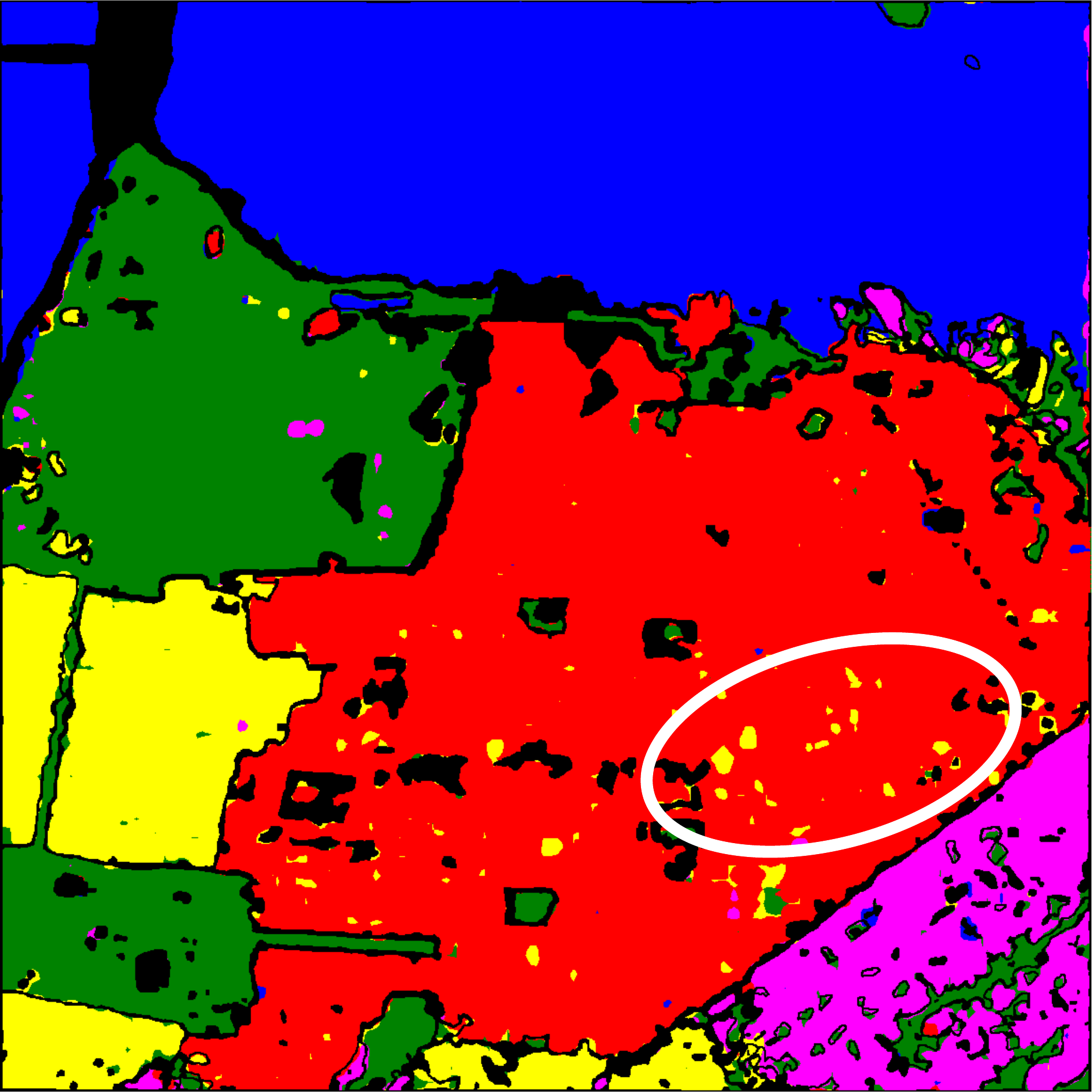}%
		\label{fig_San_CL_2}}
	\hfil
	\subfloat[]{\includegraphics[width=1.15in,height=1.15in]{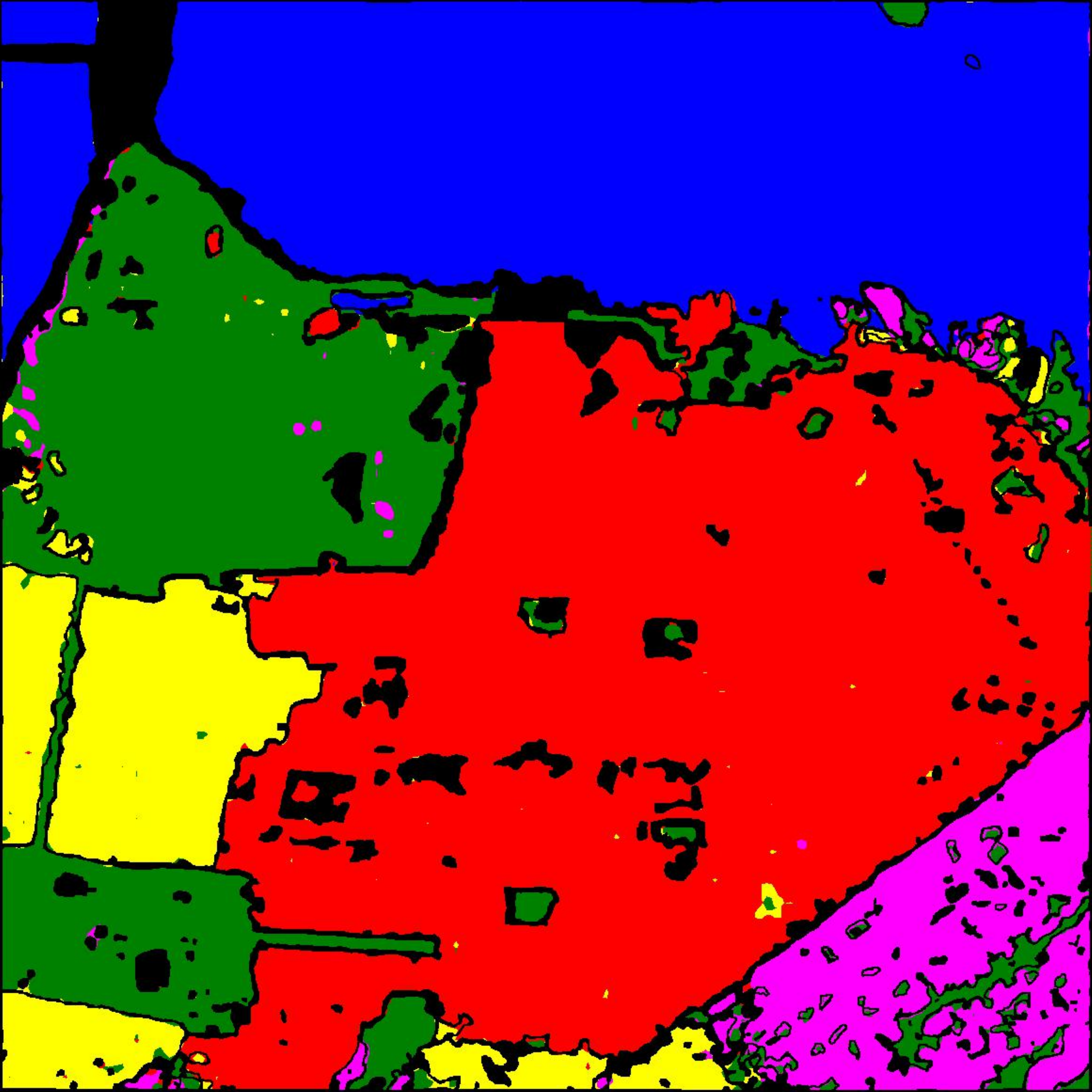}%
		\label{fig_San_CL_3}}
	\hfil
	\subfloat[]{\includegraphics[width=1.15in,height=1.15in]{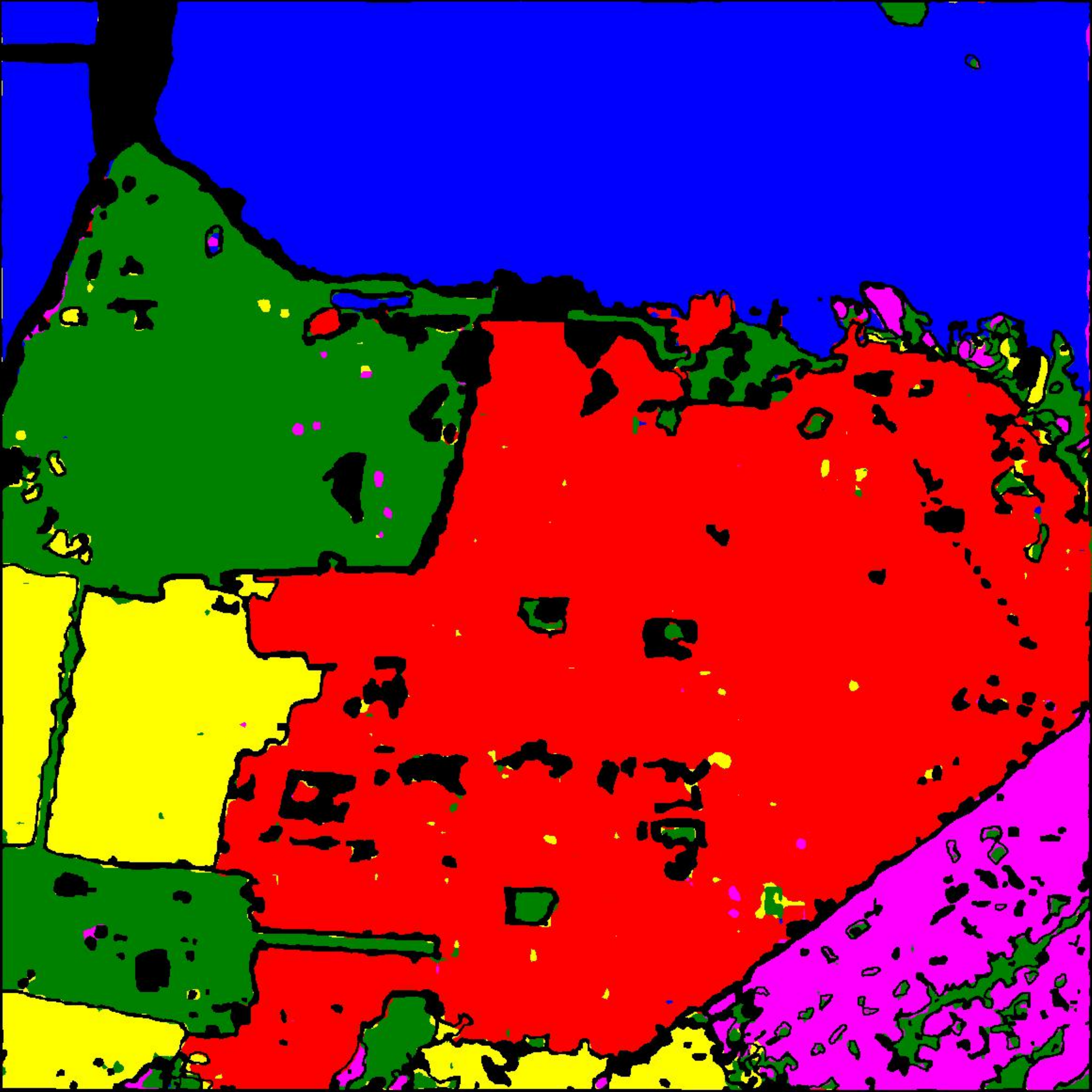}%
		\label{fig_San_CL_4}}
	\hfil
	\subfloat[]{\includegraphics[width=1.15in,height=1.15in]{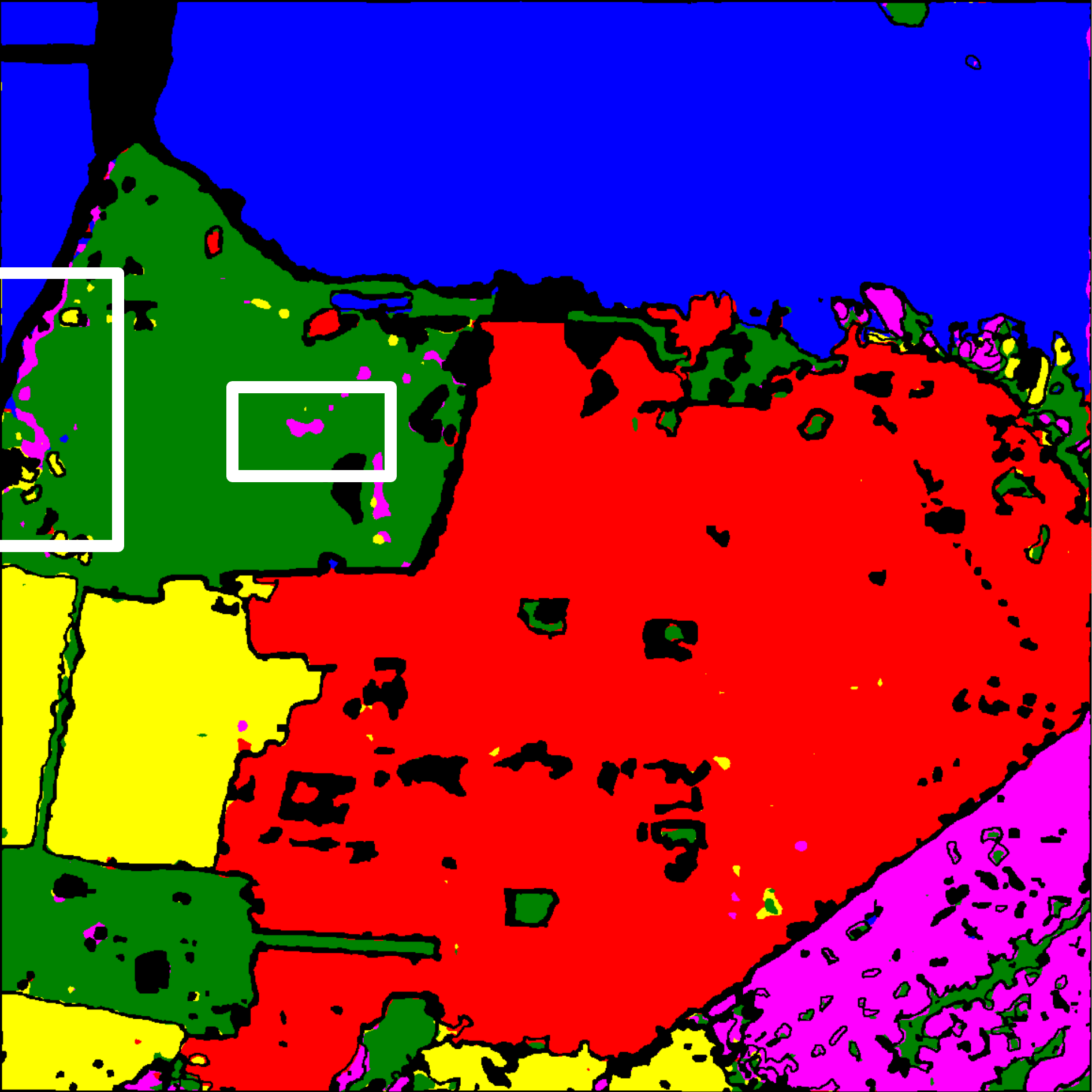}%
		\label{fig_San_CL_5}}
	\hfil
	\subfloat[]{\includegraphics[width=1.15in,height=1.15in]{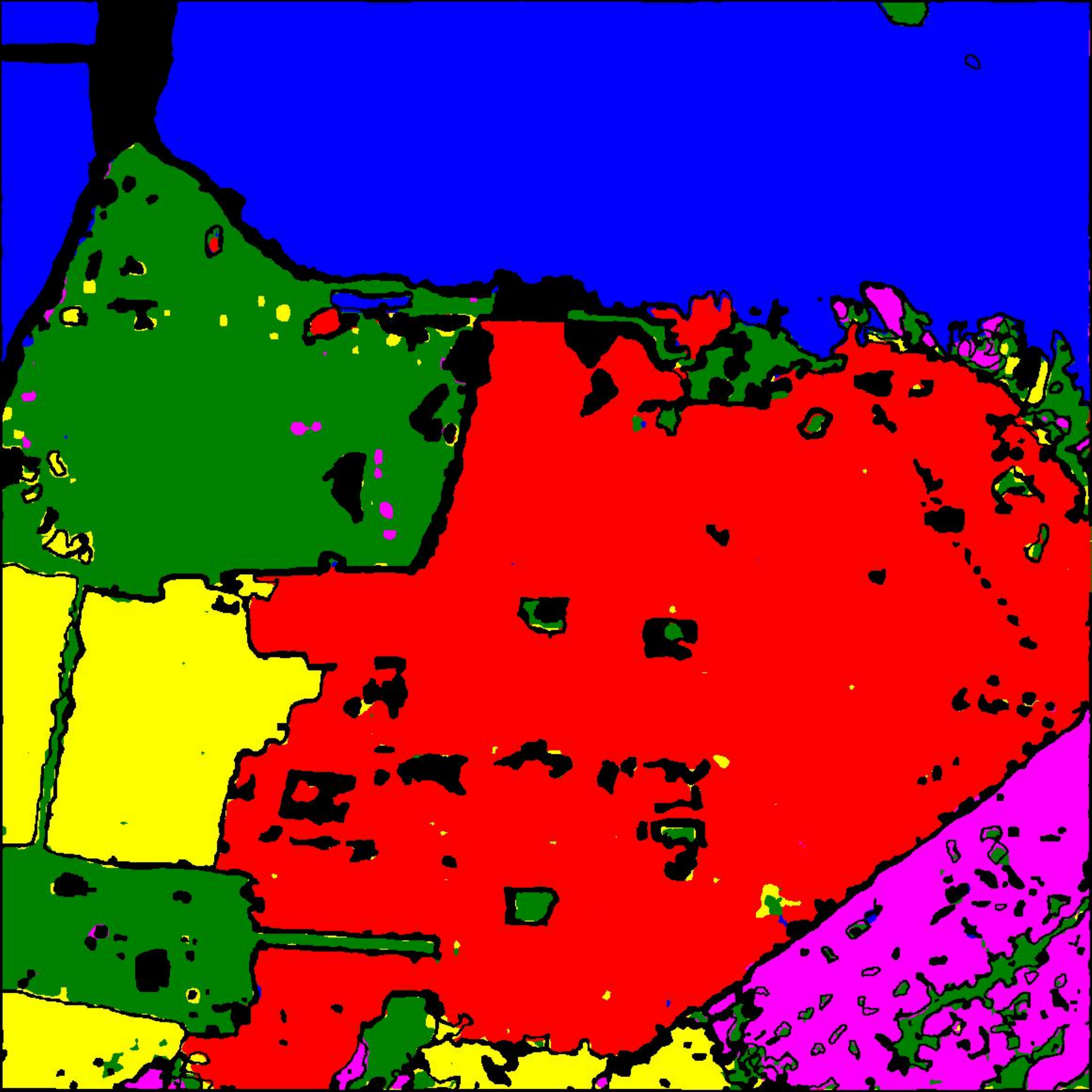}%
		\label{fig_San_CL_6}}
	\hfil
	\subfloat[]{\includegraphics[width=1.15in,height=1.15in]{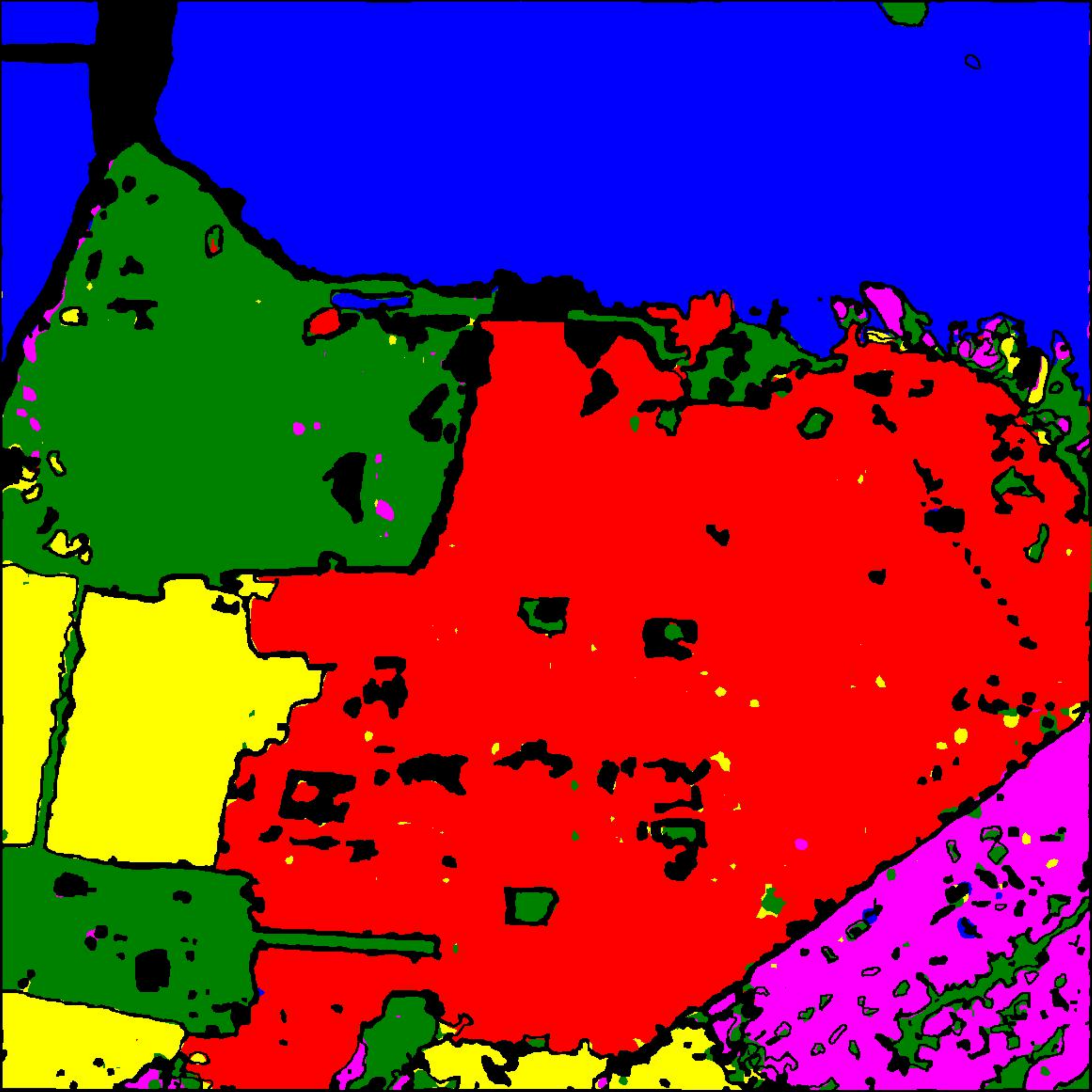}%
		\label{fig_San_CL_7}}
	\hfil
	\subfloat[]{\includegraphics[width=1.15in,height=1.15in]{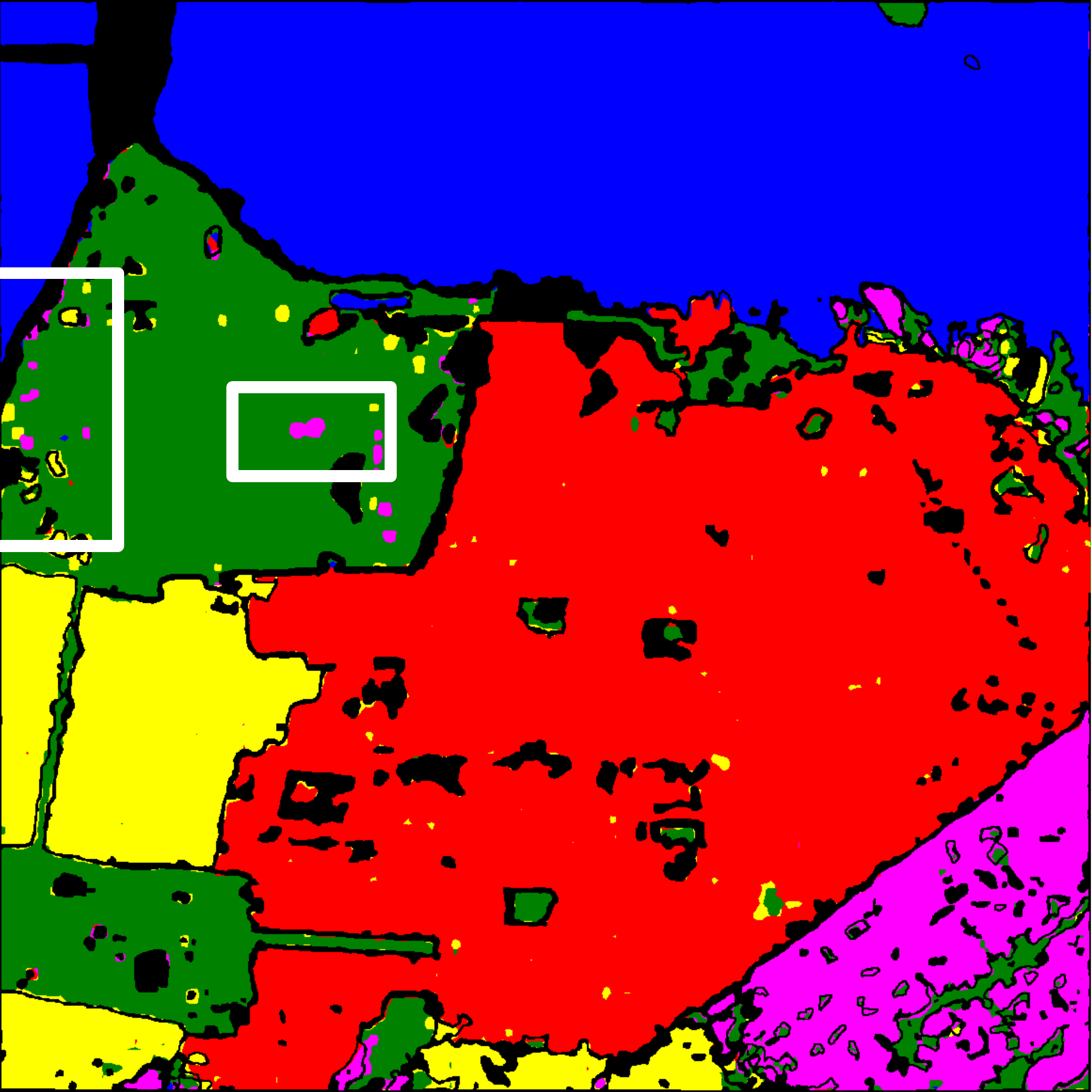}%
		\label{fig_San_CL_8}}
	\hfil
	\subfloat[]{\includegraphics[width=1.15in,height=1.15in]{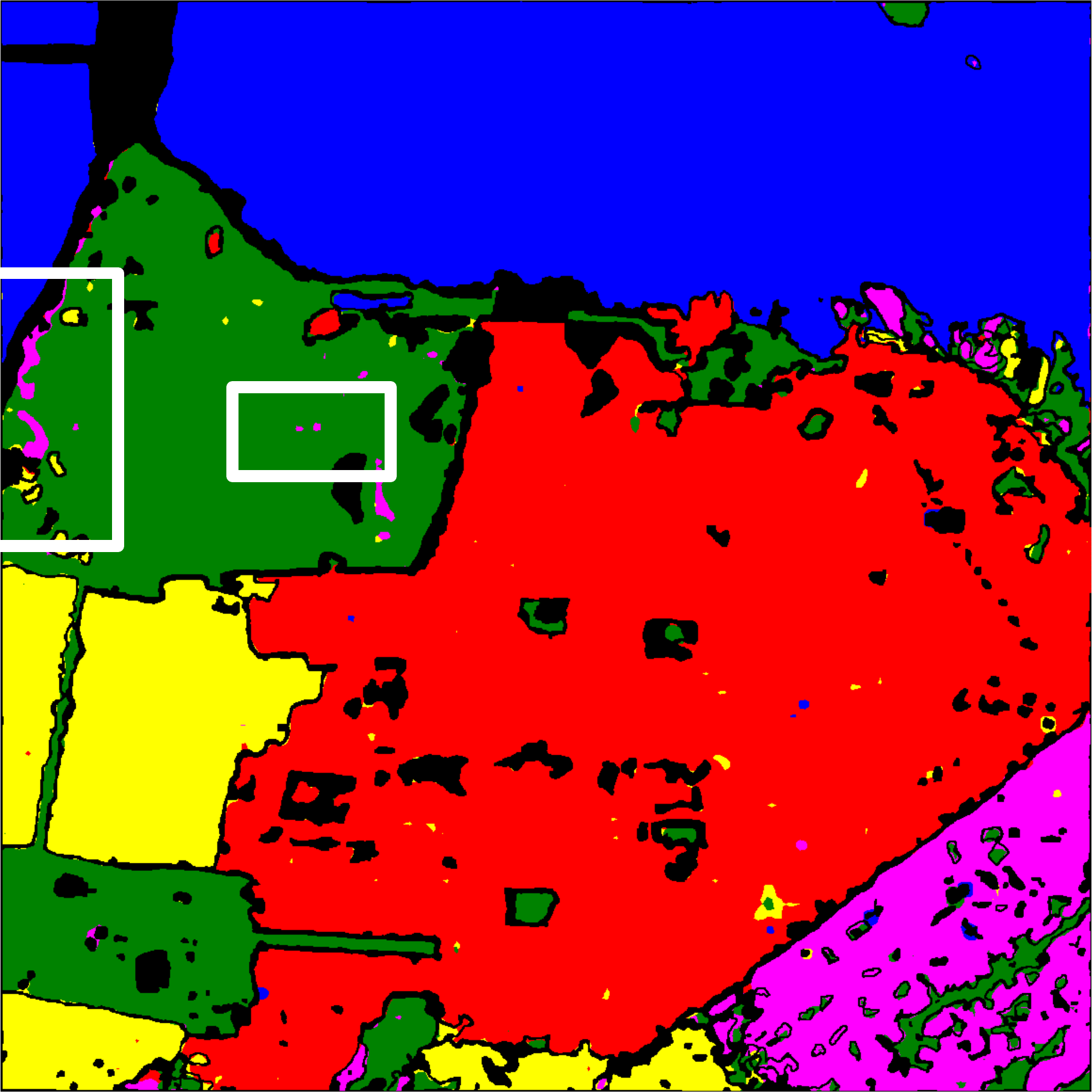}%
		\label{fig_San_CL_9}}
	\hfil
	\subfloat[]{\includegraphics[width=1.15in,height=1.15in]{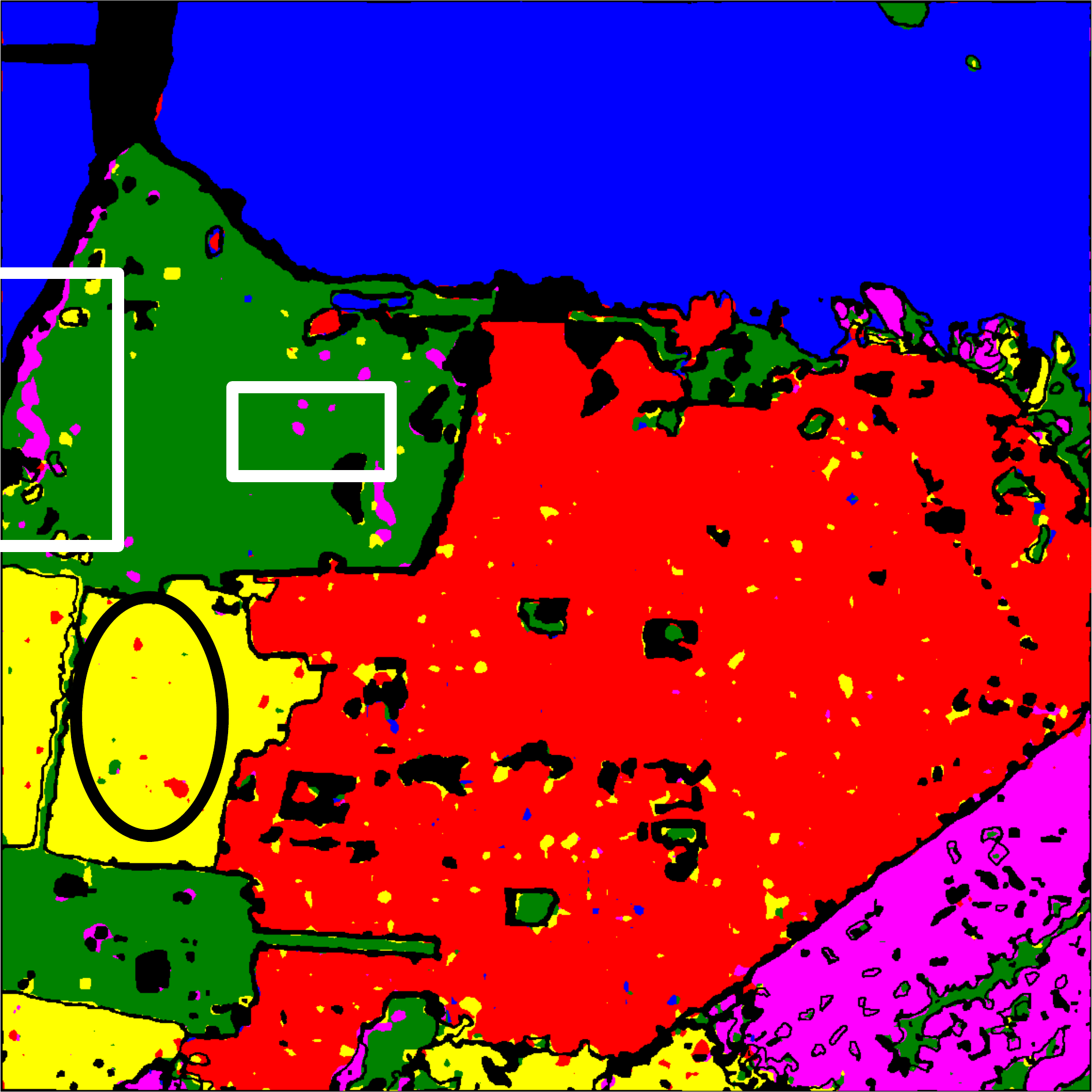}
		\label{fig_San_CL_C}}
	\hfil
	\subfloat[]{\includegraphics[width=1.15in,height=1.15in]{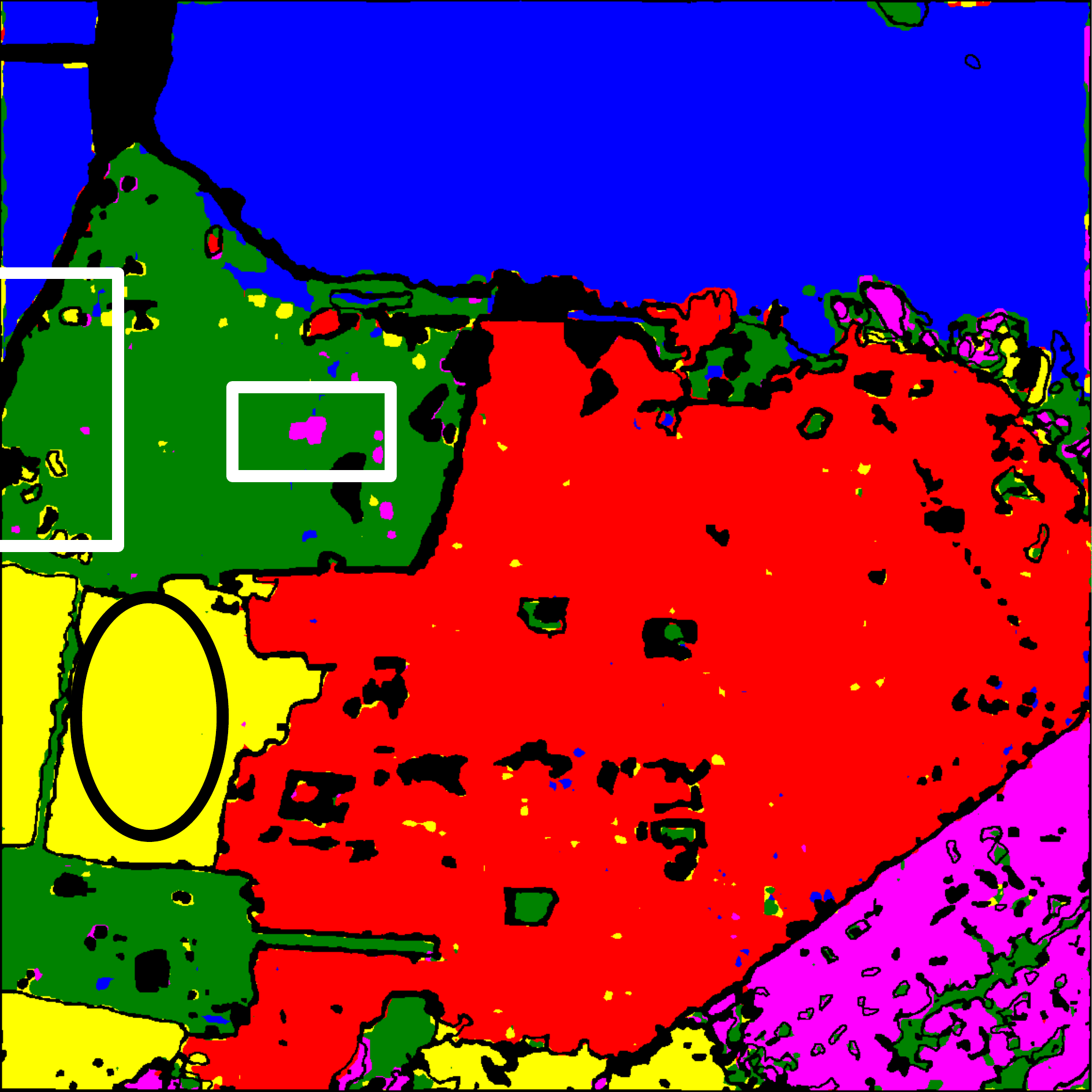}
		\label{fig_San_CL_L}}
	\hfil
	\subfloat[]{\includegraphics[width=1.15in,height=1.15in]{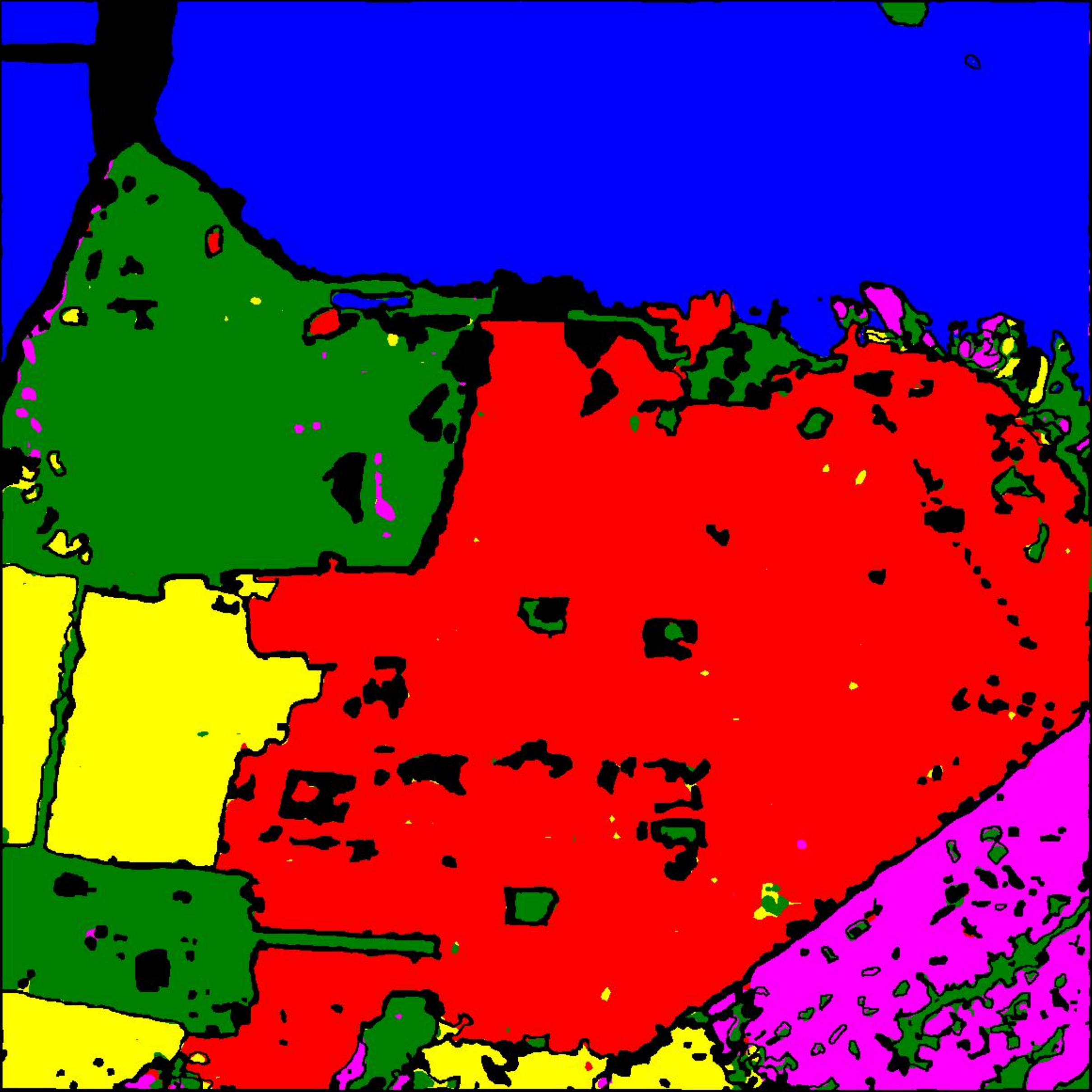}%
		\label{fig_San_CL_proposed}}
	\caption{Classification maps obtained by different algorithms on SanFrancisco\_CL. (a) Ground Truth. (b) tbCNN. (c) S2ENet.(d) GLT.(e) ExViT.(f) mCrossPA.(g) SepDGConv.(h) AsyFFNet. (i) CMGFNet. (j) Ours(C). (k) Ours(L). (l) Ours(C+L).}
	\label{Fig_San_CL}
\end{figure*}

In the following comparison experiments, the results are displayed as numerical accuracies and classification maps, which are shown in Table \ref{Tab_HB_SL}-Table \ref{Tab_Fle_CP} and Fig. \ref{Fig_HB_SL}-Fig. \ref{Fig_Fle_CP}. The best classification results in tables are highlighted in bold and the advantage values in two frequency data are highlighted in underline. The color codes in Fig. \ref{Fig_HB_SL}-Fig. \ref{Fig_Fle_CP} are the same as the ground truth maps in Fig. \ref{Fig_HB}-Fig. \ref{Fig_Fle}. Besides, in Fig. \ref{Fig_band}, we present the probability histogram of each frequency data serves as a teacher model for reference.

\begin{table*}[!t]
	\renewcommand{\arraystretch}{1.2} 
	\centering
	\caption{Classification Accuracy of different algorithms on SanFrancisco\_CL.}
	\begin{tabular}{ccccccccc|ccc}  
		\hline
		\hline
		\multirow{3}{*}{Class} &  \multicolumn{11}{c}{Performance} \\
		\cline{2-12} 
		& \multirow{2}{*}{tbCNN} & \multirow{2}{*}{S2ENet} & \multirow{2}{*}{GLT} & \multirow{2}{*}{ExViT} & \multirow{2}{*}{mCrossPA} & \multirow{2}{*}{SepDGConv} & \multirow{2}{*}{AsyFFNet} &  \multirow{2}{*}{CMGFNet} & \multicolumn{3}{c}{Ours} \\
		\cline{10-12} 
		& & & &  &  & & & & C & L & C+L\\
		\hline
		Forest&	96.62&	95.79&	97.44&	94.42&	96.16&	97.69&	96.79&	96.88& 
		91.33&	\underline{92.91}&	\textbf{97.90}\\
		
		Water&	99.52&	99.78&	99.77&	99.77&	99.94&	99.92&	\textbf{99.94}&	\textbf{99.94}& 
		\underline{99.62}&	98.51&	99.93\\
		
		Urban1 &	97.62&	99.56&	98.85&	99.63&	99.30&	99.08&	99.23&	99.32& 
		97.82&	\underline{98.56}&	\textbf{99.43}\\
		
		Urban2 &	97.70&	98.20&	98.27&	98.31&	98.25&	98.39&	98.59&	\textbf{99.09}& 
		\underline{98.38}& 95.62&	98.49\\
		
		Urban3 &	95.02&	98.39&	97.73&	\textbf{98.58}&	98.16&	96.25&	98.19&	96.67& 
		\underline{96.91}&	96.69&	97.45\\
		
		\hline
		OA(\%)&	97.80&	98.74&	98.72&	98.54&	98.74&	98.82&	98.49&	98.85&
		96.86&	\underline{97.09}&	\textbf{99.08}\\
		AA(\%)&	97.30&	98.35&	98.41&	98.15&	98.36&	98.26&	98.15&	98.38	&
		96.21&	\underline{96.55}&	\textbf{98.64}\\
		$\kappa\times$100&	96.98&	98.27&	98.24&	97.99&	98.27&	98.37&	97.93&	98.42&
		95.69&	\underline{96.01}&	\textbf{98.73}\\
		\hline
		\hline
	\end{tabular}%
	\label{Tab_San_CL}%
\end{table*}%

\subsubsection{Results on Hebei\_SL}
For quantitative evaluation, the classification accuracy of each algorithm is shown in Table \ref{Tab_HB_SL}. It can be seen that the proposed SKDNet-SSR has the highest classification accuracy in most categories. And S-band has obvious advantages on road, L-band performs better on farmland. From Fig. \ref{band_HB_SL}, in terms of road, DGSD mainly chooses S-band as the teacher model, while in farmland, DGSD mainly chooses L-band as the teacher model. Therefore, the proposed algorithm can achieve good classification results on both road and farmland, achieving the complementary utilization of S-band and L-band on different land covers.

For qualitative evaluation, Fig. \ref{Fig_HB_SL} shows the classification maps of each algorithm. In Fig. \ref{fig_HB_SL_S}, the road in the upper oval areas is classified more accurately than Fig. \ref{fig_HB_SL_L}, but at the same time, the Fig. \ref{fig_HB_SL_S} misclassifies the farmland in the lower white rectangular areas as road; In Fig. \ref{fig_HB_SL_L}, the farmland in the white rectangular areas performs well, but there are some clear misclassifications in the above road areas. These classification maps also demonstrate the advantages of S-band on road and L-band on farmland. In addition, Fig. \ref{fig_HB_SL_2} misclassifies a large number of road pixels as farmland. This is because the tbCNN uses stacking operation to process the dual-frequency data, which biases the network learning toward the L-band and the advantages of the S-band on road are ignored. Similar situation also occurs in the classification map of S2ENet. In Fig. \ref{fig_HB_SL_4}-Fig. \ref{fig_HB_SL_9},  the road in the purple oval areas is better classified, but the farmland in the white rectangular area below is misclassified to varying degrees. This is due to the emphasis on the advantages of S-band on road, while ignoring the advantages of L-band on farmland. As shown in Fig. \ref{fig_HB_SL_proposed}, the proposed algorithm utilizes gate-selection technology to emphasize different advantages of different frequency data, thus performing well on both road and farmland. 

\subsubsection{Results on SanFrancisco\_CL}
For quantitative evaluation, Table \ref{Tab_San_CL} shows the classification accuracy of all algorithms on SanFrancisco\_CL. It can be seen that the classification accuracy of each category is relatively high. This is because there are fewer categories in this region, and the separation among categories is strong. Besides, from Fig. \ref{band_San_CL}, L-band has advantages in Forest and Urban1, C-band has advantages in several other categories, which is consistent with the underlined results in Table \ref{Tab_San_CL}.

For qualitative evaluation, Fig. \ref{Fig_San_CL} shows the classification maps of each algorithm. The white oval region in Fig. \ref{fig_San_CL_2} contains significant noise, which is due to its limited use of local spatial convolution operations and insufficient learning of dual-frequency data fusion. In Fig. \ref{fig_San_CL_9}, the classification results of rectangular areas are similar to those of C-band; In Fig. \ref{fig_San_CL_8}, the classification results of rectangular areas are similar to those of L-band; In Fig. \ref{fig_San_CL_5}, the left rectangular box leans towards the C-band classification result, while the right rectangular box leans towards the L-band classification result. All of these algorithms are more or less biased towards one frequency band, and therefore do not fully utilize the complementary nature of the dual-frequency PolSAR data on different ground objects. As for the classification map of the proposed SKDNet-SSR, the advantages of C-band and L-band are both taken into account in the aforementioned regions, achieving a classification map that is closer to the ground truth. 

\begin{table*}[!t]
	\renewcommand{\arraystretch}{1.2} 
	\centering
	\caption{Classification Accuracy of different algorithms on Flevoland\_CL.}
	\begin{tabular}{ccccccccc|ccc}  
		\hline
		\hline
		\multirow{3}{*}{Class} &  \multicolumn{11}{c}{Performance} \\
		\cline{2-12} 
		& \multirow{2}{*}{tbCNN} & \multirow{2}{*}{S2ENet} & \multirow{2}{*}{GLT} & \multirow{2}{*}{ExViT} & \multirow{2}{*}{mCrossPA} & \multirow{2}{*}{SepDGConv} & \multirow{2}{*}{AsyFFNet} &  \multirow{2}{*}{CMGFNet} & \multicolumn{3}{c}{Ours} \\
		\cline{10-12} 
		& & & &  &  & & & & C & L & C+L\\
		\hline
		Grass& 	84.65&	82.32&	93.15&	94.82&	94.22&	94.95&	94.00&	94.32&
		\underline{91.11}& 89.25&	\textbf{95.93}\\
		
		Flax& 	99.79&	99.66&	\textbf{99.94}&	99.75&	99.85&	99.88&	99.71&	99.88&
		\underline{\textbf{99.94}}&	97.90&	99.89\\
		
		Potato&	97.35&	98.38&	97.79&	97.95&	97.71&	97.60&	96.81&	97.22&
		94.56&	\underline{95.83}&	\textbf{98.45}\\
		
		Wheat&	98.14&	91.12&	97.39&	98.73&	97.64&	98.46&	97.81&	97.76&
		88.86&	\underline{95.29}&	\textbf{99.29}\\
		
		Rapeseed&	99.76&	99.86&	\textbf{99.98}&	99.96&	99.94&	99.87&	99.96&	99.92&
		99.61&	\underline{99.85}&	99.93\\
		
		Beet&	89.14&	90.65&	96.01&	97.39&	95.64&	95.95&	96.49&	95.43&
		83.86&	\underline{85.77}&	\textbf{98.35}\\
		
		Barley&	97.51&	94.61&	98.23&	99.30&	98.72&	99.24&	\textbf{99.59}&	98.48&
		95.81&  \underline{96.95}&	98.68\\
		
		Peas&	95.28&	99.06&	99.74&	99.54&	99.21&	99.35&	99.35&	\textbf{99.88}&
		\underline{96.85}&	93.69&	99.74\\
		
		Maize&	92.89&	97.05&	96.44&	99.15&	96.36&	98.33&	95.38&	97.20&
		83.28&	\underline{93.97}&	\textbf{99.39}\\
		
		Beans&	96.97&	80.22&	99.29&	99.14&	99.09&	99.04&	98.69&	99.60& 
		\underline{99.80}& 97.60&	\textbf{100.00}	\\
		
		Fruit&	96.49&	\textbf{98.07}&	96.46&	96.67&	96.24&	97.44&	96.60&	97.14
		&91.65& \underline{96.14}&	96.87\\
		
		Onions&	98.64&	99.68&	99.91&	\textbf{100.00}&	99.73&	99.91&	99.91&	99.37&
		\underline{99.73}&	99.27&	99.91\\
		
		Lucerne& 99.98&	99.59&	99.96&	99.90&	99.79&	99.94&	\textbf{99.99}&	99.51&
		93.56&	\underline{99.49}&	99.82\\
		
		Building&	83.95&	84.79&	90.94&	88.89&	89.07&	90.78&	\textbf{92.24}&	91.38&
		82.83&	\underline{90.52}&	90.12\\
		
		Road&	68.69&	90.47&	79.46&	78.71&	79.02&	85.38&	76.89&	82.84&
		69.67& \underline{73.23}&	\textbf{87.15}\\
		
		\hline
		OA(\%)&	93.52&	93.51&	95.97&	96.68&	96.05&	96.96&	96.03&	96.31&
		90.22&	\underline{93.10}&	\textbf{97.66}\\
		AA(\%)	&	93.28&	93.70&	96.31&	96.66&	96.15&	97.07&	96.23&	96.66&
		91.20&	\underline{94.06}&	\textbf{97.57}\\
		$\kappa\times$100&	92.56&	92.57&	95.37&	96.18&	95.45&	96.50&	95.43&	95.76&
		88.79&	\underline{92.07}&	\textbf{97.31}\\
		\hline
		\hline
	\end{tabular}%
	\label{Tab_Fle_CL}%
\end{table*}%

\begin{figure*}[!t]
	\centering
	\subfloat[]{\includegraphics[width=1.15in,height=1.15in]{Fig/2_Pauli/GroundTruth_Fle}
		\label{fig_Fle_CL_1}}
	\hfil
	\subfloat[]{\includegraphics[width=1.15in,height=1.15in]{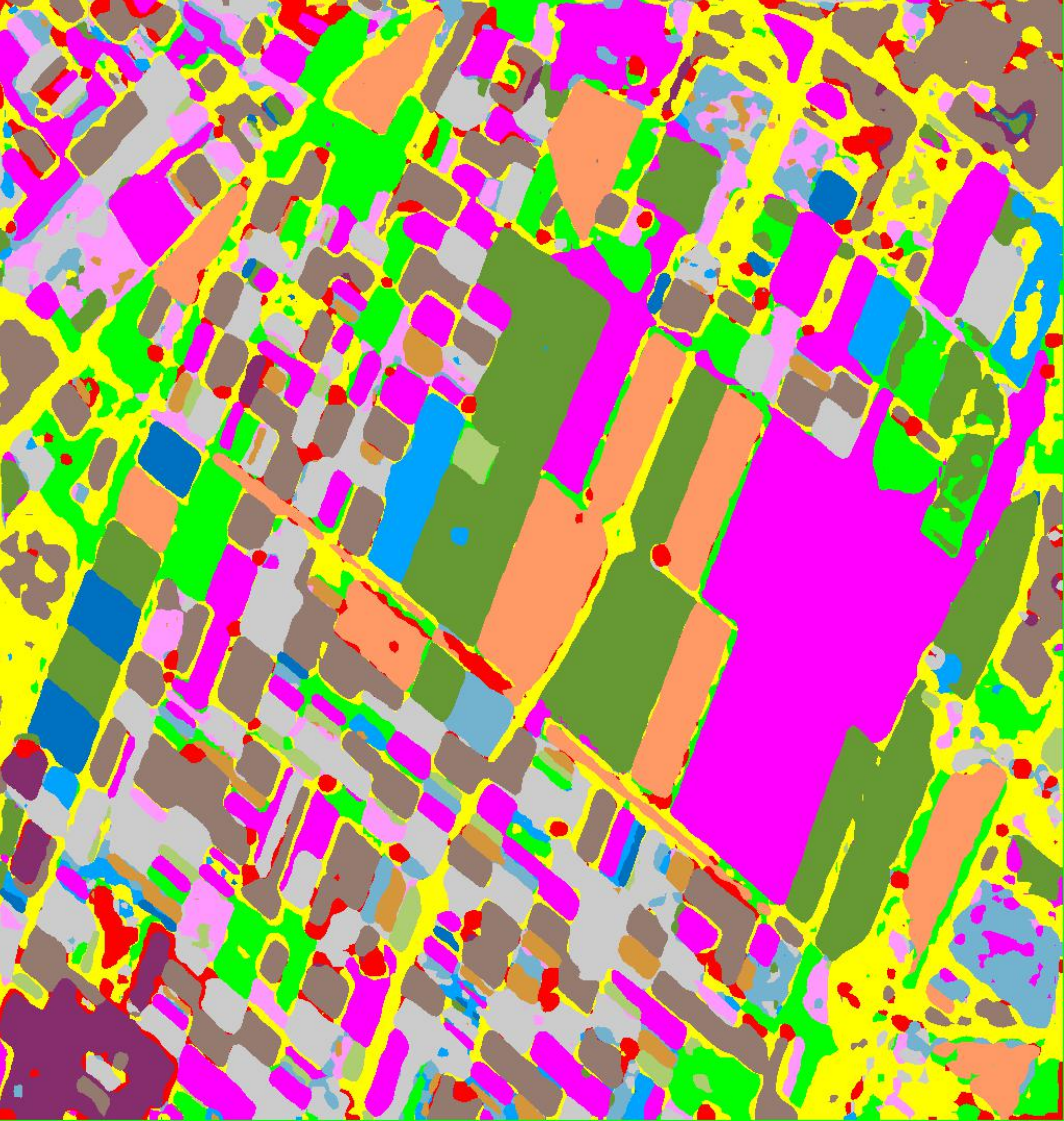}%
		\label{fig_Fle_CL_2}}
	\hfil
	\subfloat[]{\includegraphics[width=1.15in,height=1.15in]{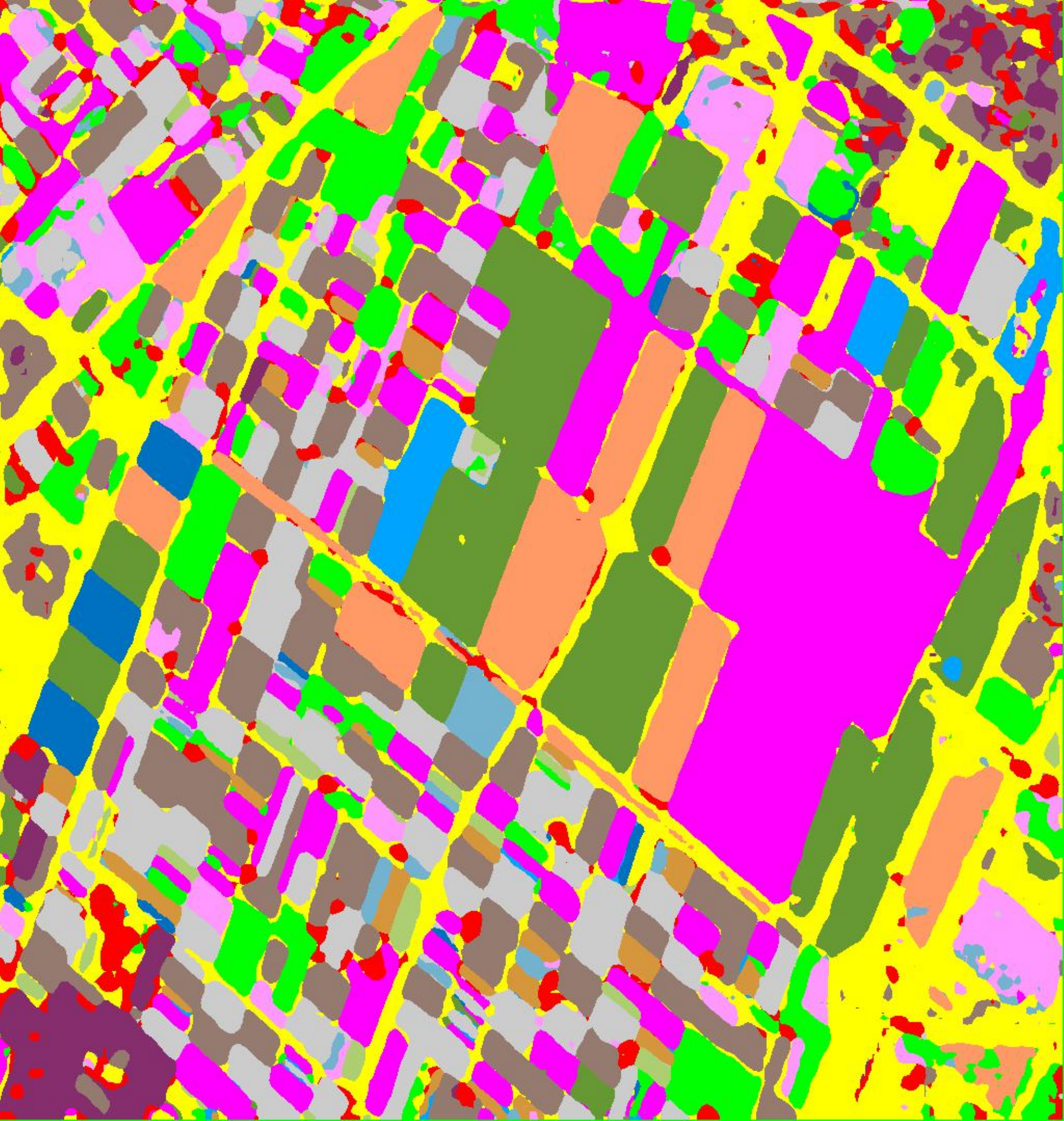}%
		\label{fig_Fle_CL_3}}
	\hfil
	\subfloat[]{\includegraphics[width=1.15in,height=1.15in]{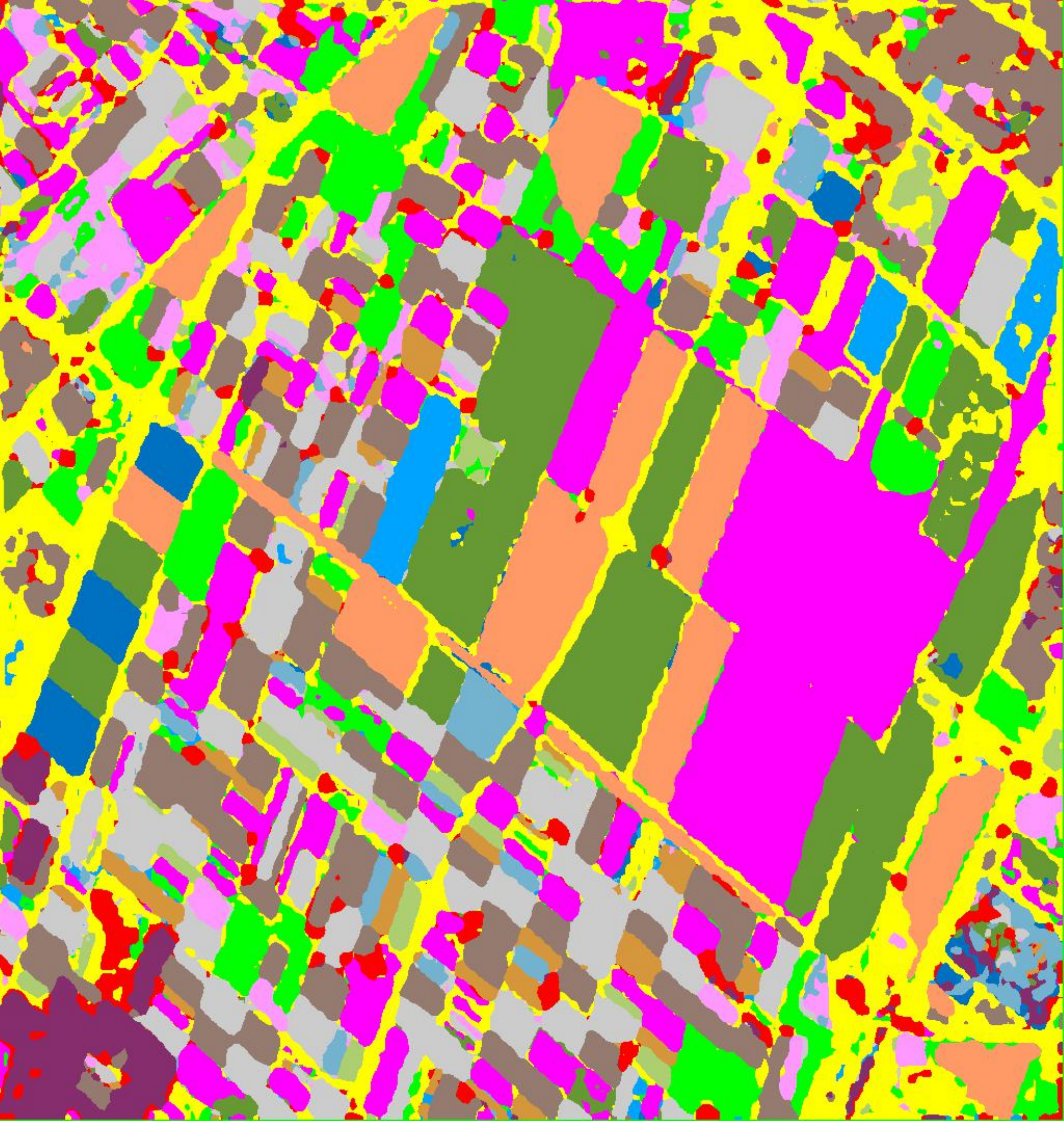}%
		\label{fig_Fle_CL_4}}
	\hfil
	\subfloat[]{\includegraphics[width=1.15in,height=1.15in]{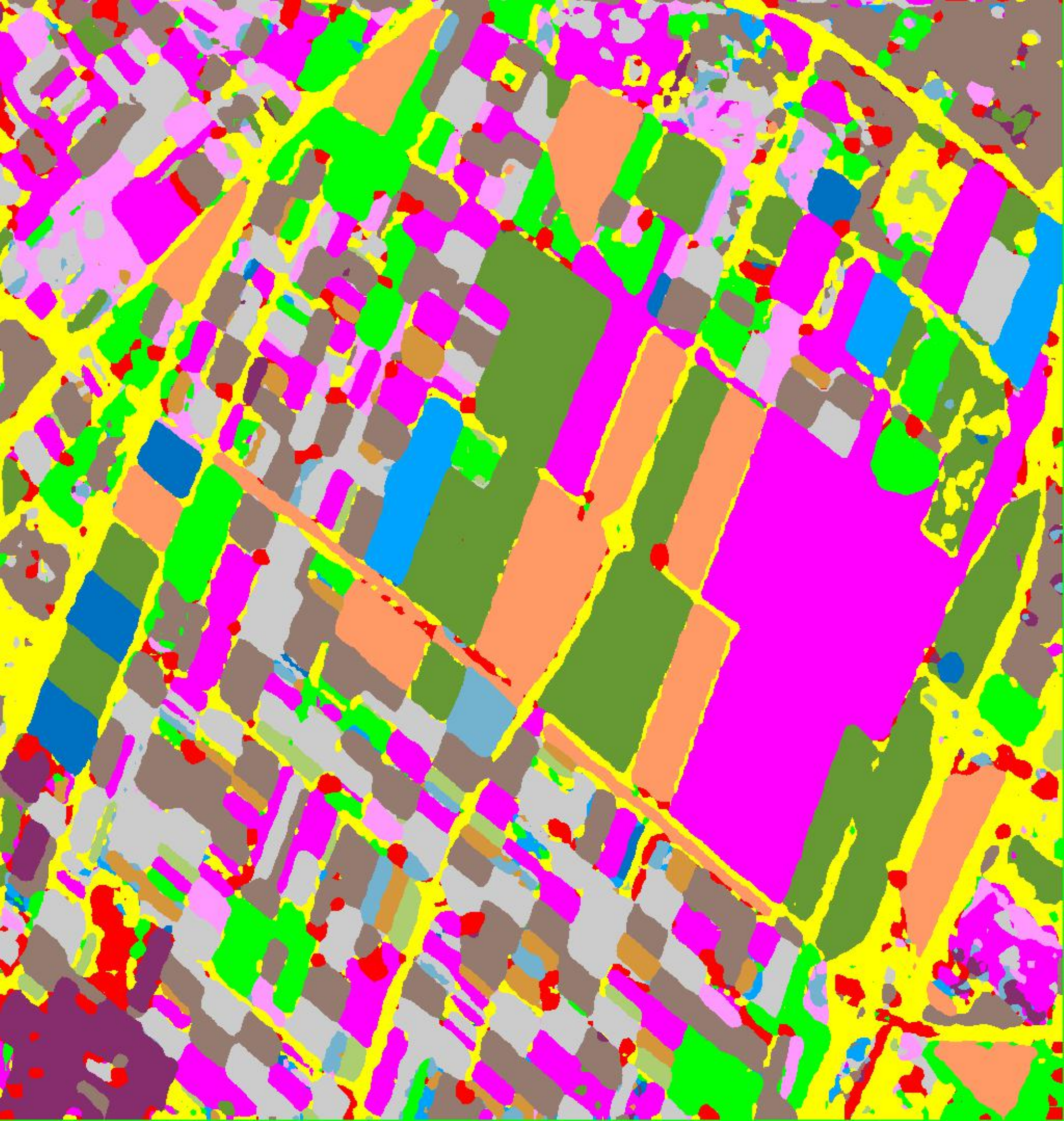}%
		\label{fig_Fle_CL_5}}
	\hfil
	\subfloat[]{\includegraphics[width=1.15in,height=1.15in]{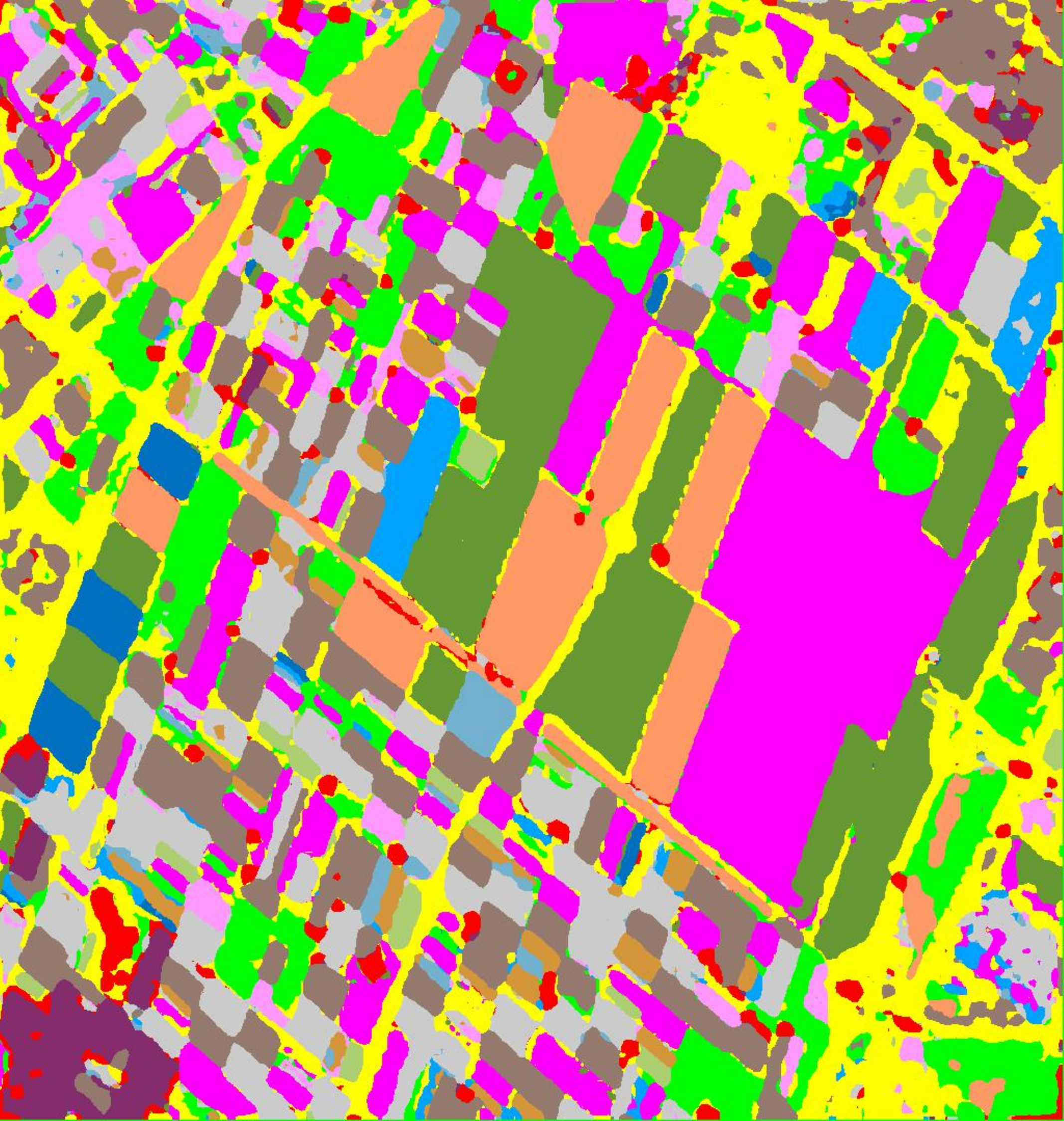}%
		\label{fig_Fle_CL_6}}
	\hfil
	\subfloat[]{\includegraphics[width=1.15in,height=1.15in]{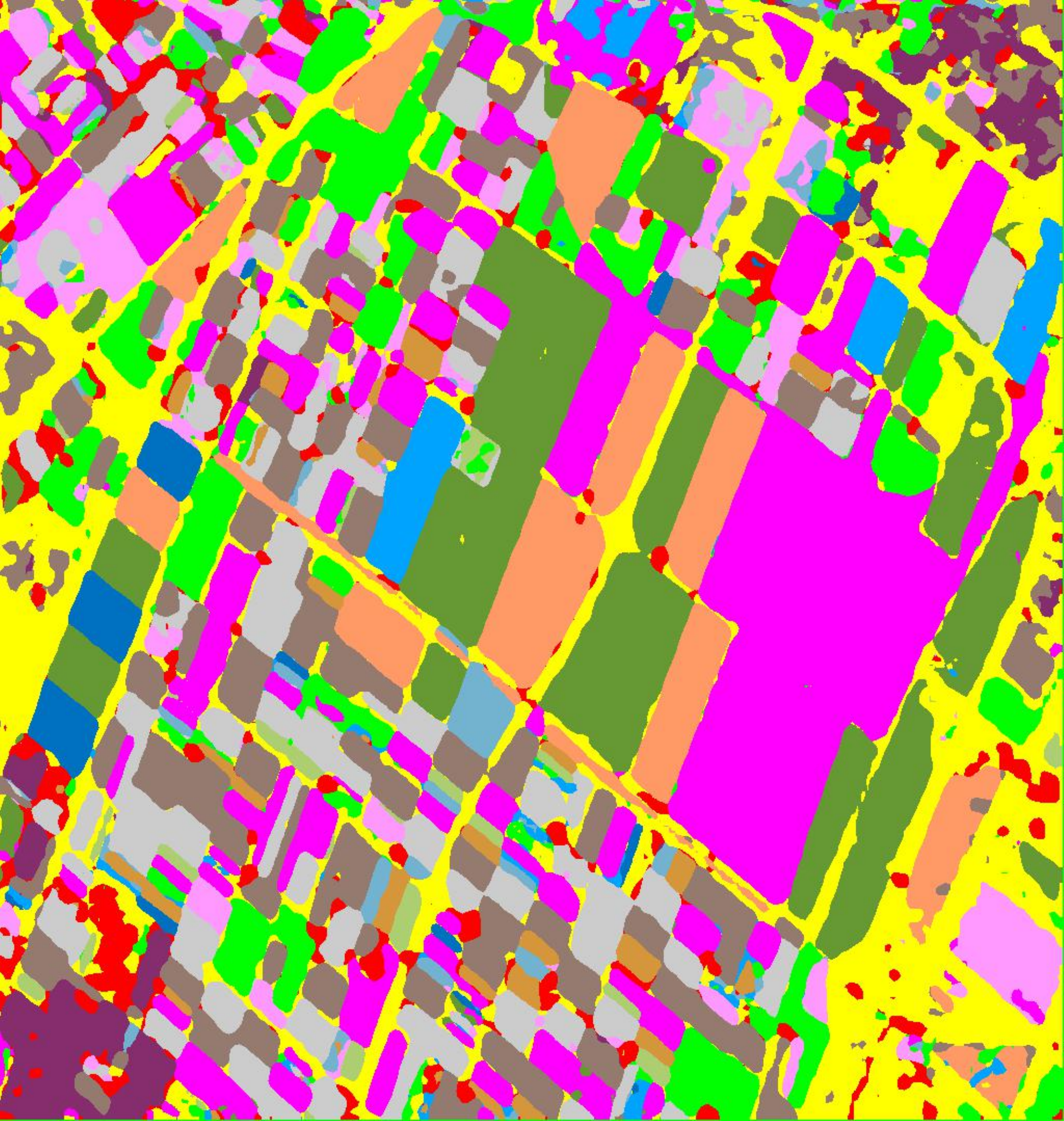}%
		\label{fig_Fle_CL_7}}
	\hfil
	\subfloat[]{\includegraphics[width=1.15in,height=1.15in]{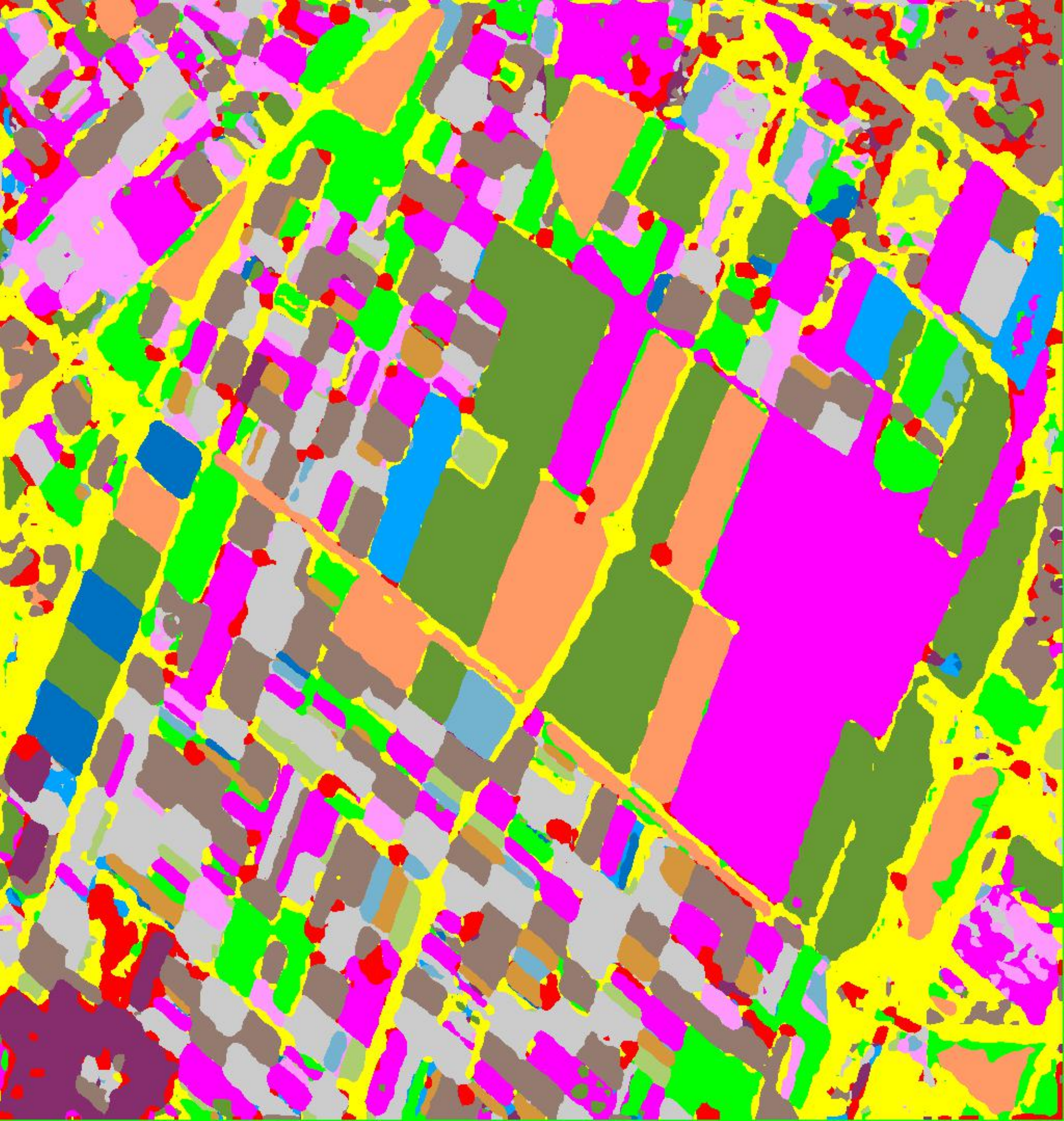}%
		\label{fig_Fle_CL_8}}
	\hfil
	\subfloat[]{\includegraphics[width=1.15in,height=1.15in]{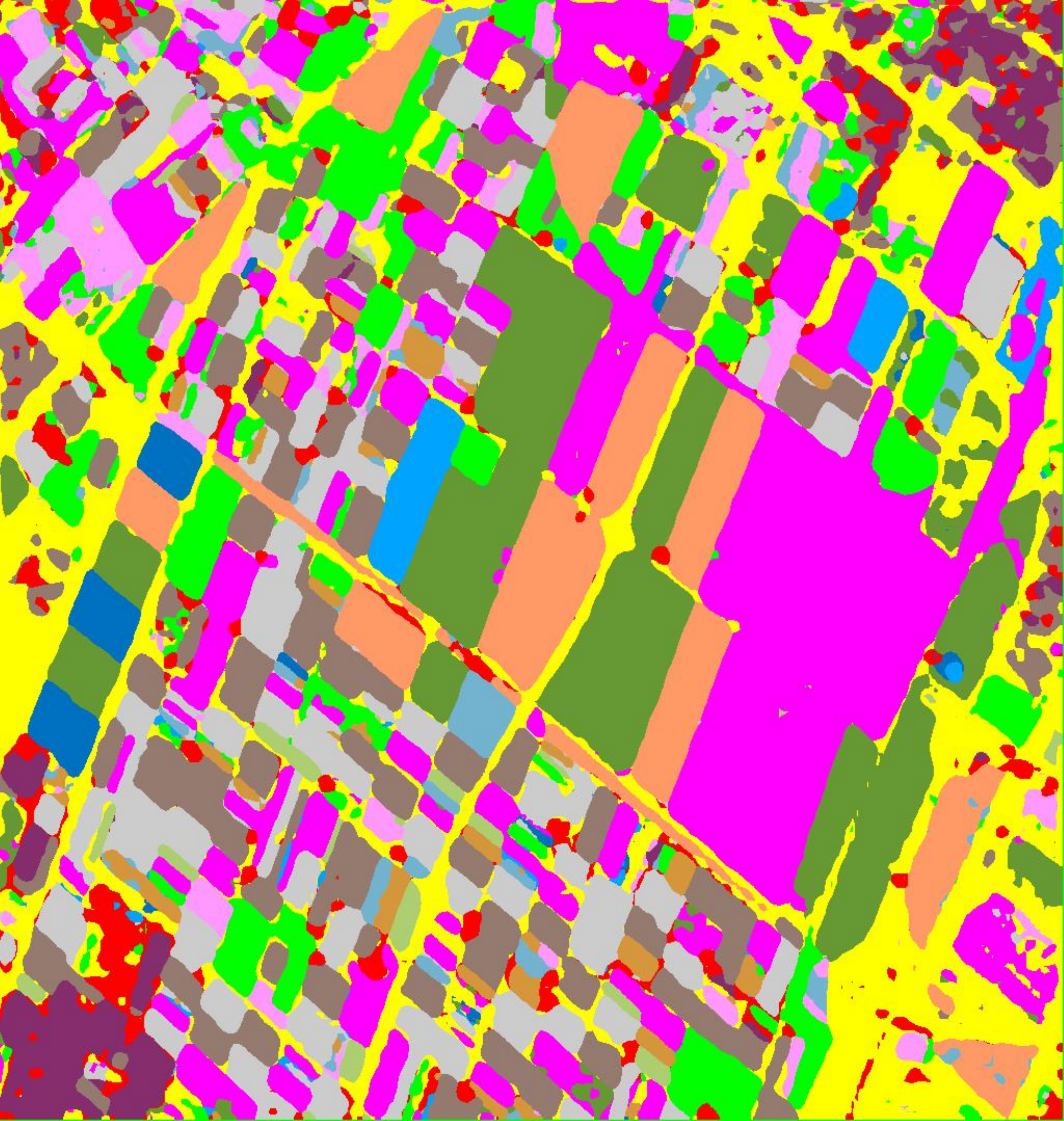}%
		\label{fig_Fle_CL_9}}
	\hfil
	\subfloat[]{\includegraphics[width=1.15in,height=1.15in]{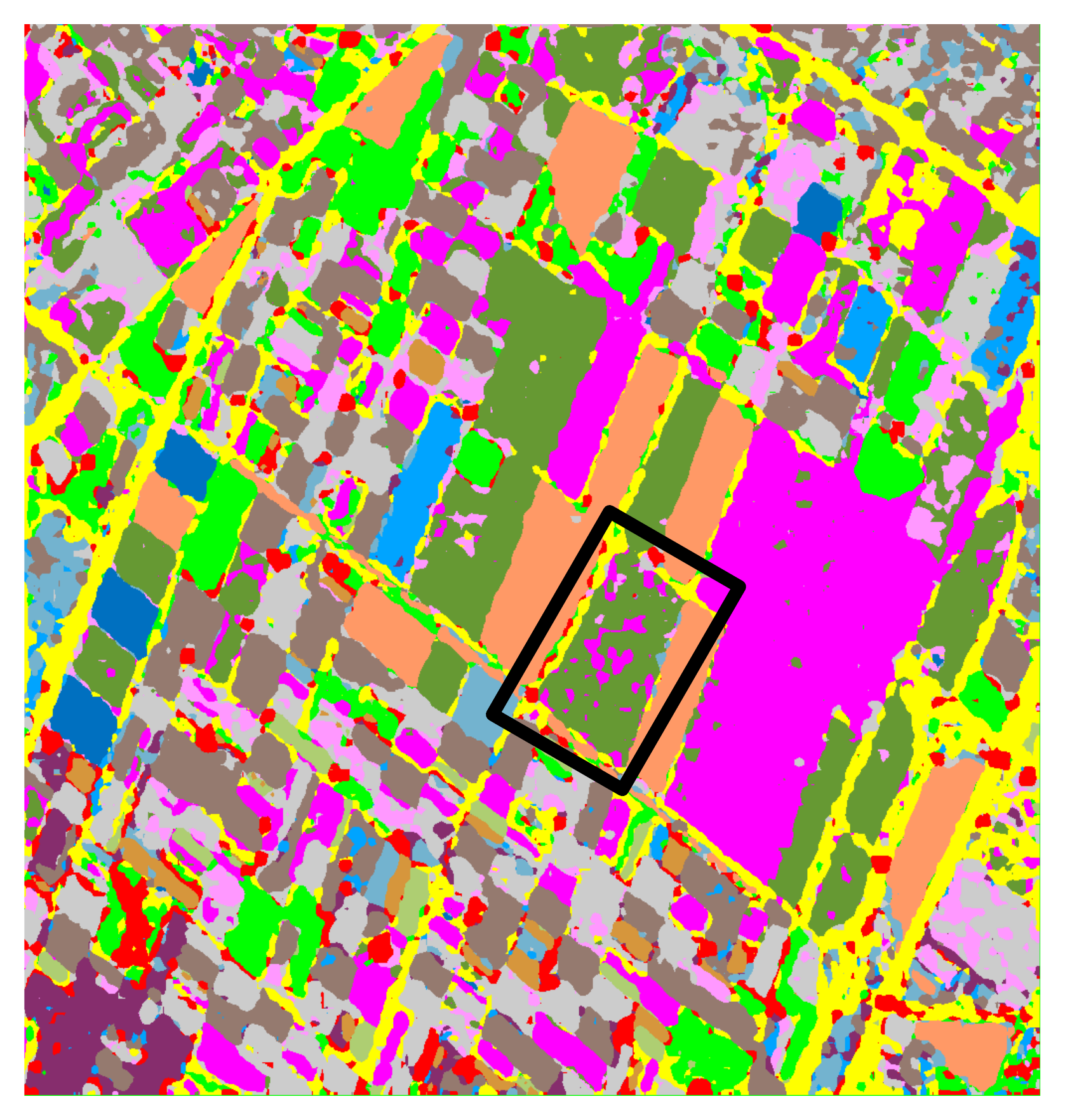}
		\label{fig_Fle_CL_C}}
	\hfil
	\subfloat[]{\includegraphics[width=1.15in,height=1.15in]{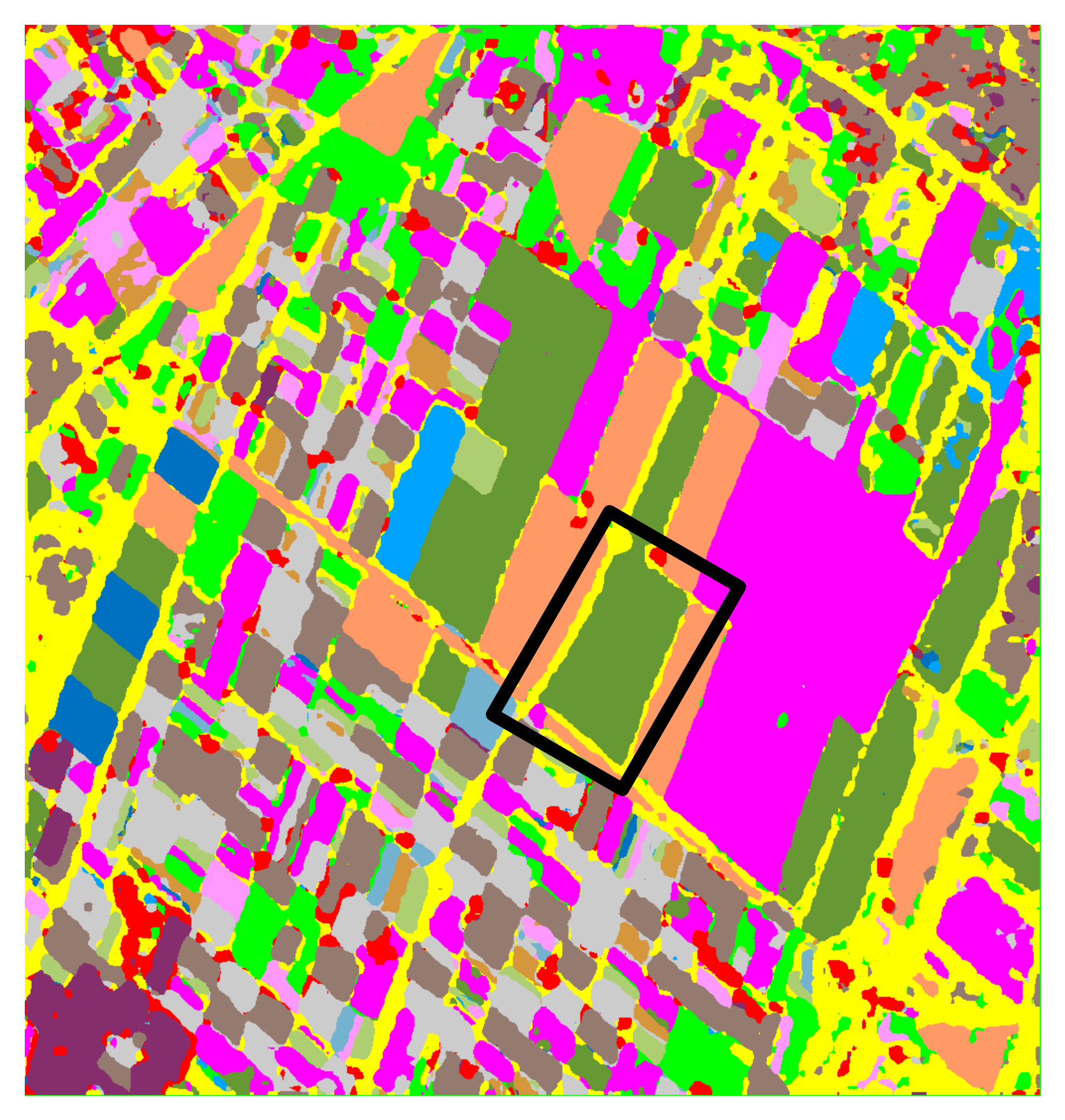}
		\label{fig_Fle_CL_L}}
	\hfil
	\subfloat[]{\includegraphics[width=1.15in,height=1.15in]{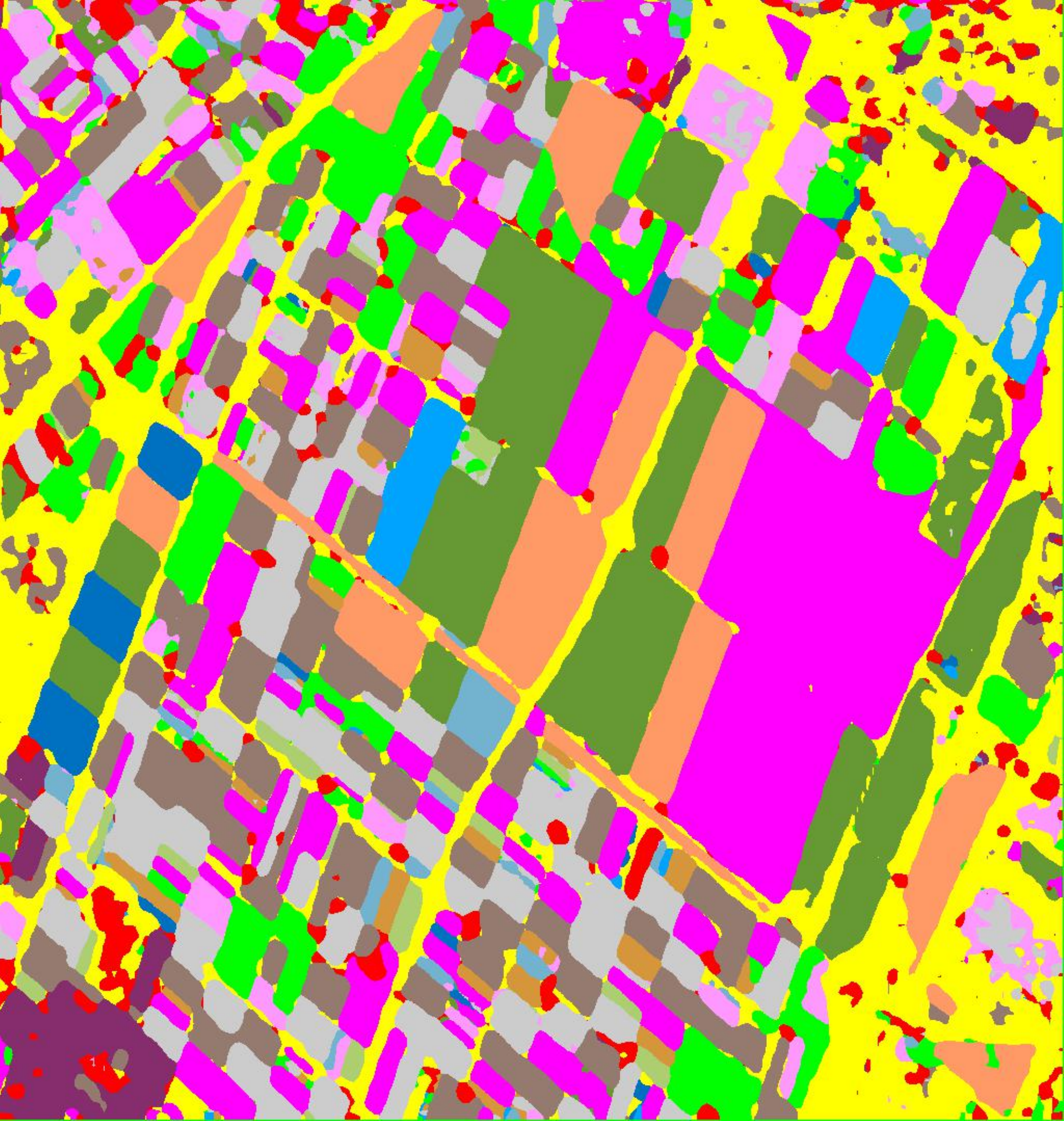}%
		\label{fig_Fle_CL_ours}}
	\caption{Classification maps obtained by different algorithms on Flevoland\_CL dataset. (a) Ground Truth. (b) tbCNN. (c) S2ENet.(d) GLT.(e) ExViT.(f) mCrossPA.(g) SepDGConv.(h) AsyFFNet. (i) CMGFNet. (j) Ours(C). (k) Ours(L). (l) Ours(C+L).}
	\label{Fig_Fle_CL}
\end{figure*}
\subsubsection{Results on Flevoland\_CL}
For quantitative evaluation, Table \ref{Tab_Fle_CL} shows the classification accuracy of different algorithms. As shown in Table \ref{Tab_Fle_CL}, compared with other classification algorithms, the proposed algorithm has the highest classification accuracy in most categories. In addition, compared with the classification results of single-frequency data, the classification performance of dual-frequency data is always better than that of single-frequency branches. 

For qualitative evaluation, Fig. \ref{Fig_Fle_CL} shows the classification maps of different algorithms. In Fig. \ref{Fig_Fle_CL}, we can clearly see that the noise in the classification map of ours(C) is more severe, especially in the black rectangular region. This is because the C-band electromagnetic wave has weak discrimination for these categories, while on ours(L), most of the categories have better classification performance. In Fig. \ref{band_Fle_CL}, we can see that most samples have chosen the L-band branch as the teacher model. Therefore, on ours(C+L), the classification performance of these categories is similar to that of the L-band. Through the DGSD strategy, the proposed SKDNet-SSR not only retains the advantages of the L-band in these categories, but also considers the supplementary information brought by C-band during the network training, thereby achieving higher overall accuracy metrics than the Ours(C) and Ours(L).

\begin{table*}[!t]
	\renewcommand{\arraystretch}{1.2} 
	\centering
	\caption{Classification Accuracy of different algorithms on Flevoland\_CP.}
	\begin{tabular}{ccccccccc|ccc}  %
		\hline
		\hline
		\multirow{3}{*}{Class} &  \multicolumn{11}{c}{Performance} \\
		\cline{2-12} 
		& \multirow{2}{*}{tbCNN} & \multirow{2}{*}{S2ENet} & \multirow{2}{*}{GLT} & \multirow{2}{*}{ExViT} & \multirow{2}{*}{mCrossPA} & \multirow{2}{*}{SepDGConv} & \multirow{2}{*}{AsyFFNet} &  \multirow{2}{*}{CMGFNet} & \multicolumn{3}{c}{Ours} \\
		\cline{10-12} 
		& & & &  &  & & & & C & P & C+P\\
		\hline
		Grass&	84.74&	64.89&	90.14&	91.07&	91.13&	94.25&	90.72&	93.08& 
		\underline{89.25}&	72.32&	\textbf{92.67}\\
		
		Flax&	99.29&	99.51&	99.08&	\textbf{99.83}&	99.78&	99.78&	\textbf{99.83}&	99.55&
		\underline{98.90}&	97.22&	\textbf{99.83}\\
		
		Potato&	97.18&	\textbf{98.13}&	96.36&	96.01&	97.26&	96.41&	95.25&	96.17&
		\underline{94.56}&	88.88&	97.67\\
		
		Wheat&	95.00&	\textbf{97.89}&	96.64&	97.36&	97.12&	97.04&	96.06&	97.10&
		88.86&	\underline{94.61}&	\textbf{97.89}\\
		
		Rapeseed&	99.91&	99.90&	\textbf{99.99}&	99.94&	99.98&	99.73&	99.87&	99.79&
		\underline{99.61}&	98.80&	99.80\\
		
		Beet&	72.20&	81.49&	89.34&	92.18&	92.12&	93.00&	92.72&	90.61&
		\underline{83.86}&	65.58&	\textbf{94.78}\\
		
		Barley&	91.27&	89.53&	97.57&	97.87&	99.01&	98.91&	98.87&	98.04&
		\underline{95.81}&	95.57&	\textbf{99.22}\\
		
		Peas&	92.54&	96.73&	96.27&	99.01&	99.11&	98.53&	98.77&	\textbf{99.59}&
		\underline{96.85}&	90.37&	98.68\\
		
		Maize&	85.01&	86.14&	93.97&	93.80&	94.14&	96.29&	96.86&	93.58&
		\underline{83.28}&	74.86&	\textbf{98.33}\\
		
		Beans&	98.03&	80.98&	\textbf{99.70}&	99.29&	99.04&	99.04&	99.39&	98.59&
		\underline{99.60}&	82.09&	99.65\\
		
		Fruit&	95.57&	\textbf{98.49}&	94.58&	95.18&	96.50&	95.01&	92.80&	94.26&
		91.65&	\underline{93.82}&	95.01\\
		
		Onions&	97.51&	98.73&	\textbf{100.00}&	99.91&	98.50&	99.68&	99.77&	99.73&
		\underline{95.29}&	94.42&	99.91\\
		
		Lucerne&	99.67&	99.71&	99.43&	99.84&	\textbf{99.88}&	99.79&	99.63&	99.57&
		93.56&	\underline{94.79}&	99.87\\
		
		Building&	86.73&	89.46&	90.81&	91.14&	92.26&	89.84&	91.69&	90.26&
		82.83&	\underline{91.27}&	\textbf{93.54}\\
		
		Road&	70.44&	\textbf{89.08}&	74.33&	80.89&	73.53&	84.09&	78.88&	81.27&
		69.67&	\underline{69.80}&	83.54\\
		
		\hline
		OA(\%)&	89.97&	91.69&	93.97&	95.08&	94.88&	95.79&	94.67&	95.01&
		\underline{90.22}&	87.12&	\textbf{96.35}\\
		
		AA(\%)	&	91.01& 91.39&	94.55&	95.55&	95.29&	96.09&	95.41&	95.39&
		\underline{91.20}&	87.02&	\textbf{96.69}\\
		
		$\kappa\times$100&	88.50&	90.46&	93.07&	94.35&	94.11&	95.17&	93.88&	94.28&
		\underline{88.79}&	85.28&	\textbf{95.80}\\
		\hline
		\hline
	\end{tabular}%
	\label{Tab_Fle_CP}%
\end{table*}%

\begin{figure*}[!t]
	\centering
	\subfloat[]{\includegraphics[width=1.15in,height=1.15in]{Fig/2_Pauli/GroundTruth_Fle}
		\label{fig_Fle_CP_1}}
	\hfil
	\subfloat[]{\includegraphics[width=1.15in,height=1.15in]{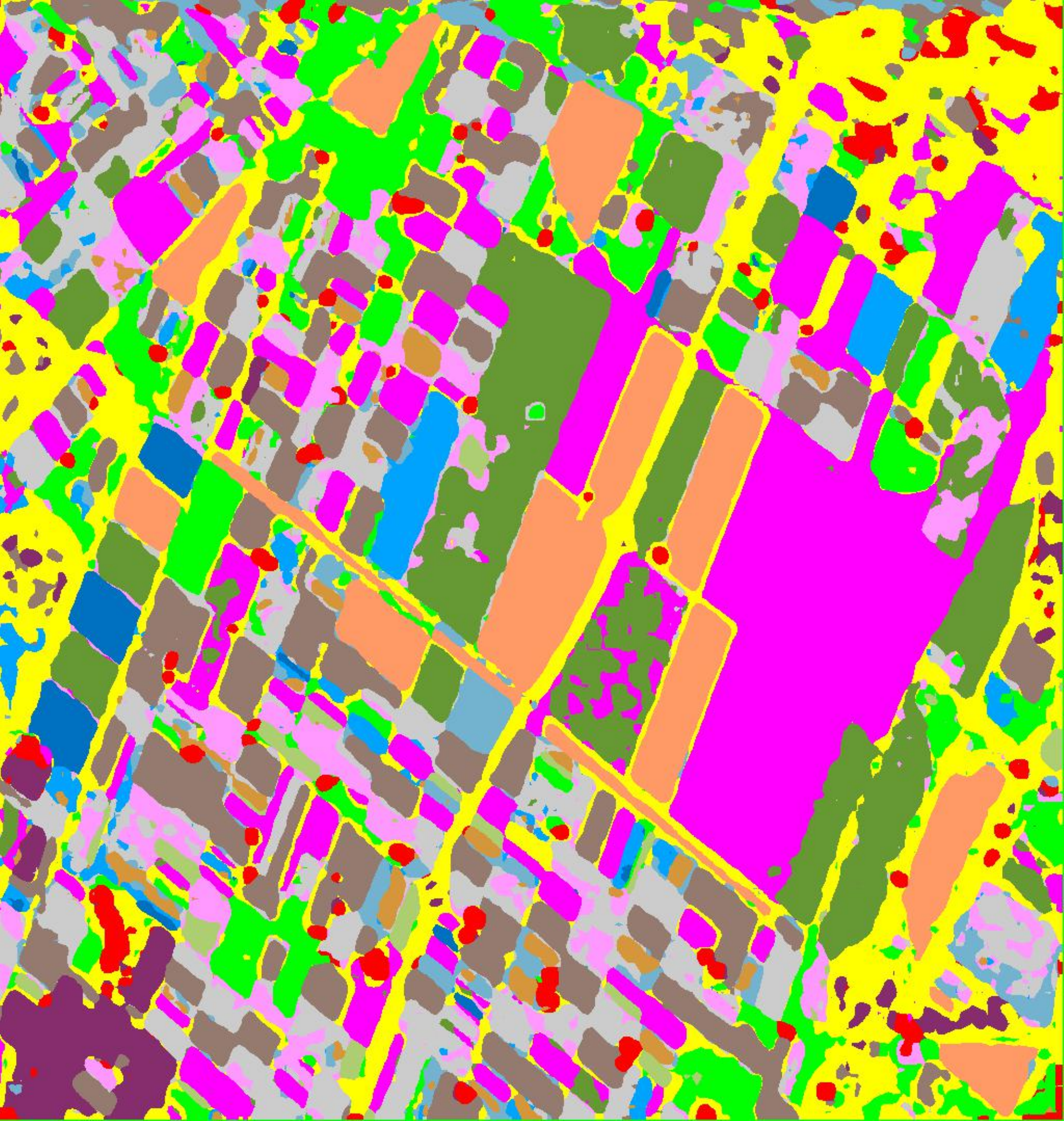}%
		\label{fig_Fle_CP_2}}
	\hfil
	\subfloat[]{\includegraphics[width=1.15in,height=1.15in]{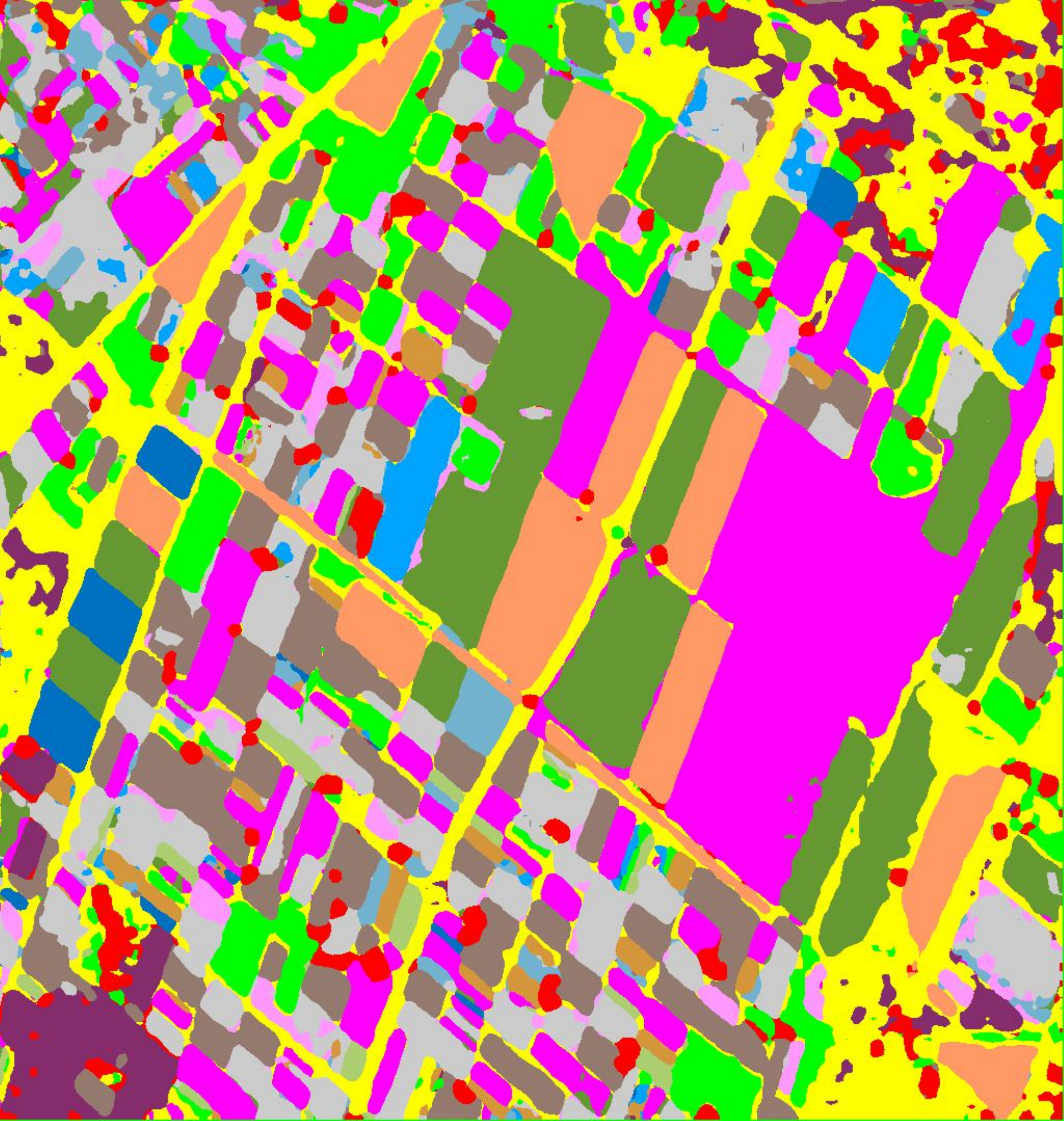}%
		\label{fig_Fle_CP_3}}
	\hfil
	\subfloat[]{\includegraphics[width=1.15in,height=1.15in]{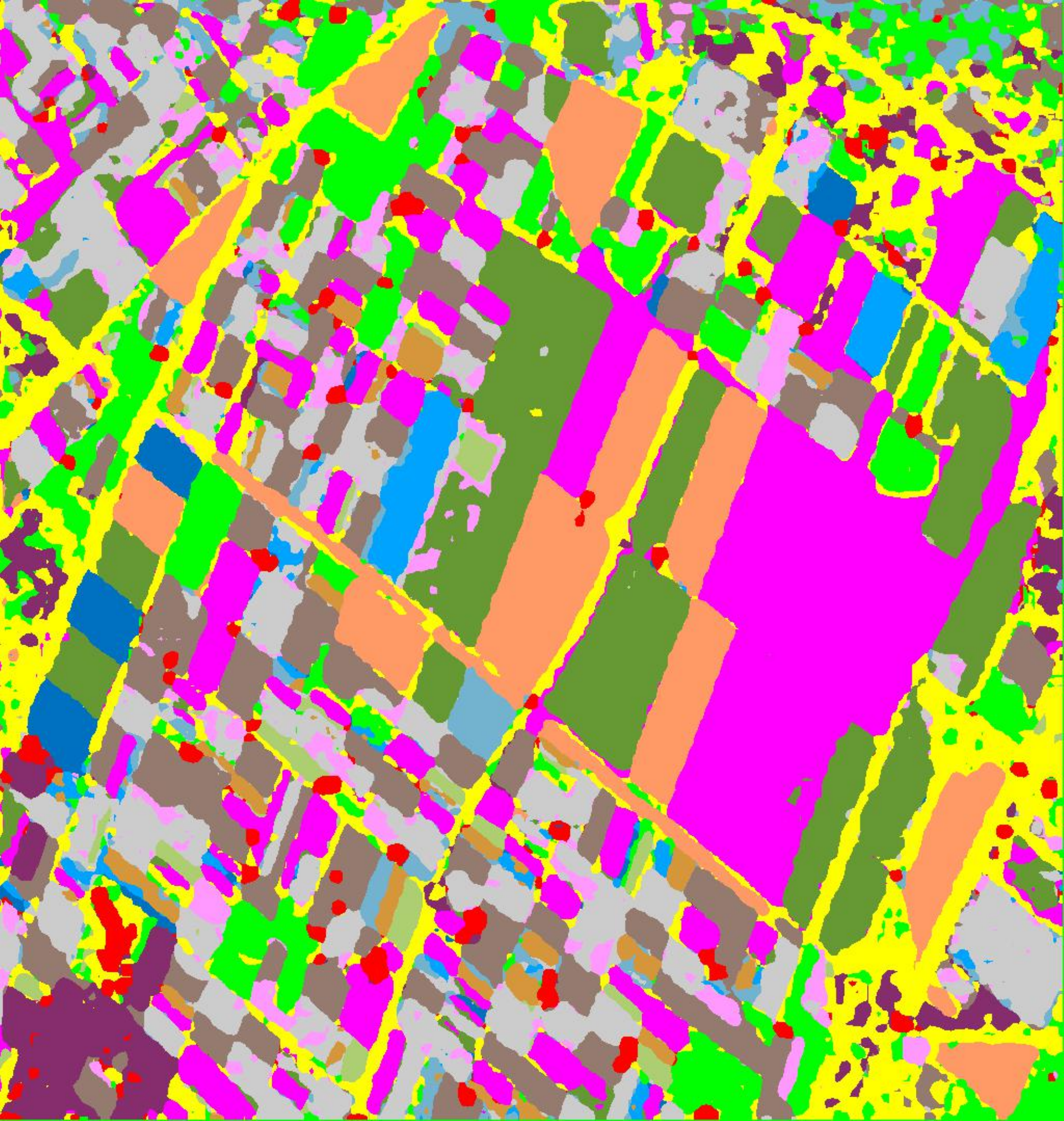}%
		\label{fig_Fle_CP_4}}
	\hfil
	\subfloat[]{\includegraphics[width=1.15in,height=1.15in]{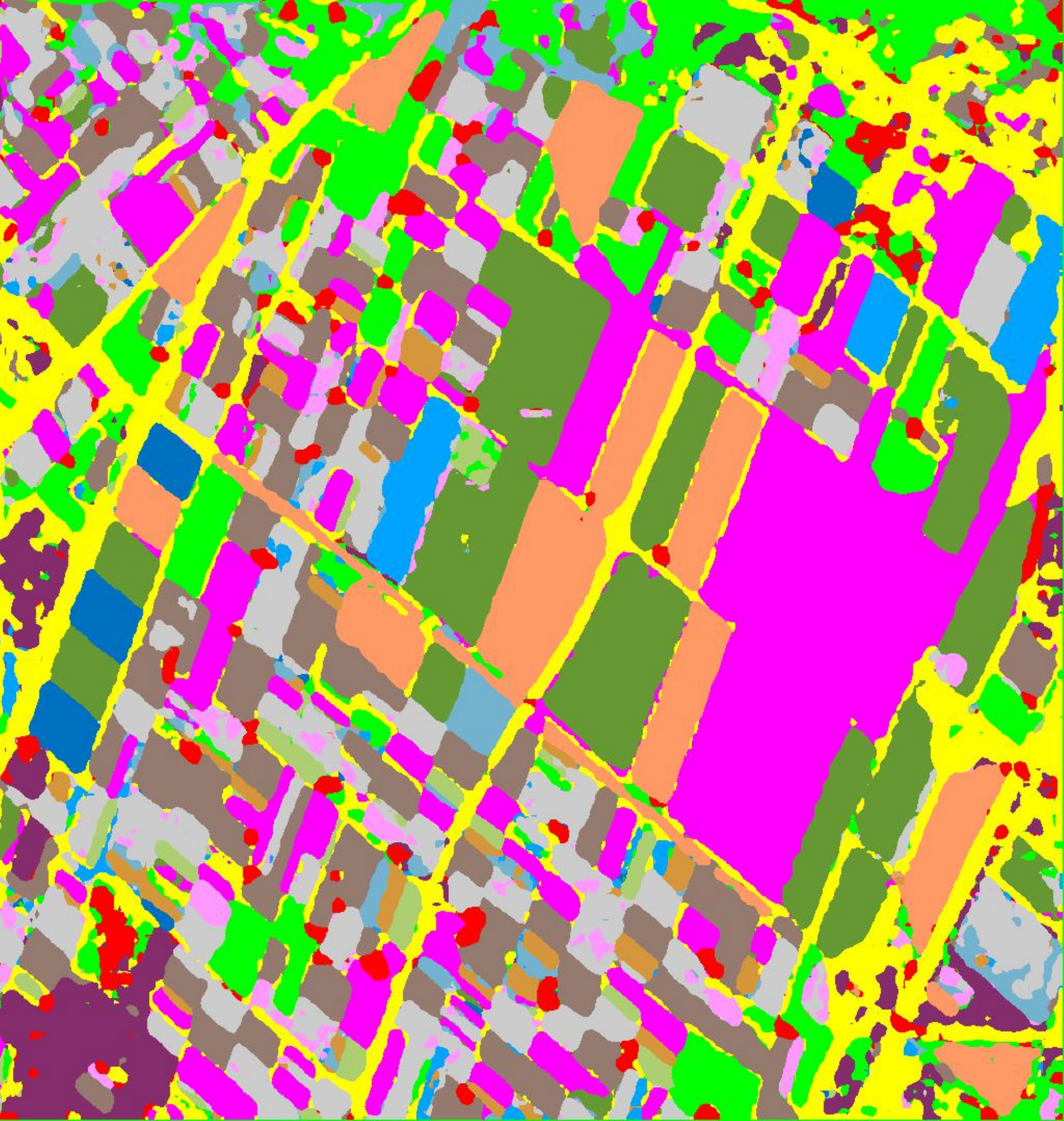}%
		\label{fig_Fle_CP_5}}
	\hfil
	\subfloat[]{\includegraphics[width=1.15in,height=1.15in]{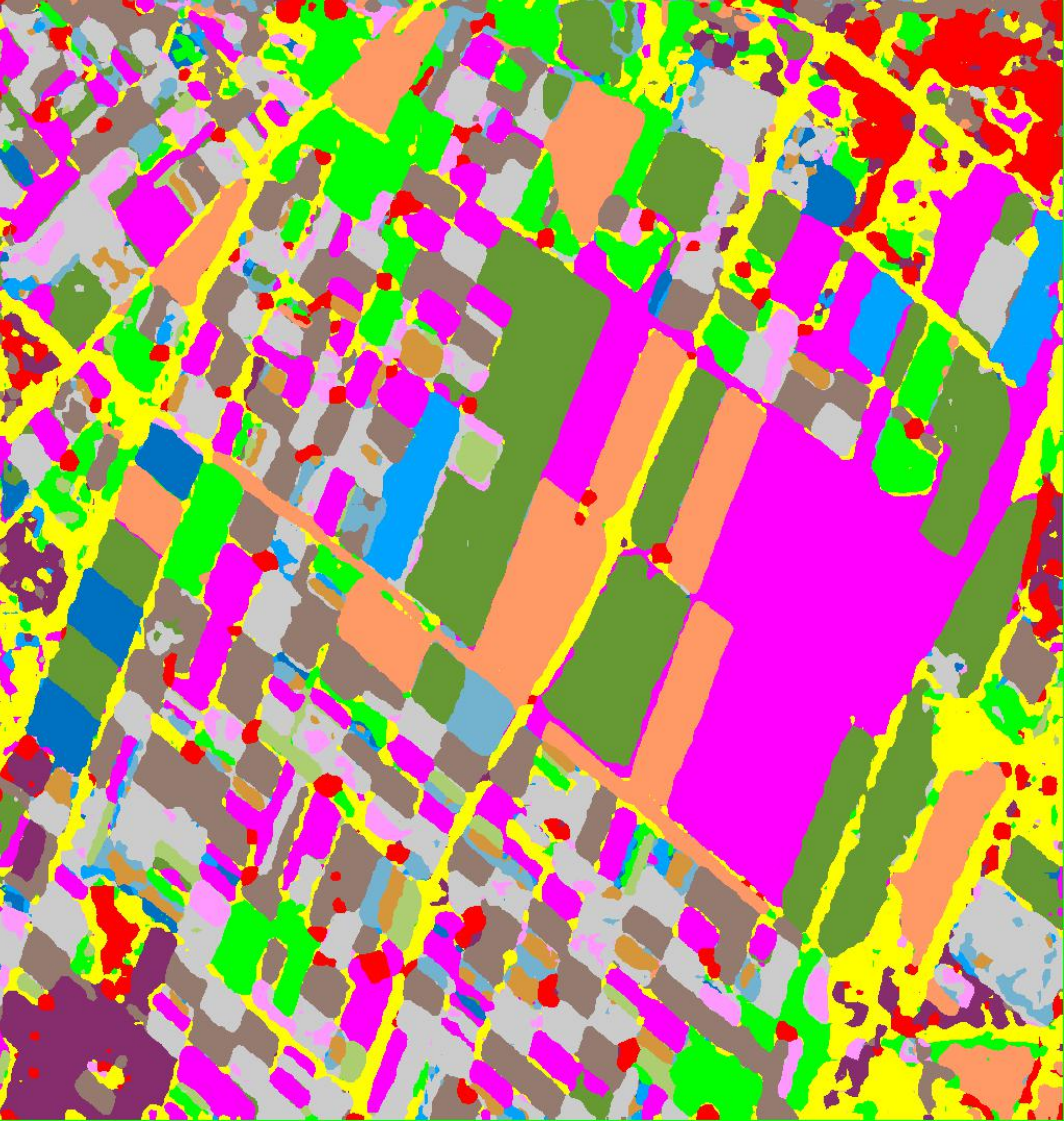}%
		\label{fig_Fle_CP_6}}
	\hfil
	\subfloat[]{\includegraphics[width=1.15in,height=1.15in]{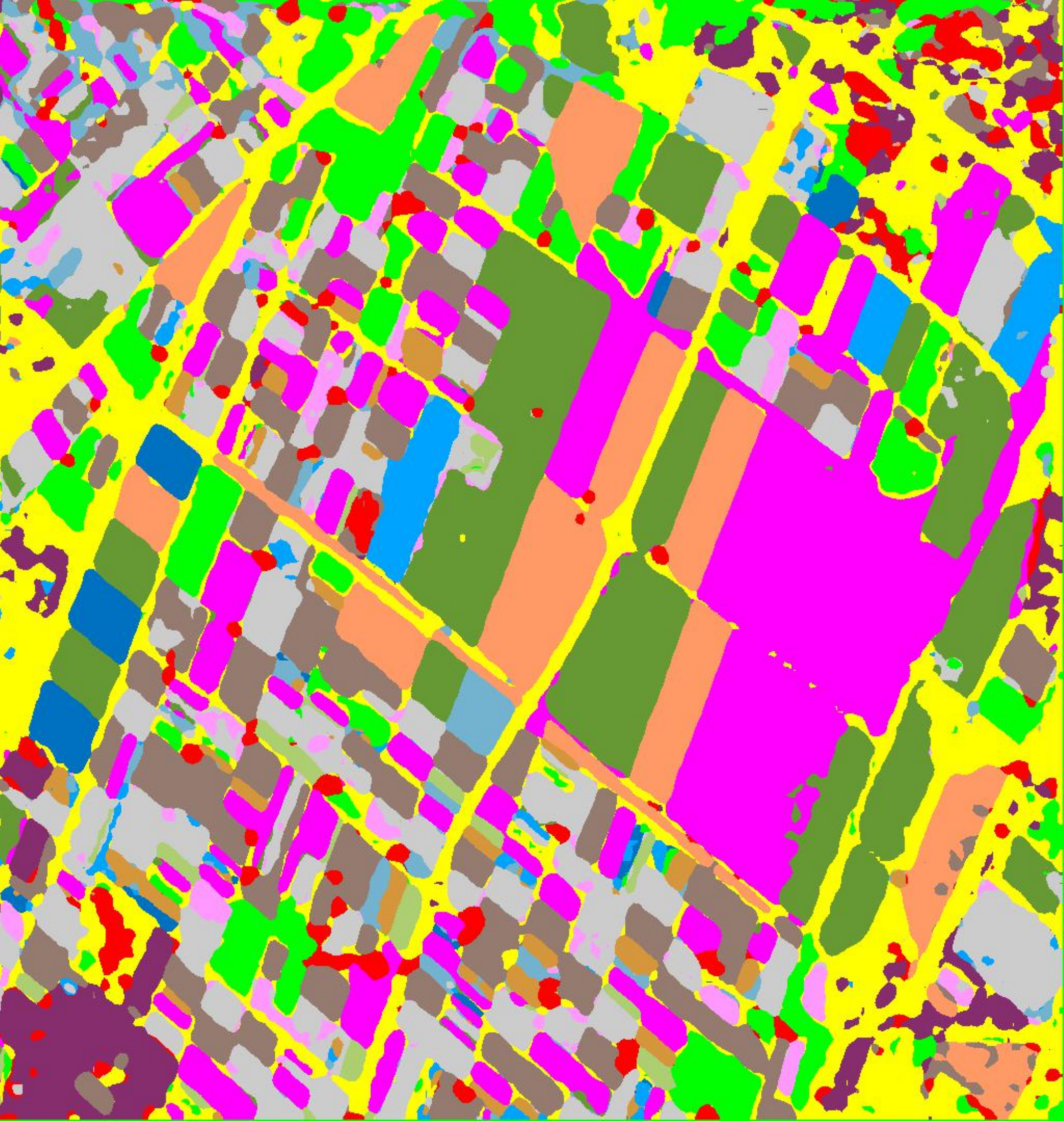}%
		\label{fig_Fle_CP_7}}
	\hfil
	\subfloat[]{\includegraphics[width=1.15in,height=1.15in]{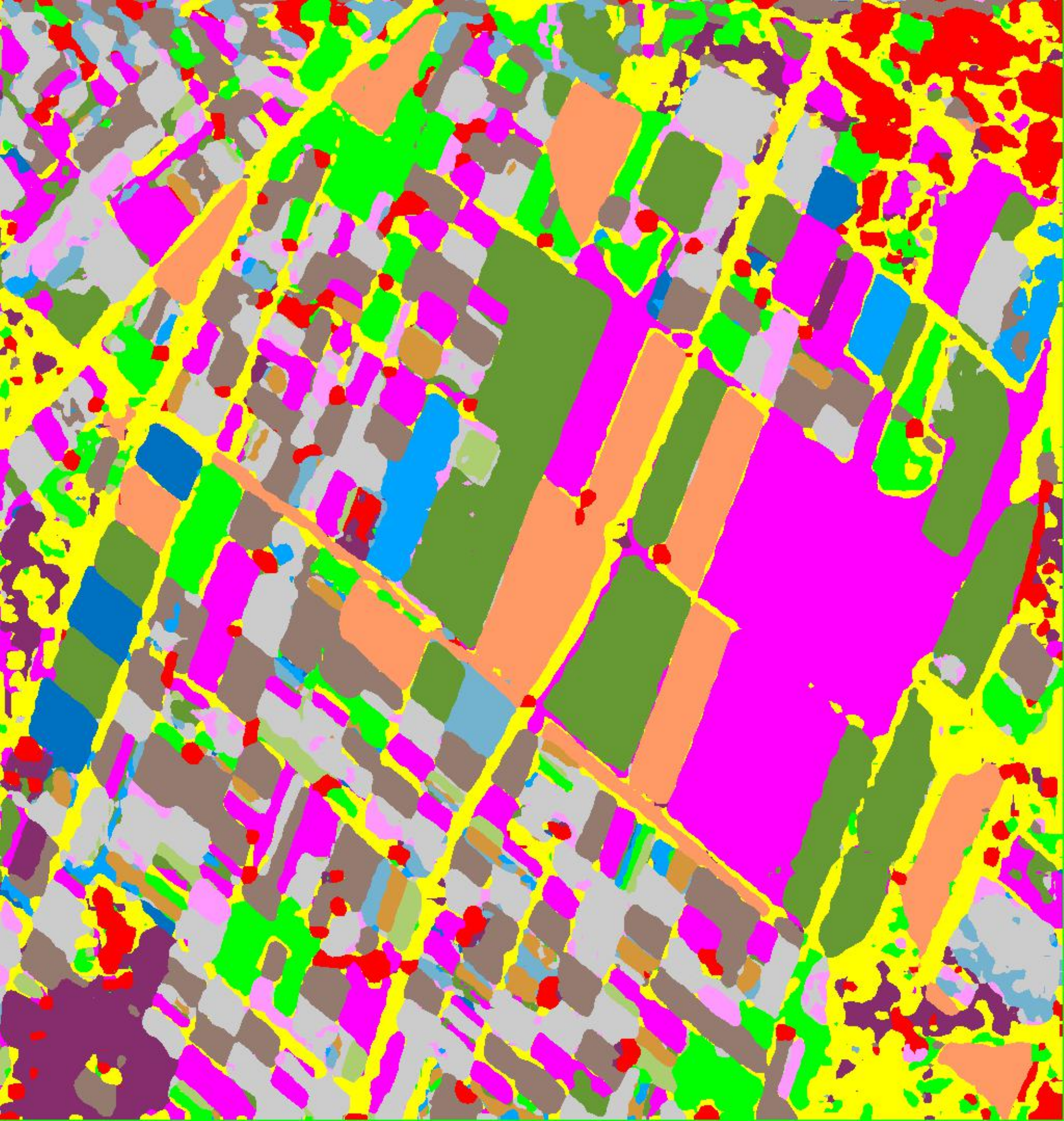}%
		\label{fig_Fle_CP_8}}
	\hfil
	\subfloat[]{\includegraphics[width=1.15in,height=1.15in]{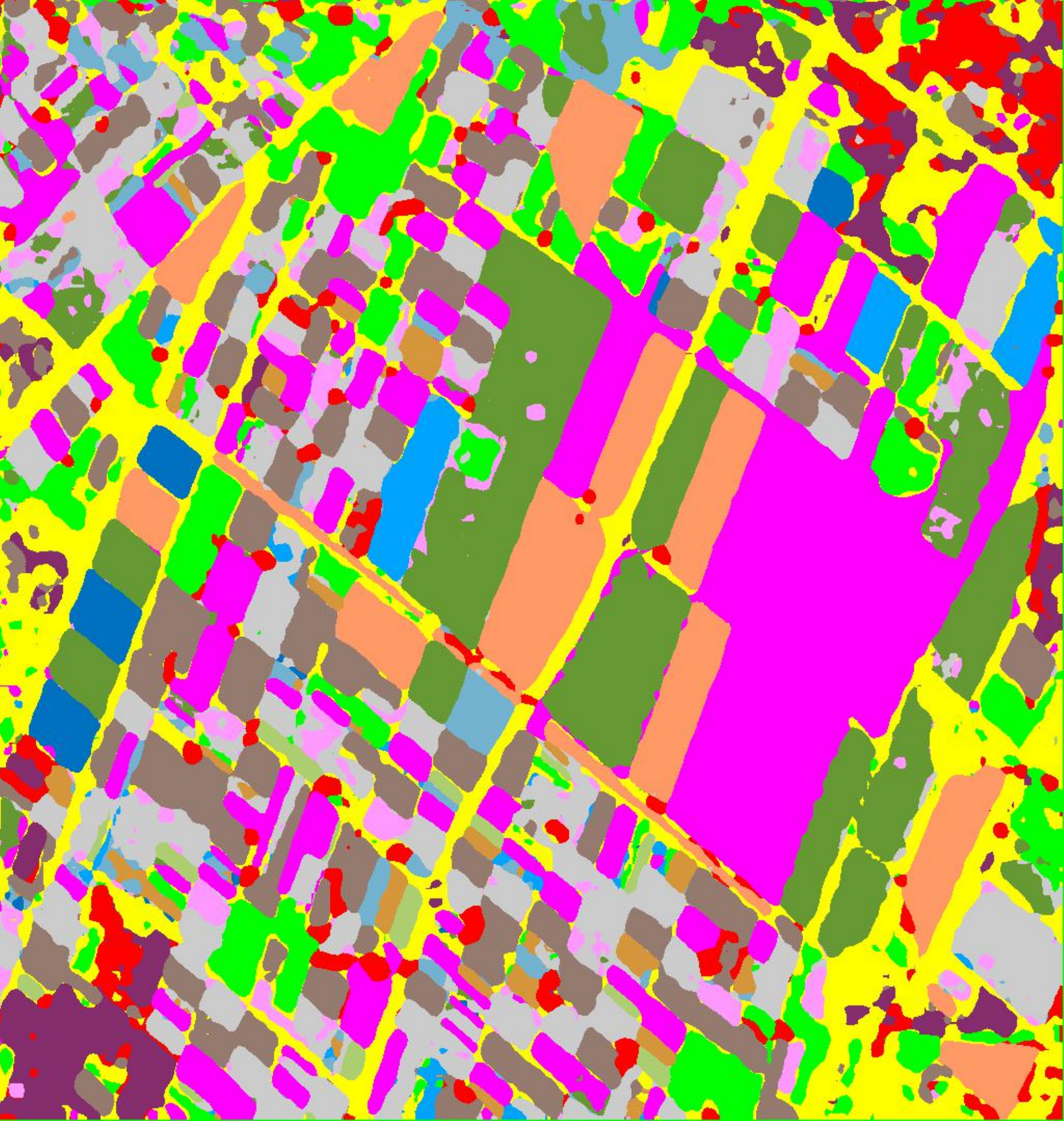}%
		\label{fig_Fle_CP_9}}
	\hfil
	\subfloat[]{\includegraphics[width=1.15in,height=1.15in]{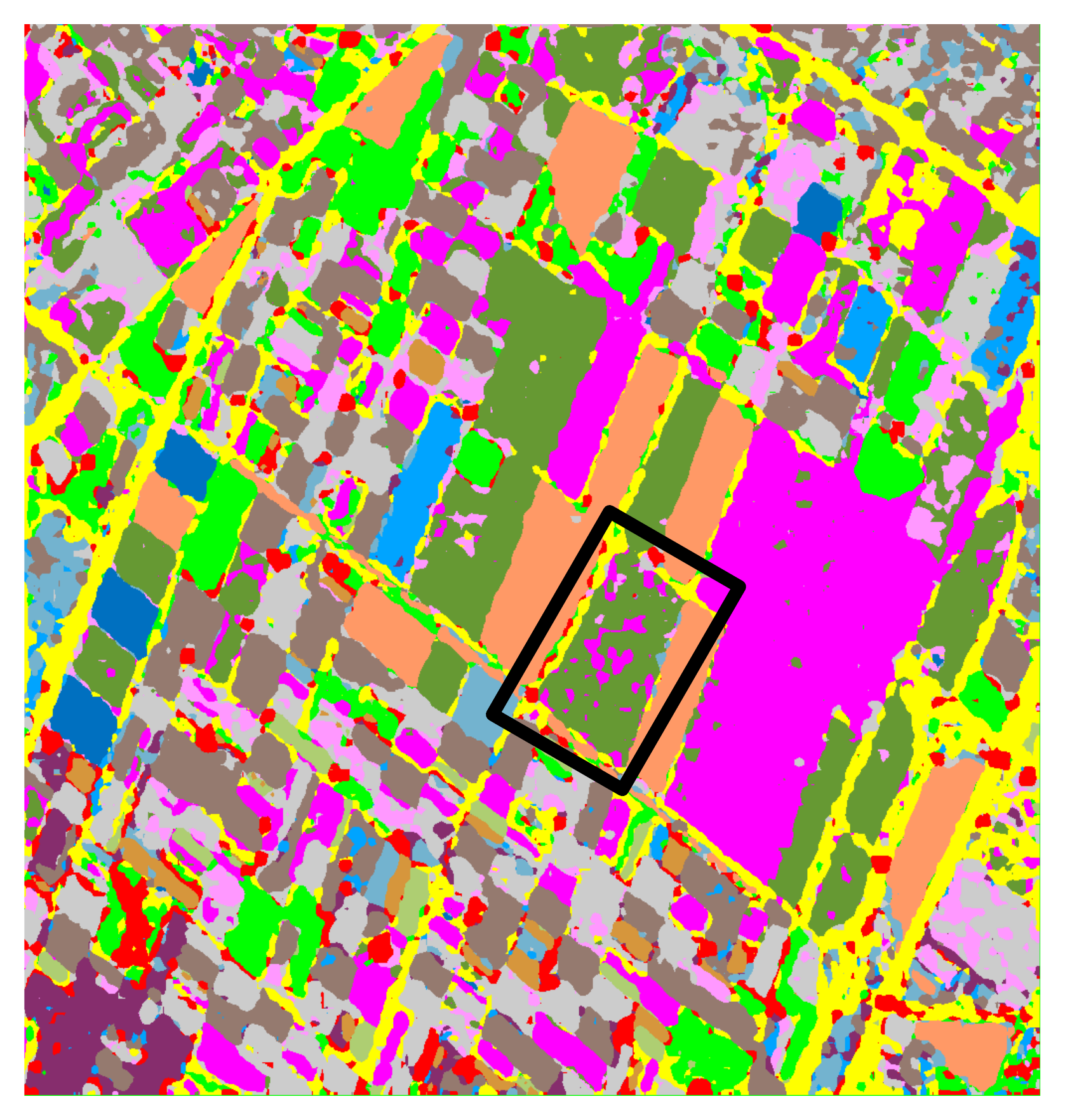}
		\label{fig_Fle_CP_C}}
	\hfil
	\subfloat[]{\includegraphics[width=1.15in,height=1.15in]{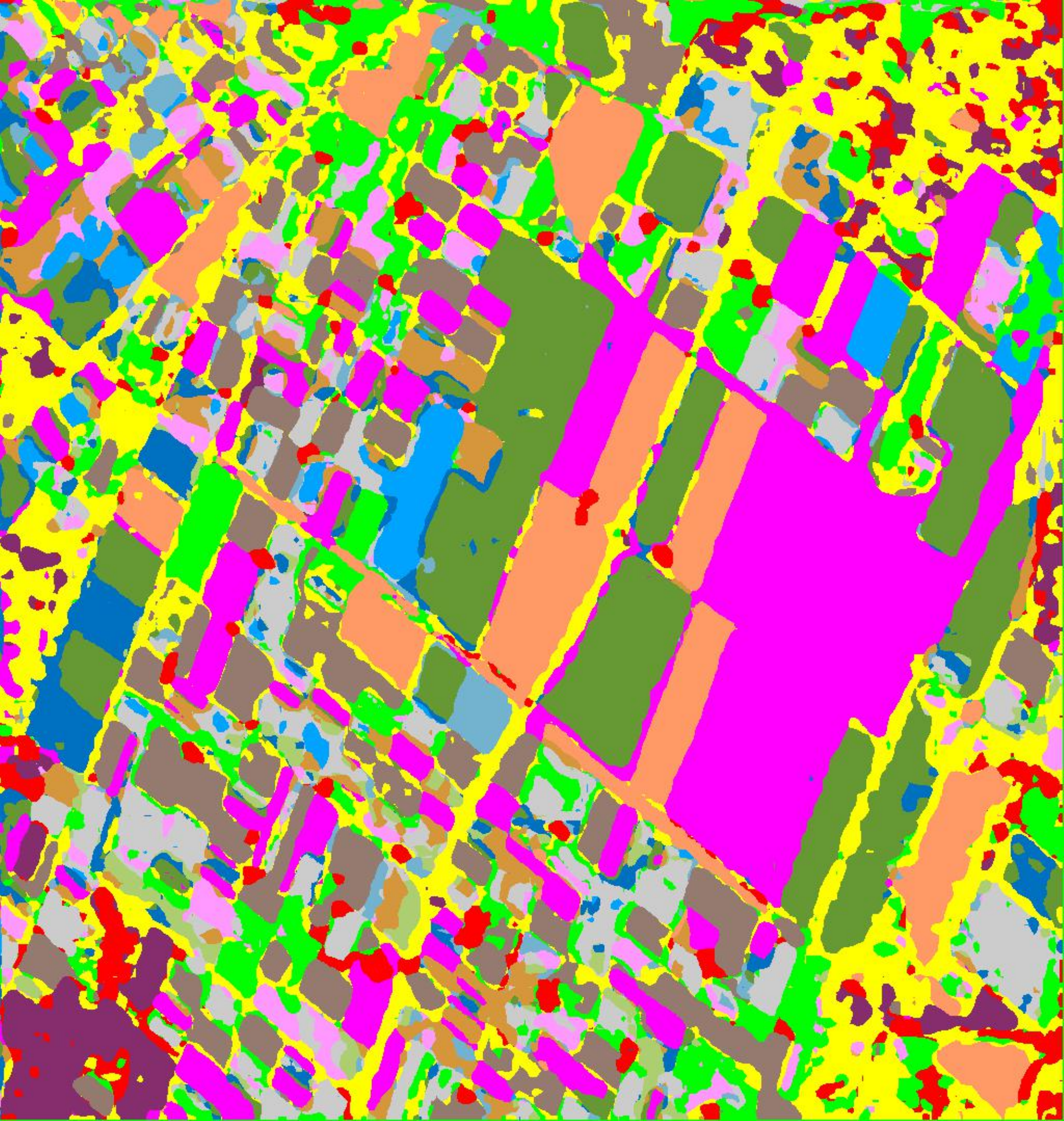}
		\label{fig_Fle_CP_P}}
	\hfil
	\subfloat[]{\includegraphics[width=1.15in,height=1.15in]{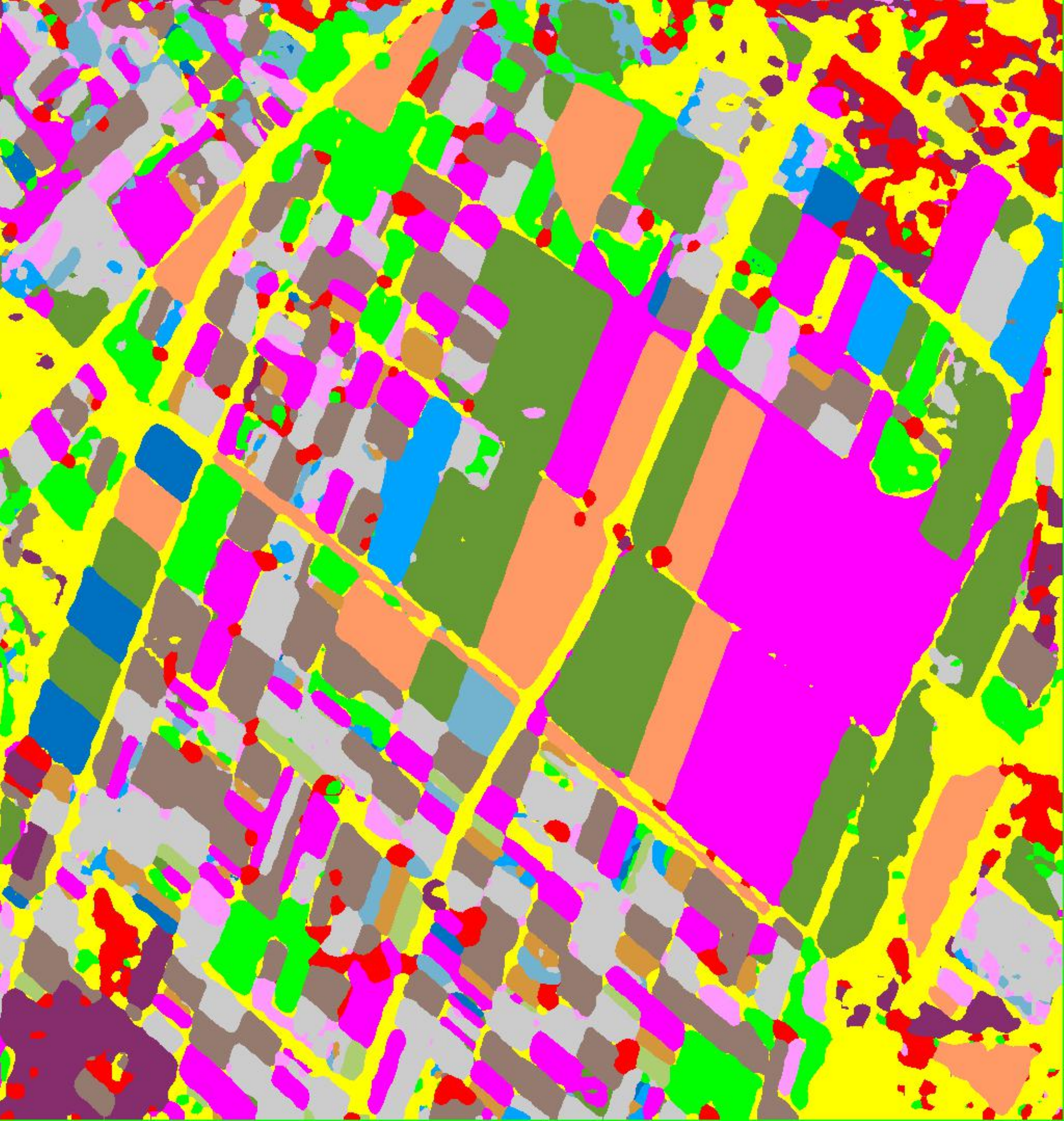}%
		\label{fig_Fle_CP_ours}}
	\caption{Classification maps obtained by different algorithms on Flevoland\_CP. (a) Ground Truth. (b) tbCNN. (c) S2ENet.(d) GLT.(e) ExViT.(f) mCrossPA.(g) SepDGConv.(h) AsyFFNet. (i) CMGFNet. (j) Ours(C). (k) Ours(P). (l) Ours(C+P).}
	\label{Fig_Fle_CP}
\end{figure*}

\begin{figure}[!t]
	\centering
	\subfloat[]{\includegraphics[width=1.5in]{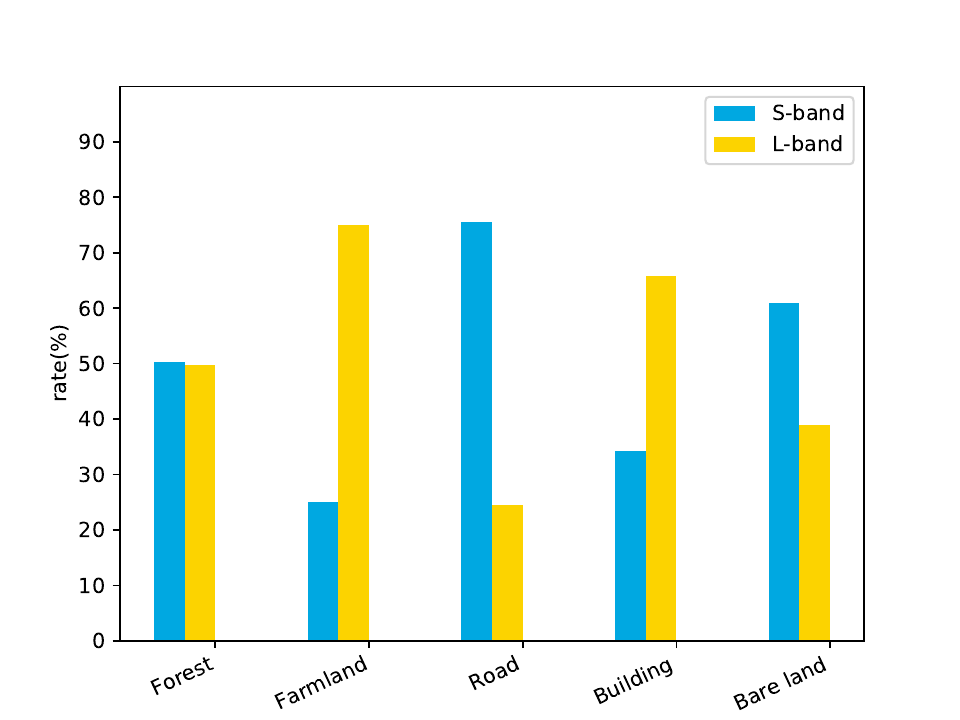}%
		\label{band_HB_SL}}
	\hfil
	\subfloat[]{\includegraphics[width=1.5in]{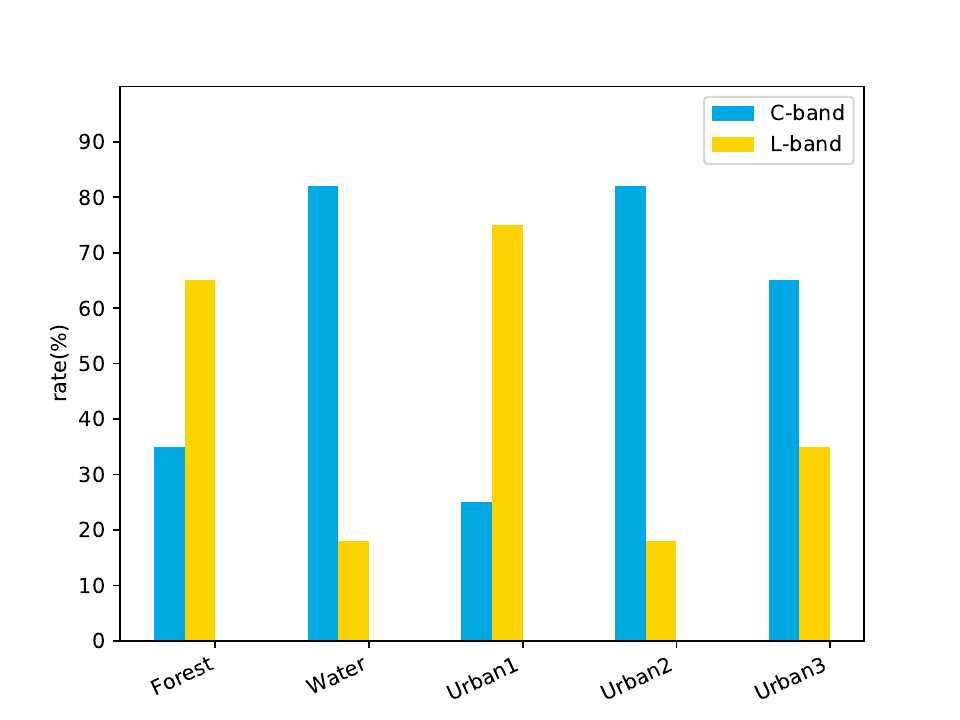}%
		\label{band_San_CL}}
	\hfil
	\subfloat[]{\includegraphics[width=1.5in]{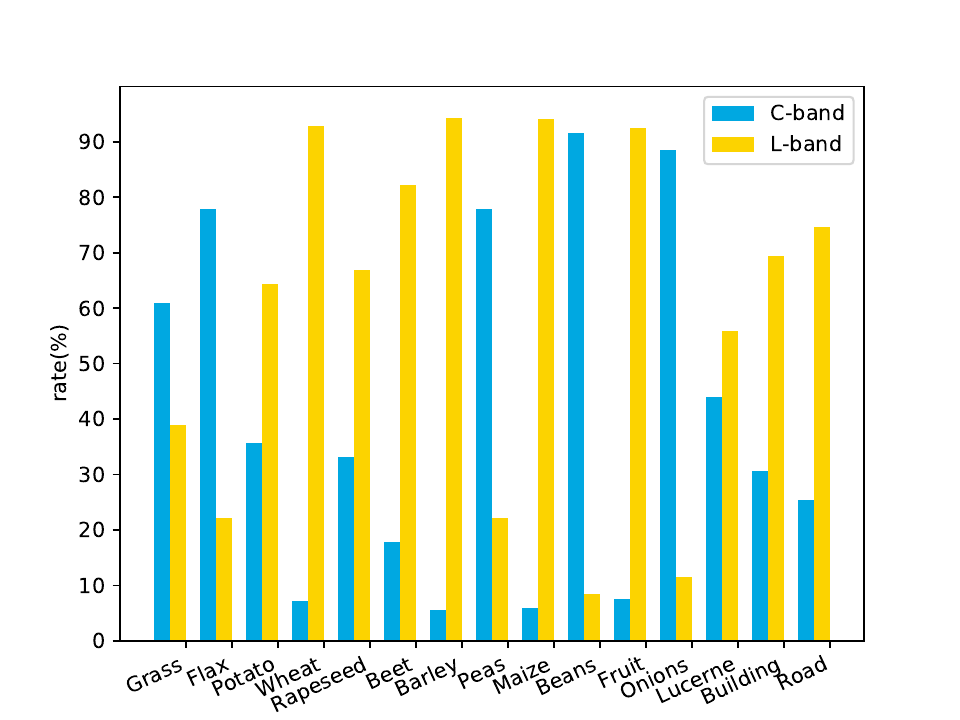}%
		\label{band_Fle_CL}}
	\hfil
	\subfloat[]{\includegraphics[width=1.5in]{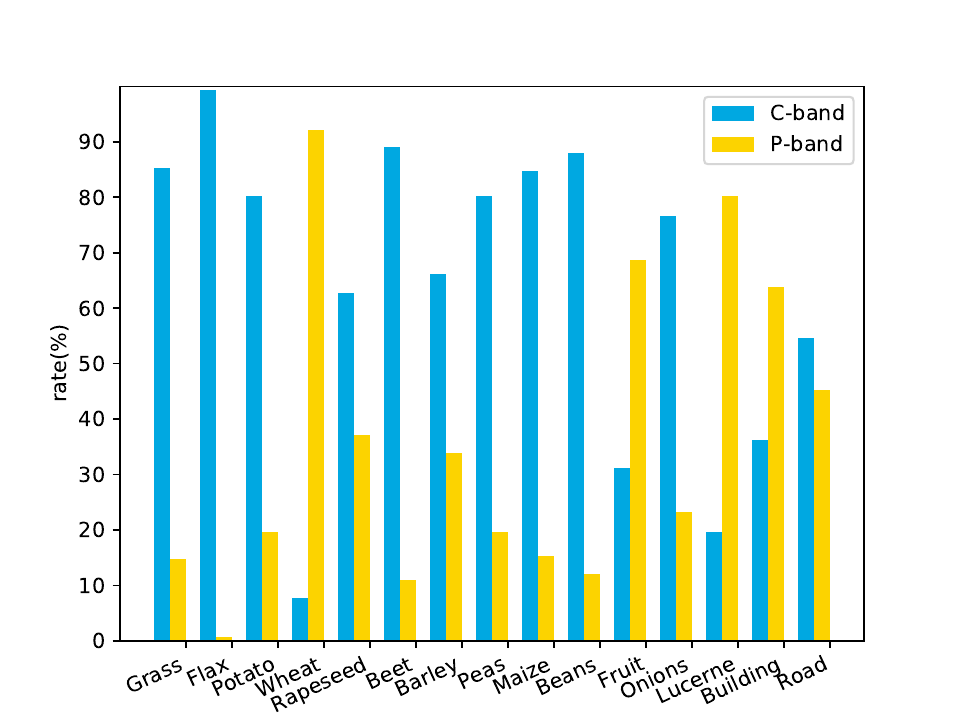}%
		\label{band_Fle_CP}}
	\hfil
	\caption{The probability of each single-frequency branch serving as a teacher model. (a) Hebei\_SL. (b) SanFrancisco\_CL. (c) Flevoland\_CL. (d) Flevoland\_CL. }
	\label{Fig_band}
\end{figure}

\subsubsection{Results on Flevoland\_CP}
For quantitative evaluation, Table \ref{Tab_Fle_CP} shows the classification accuracy of different algorithms on Flevoland\_CP. As shown in Table \ref{Tab_Fle_CP}, ours(C+P) outperforms ours(C) and ours(P) in classification accuracy for various categories, and has significantly better classification accuracy than other algorithms in many categories. Besides, the proposed dual-frequency collaborative classification algorithm has the highest accuracy in describing overall indicators such as OA, AA, and $\kappa$. 

For qualitative evaluation, Fig. \ref{Fig_Fle_CP} shows the classification maps of different algorithms. In Fig. \ref{fig_Fle_CP_C}, there are some clear misclassifications on the black rectangular area, while after the dual-frequency data fusion, the classification performance of this area is good in majority of comparison algorithms. This is because the P-band has better discrimination in this area, which supplements the discriminative features for C-band, resulting in better classification results for C+P. In addition, Fig. \ref{band_Fle_CP} shows the proportion of C-band branch and P-band branch being used as teacher models. It can be clearly seen that the C-band has a higher probability of being used as the teacher model in majority of categories, that is, the C-band is superior to the P-band in most categories. This conclusion is consistent with the results in Table \ref{Tab_Fle_CP} and Fig. \ref{Fig_Fle_CP}.

\section{Conclusion}\label{section5}
This article proposes an SKDNet-SSR method for dual-frequency PolSAR image collaborative classification. In the SKDNet-SSR structure, the SDSR module is constructed to improve the classification information learning under poor regional consistency conditions. It performs sample rectification based on the statistical Wishart distribution, which discards irrelevant noisy pixels and enables the feature interaction process of CNN and ViT to be carried out among informative pixels. In addition, based on the performances of different frequency PolSAR data in different categories and knowledge distillation technology, the DGSD module selects the best performing single-frequency branch on each sample as the teacher model, emphasizing the advantages of different frequency bands and achieving complementary learning of dual-frequency data. The ablation experiments on four real dual-frequency PolSAR datasets have demonstrated that the SDSR module can effectively improve the advanced feature extraction process, and the DGSD module can fully utilize the dual-frequency complementarity in land cover classification. In addition, the comparative experiments with several related algorithms reveal the advantages of the proposed SKDNet-SSR on dual-frequency PolSAR image classification.

\bibliographystyle{unsrt}
\bibliography{cite.bib}

\end{document}